\documentclass{IEEEtran}
\usepackage{epsfig}
\usepackage{graphicx}
\usepackage{amsmath}
\usepackage{amssymb}
\usepackage{rotating}
\usepackage{pdflscape}
\usepackage{hyperref}
\usepackage{flushend}
\usepackage{balance}
\usepackage{tabu}
\usepackage{slashbox,multirow}
\usepackage{hyperref}

\begin{document}

\title{TV News Commercials Detection using Success based Locally Weighted Kernel Combination}

\author{\IEEEauthorblockN{Raghvendra Kannao and Dr. Prithwijit Guha} \\
\IEEEauthorblockA{Department of Electronics and Electrical Engineering, IIT Guwahati\\
	Guwahati - 781039, Assam, India \\
	\{raghvendra, pguha\}@iitg.ernet.in
}}
\def\PaperID{0186}
\maketitle
\begin{abstract}
Commercial detection in news broadcast videos involves judicious selection of meaningful audio-visual feature combinations and efficient classifiers. And, this problem becomes much simpler if these combinations can be learned from the data. To this end, we propose an Multiple Kernel Learning based method for boosting successful kernel functions while ignoring the irrelevant ones. We adopt a intermediate fusion approach where, a SVM is trained with a weighted linear combination of different kernel functions instead of single kernel function. Each kernel function is characterized by a feature set and kernel type. We identify the feature sub-space locations of the prediction success of a particular classifier trained only with particular kernel function. We propose to estimate a weighing function using support vector regression (with RBF kernel) for each kernel function which has high values (near 1.0) where the classifier learned on kernel function succeeded and lower values (nearly 0.0) otherwise. Second contribution of this work is TV News Commercials Dataset of $150$ Hours of News videos. Classifier trained with our proposed scheme has outperformed the baseline methods on $6$ of $8$ benchmark dataset and our own TV commercials dataset.
\end{abstract}

\section{Introduction}

Commercial block detection in news broadcast videos have been attempted by both frequentist \cite{caco,adar,fcdb,ustc} and machine learning based approaches \cite{rlbt,rcds,tdot,evat}. The frequentist approach relies on the large number of repetition of advertisements and typically works with off-line stored data. The machine learning approaches, on the other hand, try to learn the characteristics of commercial shots and try to detect them on-the-run. The problem of detection of commercial shots in news broadcast videos involves a judicious selection of audio-visual features and suitable classifier(s). Researchers have identified a number of features based on presentation styles involving position of text, motion content, music content and other audio properties. In this work, we focus on the machine learning approach and present an intuitive idea for adaptive feature and suitable kernel type selection in the context of TV news commercials detection. 
Our main contributions in this work are --
\begin{itemize}
\item Proposal of ``Success based Locally Weighted Multiple Kernel Combination'' , a new Multiple Kernel Learning algorithm which uses a success based locally weighted linear combination of kernels. The goal of this proposals is to identify the locally best performing feature and kernel type combinations while suppressing the failed feature-kernel type combinations;

\item We have created a TV News Commercial Dataset of approximately $150$ hours of TV news videos which will be made available publicly. To the best of our knowledge this is the first publicly available dataset for TV news commercial detection which will enable benchmarking of different algorithms.
\end{itemize}

In classification problems, often selecting and fusing features available from different sources and modalities is a crucial problem. The fusion becomes even more difficult when different features have different notions of similarity. Various feature fusion techniques are well studied in the literature. The simplest one being ``early fusion'' where the features from different sources are concatenated to learn a single classifier. In case of early fusion technique poor feature selection often results in degraded performance\cite{noble:2004}. In the ``late fusion'' framework, different classifiers are trained on different feature sets or different training sets. Predictions of  these classifiers are further processed by a heuristic based or learned combiner algorithm to give final prediction\cite{Rokach2010}. The choice of combiner usually determines the overall performance of the final classifier. Bagging and boosting based approaches are some of the examples of late fusion technique. Third framework for feature fusion is ``Intermediate Fusion''\cite{noble:2004} using Multiple Kernel Learning (MKL, henceforth). In intermediate fusion a SVM is trained by combining multiple kernel functions with different features and kernel types. Empirical results in literature \cite{noble:2004,Sonnenburg:2006,Lanckriet:2004a,pavlidis:2001} have shown the superiority of intermediate fusion framework over early fusion and some of the late fusion techniques. In this work we have proposed an intermediate fusion technique .

The support vector machine (SVM) determines the discriminative hyperplane with  maximum margin in an implicitly induced feature space.  The discriminative hyperplane obtained after training is;
\begin{equation}
f(x) = \langle \boldsymbol{ w , \Phi(x)} \rangle + b = 0
\label{eq:hyperPlane}
\end{equation}

where, $ \boldsymbol{w}$ is hyperplane coefficient vector, $b$ is bias and $ \boldsymbol{\Phi(x)}$ is the mapping function. From the dual formulation of SVM, hyperplane coefficient vector $ \boldsymbol{w}$ can be substituted by $\sum_{i=1}^{N} \alpha_{i} y_{i} \boldsymbol{\Phi(x_{i}) }$. Equation \ref{eq:hyperPlane} can be rewritten as;
\begin{equation}
f(x) = \sum_{i=1}^{n} \alpha_{i} y_{i} \underbrace{\langle \boldsymbol{ \Phi(x_{i}) , \Phi(x)} \rangle}_{\boldsymbol{k(x_{i},x)}} + b = 0
\label{eq:kernel}
\end{equation}
where, $n$ are the number of training instances, $\boldsymbol{x_{i}}$ having labels $y_{i}$ and Lagrange multiplier $\alpha_{i}$. The inner product $\langle \boldsymbol{ \Phi(x_{i}) , \Phi(x)} \rangle$ can be replaced by a function called kernel function. The kernel function $\boldsymbol{k(x_{i},x)}$ computes the similarity between pair of data points thus avoids explicit definition of the mapping function $ \boldsymbol{ \Phi(.)}$. Different kernel functions and hence mapping functions leads to different hyperplanes in original feature space. Hence choosing proper kernel function is decisive step in training SVM based classifier and usually selected by cross validation.

Several different types of general and domain specific kernels are proposed in the literature. Each kernel has different similarity measure and captures different representation from the features. When multiple features are available, instead of using single kernel by concatenating all the features, multiple kernels can be simultaneously in  MKL framework.

Using combination multiple kernels not only enables the use of different similarity measures for different features but also allows feature selection by learning the weights for each kernel.

MKL is a well studied problem and a vast literature is available on the same. While combining multiple kernels, each one is associated with a non-negative weight (which defines its importance) and they can be combined either linearly or non linearly.

Gonen et. al \cite{alpaydinSurvey:2011} in a latest survey paper presented the taxonomy for different multiple kernel learning methods. They have identified six key properties for characterizing MKL algorithms -- viz. learning method, functional form, the target function, the training method, the base learner, and computational complexity. Based on these six key properties, the MKL algorithms are grouped into twelve different categories. In \cite{pavlidis:2001} an unweighted sum of heterogeneous kernels ( each kernel has the same weight) performed well over combination of SVMs trained on individual features. Diego et.al.\cite{Diego:2004} have proposed to use data dependent weight for kernels. The weights for kernels were set to conditional class probabilities estimated using nearest neighbor approach; while Tanabe et. al.\cite{tanabe:2008} have used the F-measure of the classifier trained on individual kernels as weight of the kernels in linear combination. The approach proposed in \cite{tanabe:2008} is one of the simplest method for combining multiple kernels. The hyperplane for combined kernel SVM is given by ;
\begin{equation}
f(x) = \sum_{m=1}^{p} \eta_{m} \sum_{i=1}^{n} \alpha_{i} y_{i} \underbrace{\langle \boldsymbol{ \Phi_{m}(x_{i}) , \Phi_{m}(x)} \rangle}_{\boldsymbol{k_{m}(x_{i},x)}} + b = 0
\label{eq:kernelMKL}
\end{equation}
where, $p$ are the number of kernels, $\eta_{m}$ is the weight of $m^{th}$ kernel $\boldsymbol{k_{m}(.,.)}$.
Apart from heuristics and data dependent methods, kernel weight estimation is also formulated as an optimization problem. The kernel weights are selected such that it optimizes one of the properties of the classifiers and/or kernel. Various properties of a classifier/kernel include structural risk, kernel similarity, kernel alignment and VC dimension. Kandola et.al. \cite{kandola:2002} proposed the estimation of non-negative kernel weights by formulating it as an optimization problem to maximize the alignment between a non negative linear combination of kernels and the ``ideal kernel''. In \cite{he:2008} instead of optimizing the kernel alignment, distance between combined kernel matrix and the ideal kernel is optimized. Varma et.al \cite{varma:2007} formulated the linear kernel weight combination as a single step structural risk minimization problem with regularized non-negative kernel weights. In \cite{ong:2008}, the proposed approach learns a kernel function instead of kernel weights for individual kernels to minimize the structural risk where the kernel function includes convex combinations of an infinite number of point-wise non-negative kernels. While semi infinite programming is used in\cite{shogun:2010}.

Alpaydin et.al. \cite{alpaydin:2008} proposed a Localized Multiple Kernel learning (L-MKL, henceforth) for estimating the kernel weights locally, by defining the region of influence of each kernel. A gating model defined by a combination of perceptrons decides the weights for kernels. The weights were estimated using a two step optimization process. In the first step, the parameters of the canonical SVM (Lagrange multipliers) are estimated by keeping the parameters of the gating model fixed. In second step, the parameters of the gating model ( perceptron weights ) are re-estimated. This two step process is continued till convergence. The gating model non-linearly selects the weights for each kernel depending on the data points. In \cite{christoudias:2009} a Gaussian Process framework was used for combining different feature representations in a data dependent way using a Bayesian approach. Boosting and ensemble learning based methodologies are also proposed in the literature \cite{Bennett:2002}. Extensive Literature Review of MKL methods is out of scope of this work.
Most recent works have focused on either domain specific kernels \cite{pyramidKernel} or optimization based MKL with more focus on faster convergence, reducing number of support vectors etc.\cite{ManikVarma:2014, Castro2014}. Though most recent methods have almost comparable performance with approach proposed in \cite{shogun:2010} hence can be used as benchmark. Interested readers may refer to the survey on MKL by Gonen and Alpaydin \cite{alpaydinSurvey:2011} and a recent survey in the context of visual Object recognition \cite{pamiMKL:2014}.

In the proposed approach, the video stream is first segmented into shots based on color distribution consistency. Audio-visual features computed from these shots are used to characterize the commercials. We have used existing features from the literature viz. shot length \cite{rtcd}, scene motion distribution \cite{rlbt,ZhuJacobsonPanEtAl2011}, overlay text distribution \cite{evat}, zero crossing rate \cite{aiac,rcds}, short time energy (STE) \cite{rcds}, fundamental frequency, spectral centroid, flux and roll-off frequency \cite{evat} and MFCC Bag of Words \cite{MaehlingEwerthZhouEtAl2012}. We observed that, SVMs trained on a certain set of features fail to detect the commercial shots when ever the basic assumption involving those features are violated. Moreover features extracted from different modalities have different notions of similarity.

This motivated us to use a intermediate fusion (MKL) approach. We combine different kernel functions linearly. Each kernel function ( or kernel) is characterized by a feature and kernel type( e.g. linear, RBF etc.). We also identify the points in feature sub-spaces where individual classifiers trained with particular kernel function succeed. We use this success information to estimate a weighing function using support vector regression (with RBF kernel only). This success based weighing functions are directly used as the linear combination parameters for multiple kernels thereby producing a locally weighted kernel combinations linked to kernel function success. The motivation of this approach was to enhance the kernels from successful feature-kernel type combinations while suppressing the failed ones.We have benched marked our results on our own commercial shot datasets of $150$ hours along with $8$ standard data sets to verify our claim.

This paper is organized in the following manner. In Section~\ref{sec:comFeat}, we briefly describe the different audio-visual features used for characterizing commercials. The proposal of the success based locally weighted kernel combinations is explained in Section~\ref{sec:mkl}. The TV news commercial dataset is described in Section~\ref{sec:dataset}. The results of experimentation in terms of comparative f-measures and generalization performances and discussions on results are presented in Section~\ref{sec:expt}. Finally, we conclude in Section~\ref{sec:conc} and outline the future extensions.

\section{Audio-Visual Features}
\label{sec:comFeat}

\begin{figure*}[tbph]
\centerline{\includegraphics[width=0.9\textwidth]{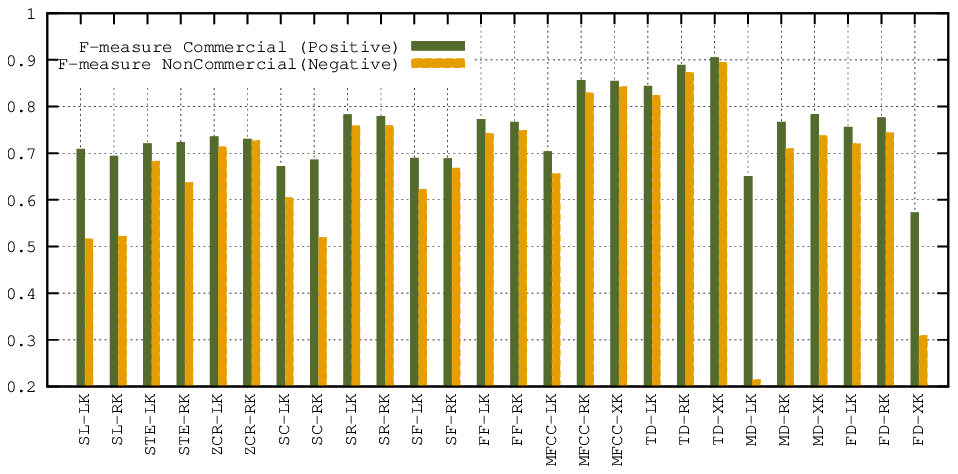}}
\caption{\small{The performance analysis in terms of F-measures for both commercial and non-commercial shot classification with different feature-kernel combinations. We have used linear (LK) and RBF (RK) kernels for all the features and the $\chi^2$ kernels (XK) for motion, frame difference, MFCC BoW and text distribution features. Note the varying capabilities of the different feature-classifier combinations and their biases towards positives and negatives. Also, we observe that scene text distribution and MFCC Bag of Words using SVMs with RBF or $\chi^{2}$ kernel outperforms the other feature-classifier combinations.}}
\label{fig:featAnal}
\end{figure*}
We choose a video shot as basic unit for commercial detection as shot boundaries will mostly overlap with commercial- non commercial boundary. The television video broadcast is first segmented into shots based on simple color distribution consistency \cite{Koprinska2001}. We extract $11$ different audio-visual features from each video shots which are used to characterize the commercials and are briefly described as follows.

\textbf{Video Shot Length} \cite{rlbt} is considered as a discriminating feature as the commercial shots are mostly of very short duration compared to news reports. \textbf{Overlay Text Distribution} has been used as an important clue for identifying commercials \cite{LiuZhaoZhuEtAl2010}. It is observed that the major ticker text bands situated in the upper and lower portions of the scene are generally present during news and other programs. However, during commercials only the lower most band remains \cite{6123389} while commercial specific small text patches containing product information appear through out the frame. Following existing work \cite{evat}, we have divided the scene into a $5 \times 3$ grid and have constructed a $30$ dimensional feature vector storing mean and variance of the fractions of text area in each grid block of each frame over entire shot. We have used the method described in \cite{Shivkumar:PAMI2011} for the purpose of text detection. The \textbf{Motion Distribution} is a significant feature as many previous works have indicated that commercial shots mostly have high motion content as they try to convey maximum information in minimum possible time. This motivates us to compute dense optical flow (Horn-Schunk formulation) between consecutive frames and construct a distribution of flow magnitudes over the entire shot with $40$ uniformly divided bins in range of $[0,40]$ \cite{rlbt,ZhuJacobsonPanEtAl2011}. Often pixel intensities of regions suddenly change while the boundaries of the region do not move. Such changes are not registered by optical flow. Thus, \textbf{Frame Difference Distribution} is also computed along with flow magnitude distributions. We obtain the frame difference by averaging absolute frame difference in each of $3$ color channels and the distribution is constructed with $32$ bins in the range of $[0,255]$ \cite{rlbt}.

\textbf{Short Time Energy} (STE, henceforth) is defined as sum of squares of samples in an audio frame. To attract user's attention commercials generally have higher audio amplitude leading to higher STE \cite{rcds}. The \textbf{Zero Crossing Rate} measures how rapidly an audio signal changes. ZCR varies significantly for non pure speech (High ZCR), music(Moderate ZCR) and speech(Low ZCR). Usually commercials have background music along with speech and hence the use of ZCR as a feature \cite{aiac,rcds}. Audio signals associated with commercials generally have high music content and faster rate of signal change compared to that of non-commercials \cite{rlbt}. This motivated the use of spectral features where higher \textbf{Spectral Centroid} signify higher frequencies (music), higher \textbf{Spectral Flux} indicate faster change of power spectrum and \textbf{Spectral Roll-Off Frequency} discriminates between speech, music and non-pure speech \cite{evat}. Along with the spectral features, \textbf{Fundamental Frequency} is also used as non-commercials (dominated by pure speech) will produce lower fundamental frequencies compared to that of commercials (dominated by music) \cite{aspRao}. For all the above mentioned audio features, we have used the non overlapping frames of $20$ msec duration and sampling frequency of $8000$ Hz. The Mean and standard deviation of  all audio feature values are calculated over the shot, generating a $2D$ vector for each feature.

The \textbf{MFCC Bag of Audio Words} have been successfully used in several existing speech/audio processing applications \cite{MaehlingEwerthZhouEtAl2012}. This motivated us to compute the MFCC coefficients along with Delta and Delta-Delta Cepstrum from $150$ hours of audio tracks. These coefficients are clustered into $4000$ groups which form the Audio words. Each shot is then represented as a Bag of Audio Words by forming the normalized histograms of the MFCC co-efficients extracted from overlapping windows in the shots.

Existing approaches have experimented with different combinations of the above mentioned features while constructing higher dimensional vectors by concatenating the different feature vectors. Classifiers (mainly SVM, AdaBoost etc.) learned on those feature spaces have been used to detect the commercial blocks. We observe that at different locations of the feature space, a particular combination of features is generally successful in identifying the commercial shots. This motivated us to propose a spatially varying composition of kernels, weights of each one being calculated based on local success. These locally varying weights effectively work as feature selectors. Our proposed methodology for Success based locally weighted multiple kernel learning is described next.

\section{Success based Locally Weighted Kernel Combination}
\label{sec:mkl}

Consider a binary classification problem where $y_{i} \in \{-1, +1 \}$ is the class label of $D$ dimensional instance $ \boldsymbol{x}_{i}$. Let, the training data set containing $n$ independent and identically distributed instances be $ \boldsymbol{S} = \{ ( {x}_{i},y_{i} ); i=1, \ldots n \}$. Each data instance $ \boldsymbol{x}_{i}$ consists of $m$ different kinds of features such that $ \boldsymbol{x}_{i} = [ ^{1}\boldsymbol{x}_{i}, \ldots ^{j}\boldsymbol{x}_{i}, \ldots ^{m}\boldsymbol{x}_{i} ]^{T}$ where the leading superscript denotes the $j^{th}$ ($j=1, \ldots m$) feature of the $i^{th}$ data vector in $S$. The $j^{th}$ ($j=1, \ldots m$) feature has $D_{j}$ dimensions.

Solving such classification problems often involve a scheme for selecting a suitable combination of features to maximize the performance. Moreover if SVM is used as classifier,  selecting appropriate feature and suitable kernel type ( and it's parameters) are very crucial steps in training. Generally features, kernel type and its parameters are selected by cross-validation.
We propose to use linear combinations of various feature and kernel types  ( Each pair is a kernel function or kernel) in multiple kernel learning framework where weight for each kernel function are learned locally. Let, $q_{j}$ be the number of kernel types ( e.g.$RBF, \chi^{2}, Linear$) used with the $j^{th}$ feature.  Thus, we will have a total of $q = \sum_{j=1}^{m} q_{j}$ number of kernel functions. $\boldsymbol{k}_{jr}(.,.)$ ($j=1, \ldots m;r = 1, \ldots q_{j}$) denotes the kernel function (or kernel) defined for $j_{th}$ feature with $q_{j}^{th}$ kernel type.

One of the simplest formulation for multiple kernel learning is proposed by Tanabe et.al.\cite{tanabe:2008}. They have used the F-measure ( on cross-validation set) of the classifier $\boldsymbol{C}_{jr}$ classifier as linear combination weight for $\boldsymbol{k}_{jr} ^{th}$ kernel in MKL. The Classifier $\boldsymbol{C}_{jr}$  ($j=1, \ldots m;r = 1, \ldots q_{j}$) is learned over the training set $\boldsymbol{S}_{j} = \{ ( ^{j}\mathbf{x}_{i},y_{i} ); i=1, \ldots n \}$ with $q_{r} ^{th}$ kernel type. Hyperplane of F-measure weighted multiple kernel SVM is given by;
\begin{eqnarray}
f(x)=\sum_{j=1}^{m} \sum_{r=1}^{q_{j}} \beta_{jr} \sum_{i=1}^{n} \alpha_{i} y_{i} \underbrace{\langle \boldsymbol{ \Phi_{jr}(x_{i}) , \Phi_{jr}(x)} \rangle}_{\boldsymbol{k_{jr}(x_{i},x)}} + b = 0 \\
\beta_{jr}=\frac{\eta_{jr}}{\sum_{j=1}^{m} \sum_{r=1}^{q_{j}} \eta_{jr}} \nonumber 
\label{eq:kernelMKLjr}
\end{eqnarray}
where, $\eta_{jr}$ is the  F-measure of the $\boldsymbol{C}_{jr} ^{th}$ classifier which acts as weight of a kernel. However, in most practical cases, fixed set of classifier weights $\alpha_{jr}$ over the entire feature space have not shown great performance, specially in cases having high intra-class variance \cite{alpaydin:2008}.

We note that the classification success is rather a local phenomenon. For cases involving many kernel functions -- where a set of kernel functions could not linearly separate ( misclassification) the data even in kernel space, another complimentary set of kernel functions may succeed in linearly separating ( correct classification) the data without over-fitting~\cite{alpaydin:2008}. This motivates us to learn a set of spatially varying weighing functions $g_{jr}$ for every kernel $\boldsymbol{k}_{jr}(.,.)$ which will have higher values (near to $1.0$) in the zones of the classifier success and very low values (nearly $0.0$) otherwise. Such a success based weighing scheme will assign more importance to useful kernel functions while suppressing the erroneous predictions in the classifier output.

To learn the function $g_{jr}$, we create the training data set $\boldsymbol{S}_{jr} = \{ ^{j}\boldsymbol{x}_{i} , \delta( \hat{y}_{ijr} - y_{i} ); i = 1, \ldots n \}$ where, $\delta(.)$ is the Kronecker Delta function and $\hat{y}_{ijr}$ is the class label predicted by the classifier $\mathcal{C}_{jr}( ^{j}\boldsymbol{x}_{i} )$ for the data vector $\boldsymbol{x}_{i}$.  The function $\hat{g}_{jr}$ is then estimated by using Support Vector Regression using RBF kernels. Thus, in the proposed framework of success based locally weighted multiple kernel learning, the discriminative hyperplane is given by 
\begin{equation}
f(x) = \sum_{i}^{n} \alpha_{i} y_{i} \mathcal{K}(\boldsymbol{x , x_{i}}) = 0
\label{eq:swk}
\end{equation}
where, the combined kernel function  $\mathcal{K}(\boldsymbol{x , x_{i}})$ is 
\begin{equation}
\mathcal{K}(\boldsymbol{x , x_{i}}) = \frac{ \sum_{j=1}^{m} \sum_{r = 1}^{q_{j}} \hat{g}_{jr}( ^{j}\boldsymbol{x})   \boldsymbol{k}_{jr}( ^{j}\boldsymbol{x}, ^{j}\boldsymbol{x}_{i}) \hat{g}_{jr}( ^{j}\boldsymbol{x}_{i} ) }{ \sum_{j=1}^{m} \sum_{r = 1}^{q_{j}} \hat{g}_{jr}( ^{j}\boldsymbol{x} ) \hat{g}_{jr}( ^{j}\boldsymbol{x}_{i} ) }
\end{equation}

We note that the values of $\hat{g}_{jr}$ always lie in the interval $[0,1]$ and hence the above expression provides a non-negative linear combination of individual kernel functions. It can also be shown that the proposed linear combination of the kernel functions satisfy the Mercer's condition \cite{NCSVM:2000} and hence $\mathcal{K(.,.)}$ can be used as a kernel function for learning a single SVM based classifier. Also, this linear combination is weighted by the success level predictions ( $\hat{g}_{jr}( ^{j}\boldsymbol{x})$, $\hat{g}_{jr}( ^{j}\boldsymbol{x}_{i})$ ) of both the inputs ($^{j}\boldsymbol{x}$, $^{j}\boldsymbol{x}_{i}$ ) of the kernel function thereby enhancing the contributions from successful kernel functions at particular instance while suppressing the failure cases.

The performance of the proposed approach was found to be superior compared to two baseline MKL methods over $8$ standard datasets and our own commercial shot dataset. The proposed approach has provided better performance on all the data sets compared to the baseline methods. The visualization of proposed method on a 2D toy dataset is shown in Figure~\ref{fig:toydata}. Next we describe our TV News commercial dataset.

\section{TV News Commercials Dataset}
\label{sec:dataset}
TV News commercial detection is semantic video classification problem. Though while classifying commercials in most of the approaches presentation format dominates the actual content of the videos. The domination of presentation format can be justified by the large intraclass variability and interclass similarity of commercials as well as news. For example, a car may appear in commercials as well as non commercials (same content). The presentation format typically includes placement of overlay text, shot duration, background music etc. and are defined by the editing policy of each channel. Hence there is significant amount of variations among different News channels. 

To best of our knowledge no TV news commercial detection dataset is publicly available. Hence benchmarking and comparing different commercial detection algorithms is tough. We have created a TV News commercials detection dataset of approximately $150$hours of TV news broadcast with $30$ hours of news broadcast from each of the $5$ television news channels -- \emph{CNN-IBN}, \emph{TIMES NOW}, \emph{NDTV} $24 \times 7$, \emph{BBC WORLD} and \emph{CNN}. Indian News channels are specifically selected as they do not follow any particular news presentation format( e.g. no blank frame before or after commercials), closed caption text is not provided, have large variability and dynamic nature presenting a challenging machine learning problem.
Recording is performed at $25$ FPS, in $720 \times 576$ PAL-B format with audio sampling rate of $44.1$ kHz in chunks of $1$ hour videos using a satellite receiver and audio-video capture card over a span of $1$ week and are stored in MPEG4 format. 3 Indian channels are recorded concurrently while 2 International are recorded simultaneously. Video shots are used as unit for generating instances. Broadcast News videos are segmented into video shots using RGB Colour Histogram matching Between consecutive video frames. From each shot $11$ audio visual features described in Section~\ref{sec:comFeat} are extracted. This TV news commercials dataset is publicly available\footnote{Available from UCIF ML Repository \tiny{\url{http://archive.ics.uci.edu/ml/datasets/TV+News+Channel+Commercial+Detection+Dataset}} }. The channel wise distribution of shots is tabulated in Table~\ref{tab:dataset}.

\begin{table}[!t]
\caption{\small{Channel wise distribution of shots in TV News Commercials Dataset. Commercials shots( positives) dominates the dataset}}
\vspace{-1em}
\begin{center}
{\tabulinesep=1.2mm
\begin{tabu}{cccc}
\hline
Channel & Number of Shots & Positives & Negatives \\ \hline
TIMES NOW & 39252 & 25147 & 14105 \\ 
NDTV & 17052 & 12564 & 4487 \\ 
CNNIBN & 33117 & 21693 & 11424 \\
BBC & 17720 & 8416 & 9304 \\ 
CNN & 22535 & 14401 & 8134 \\
Total Shots & 129676 & 82221 & 47454 \\ \hline
\end{tabu}}
\end{center}
\label{tab:dataset}
\vspace{-3em}
\end{table}

Next section presents the experiments and implementation details. 
\section{Experimentation}
\label{sec:expt}

Discriminating hyperplane of a SVM based classifier (Equation~\ref{eq:swk}) directly depends on training instances. Classifiers trained on imbalanced datasets ( in terms of number of positive and negative instances and variability) will either lead to  biased classification ( biased towards majority class) or over fitting ( over-fitting on minority class ). Biased classifier is the consequence of comparatively large number of support vectors from  majority class( due to inter class imbalance) and high intraclass imbalance in minority class. Whereas, over fitting is the result of interclass imbalance and high intraclass imbalance of minority class. To avoid ill effects of interclass and intraclass imbalance of the training data we have used cluster based over sampling (CBO, henceforth) scheme proposed in \cite{Jo2004}.

For each dataset we have several Kernel-Feature Combinations. We have used Linear( L-K) , RBF( R-K) and $\chi^{2}$ (X-K) kernels with first stage classifiers and RBF kernel for Regression. Though the $\chi^{2}$ kernel is used only for distribution like features. On each Feature-Kernel type combination a separate classifier and regressor  are trained. The trained classifier is evaluated on training set to identify the ``regions of success'' in  feature space. These regions of success of each classifiers are modeled by SVR.

The results are reported by dividing the available datasets into testing ($40\%$) and training sets( $60\%$) with stratification. Only the training set is balanced using CBO while the testing set is kept untouched. We have also reported the results on five other methods other than the proposed method (S-MKL) -- Concatenation(Concat) of all features (early fusion) with single SVM , F-measure Weighted ensemble (F-EC) of classifiers trained on each Feature Kernel combination; optimization based MKL (SG-MKL) \cite{shogun:2010}, data dependent Localized MKL (L-MKL) \cite{alpaydin:2008} and F-Measure weighted multiple kernel learning (F-MKL) \cite{tanabe:2008}.In case of SG-MKL and L-MKL, same number of kernels as in S-MKL are used.
To establish the  unbiased behavior of the classifiers we have reported the results on both positive as well as negative class on testing set. Complete experiment is repeated $10$ times to establish the consistency in the reported values. Fraction of training vectors which are selected as support vectors are also reported. For our proposed method total number of support vectors of final classifier are reported. Moreover the generalization capabilities of different methods are tested by varying the training dataset size from $10\%$ to $90\%$ (in steps of $10\%$) of the total data set.

We have implemented feature extraction codes in C++ using OpenCV\cite{opencv} library for visual features and LibSND\cite{LibSND} library for audio features. For support vector based classification (C-SVC), L-MKL and regression ($\epsilon$-SVR), we have used the publicly available LibSVM library \cite{libSVM} and for SG-MKL we have used Shogun Library \cite{shogun:2010}. All the datasets are scaled to range $[0,1]$ before training and testing. The hyper parameters for C-SVC and $\epsilon$-SVR ($C$, $\epsilon$ and $\gamma$ for RBF kernel) are obtained by a grid search using available functionalities of libSVM with the objective of maximizing the balanced accuracy and minimizing the MSE. Hyper parameters for L-MKL are also grid searched and the best parameters are chosen. Use of balanced accuracy instead of accuracy of a single class ensures the unbiasedness of the classifier.

\begin{figure*}[htbp]
\centerline{\includegraphics[width=1\columnwidth]{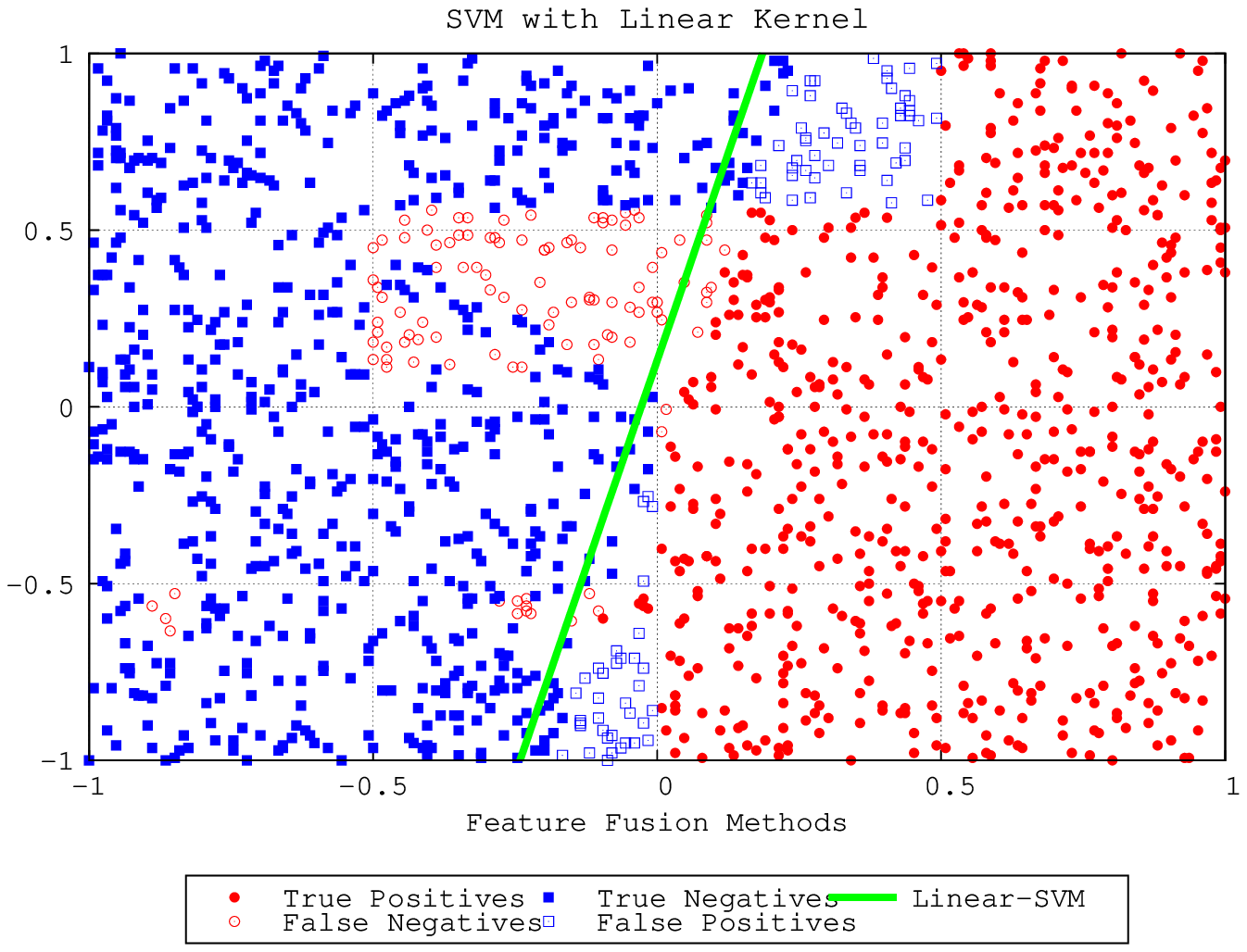}
\includegraphics[width=1\columnwidth]{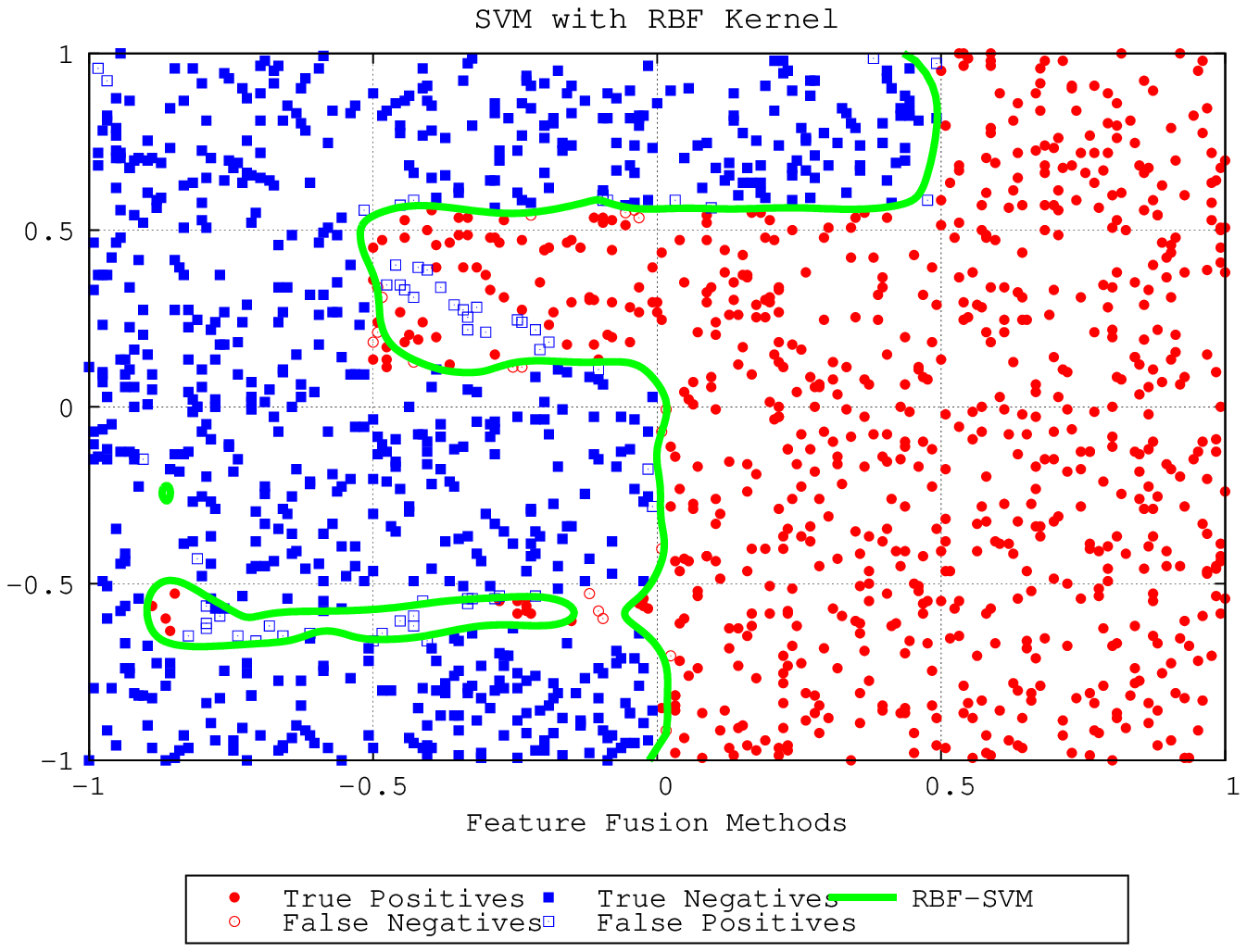}}
\centerline{(a)\hspace{\columnwidth}(b)}
\centerline{\includegraphics[width=1\columnwidth]{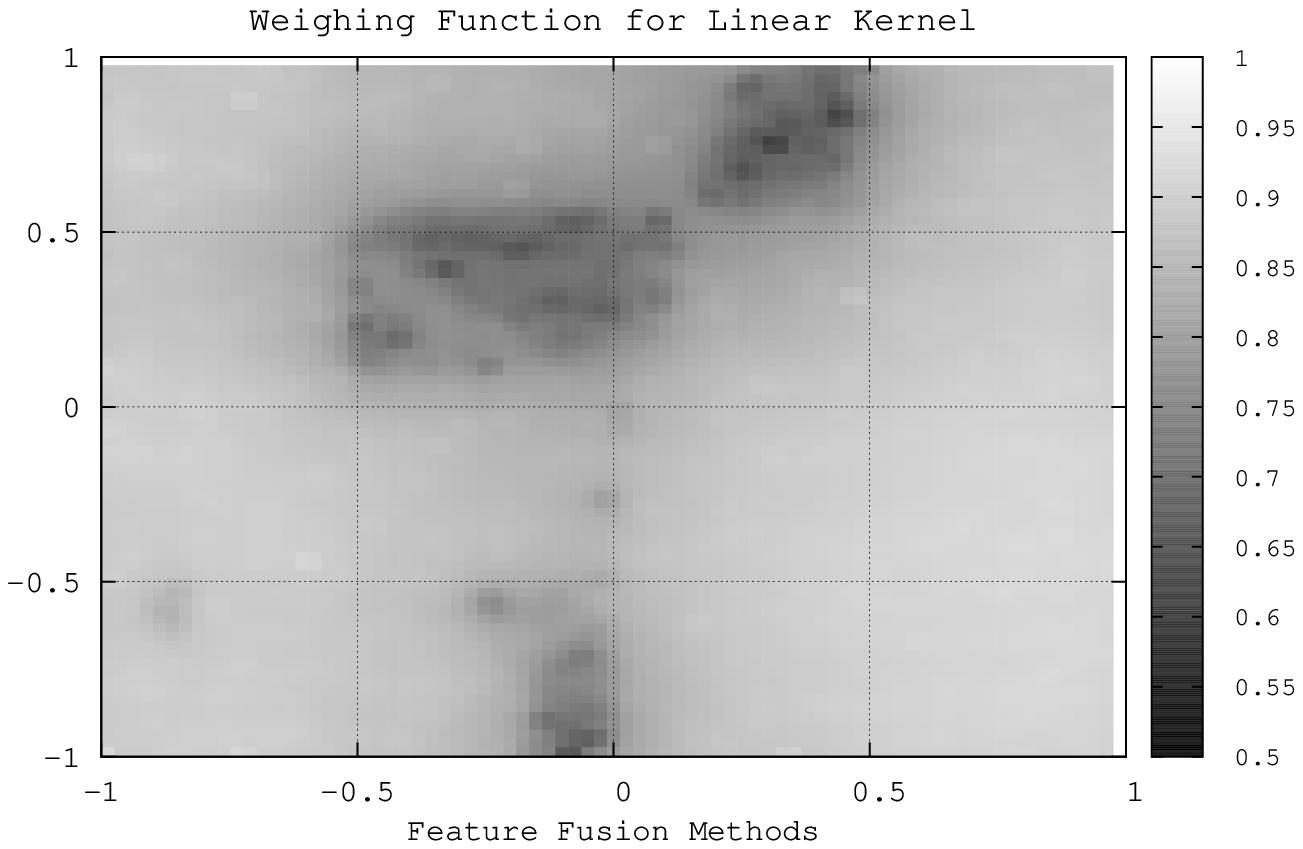}
\includegraphics[width=1\columnwidth]{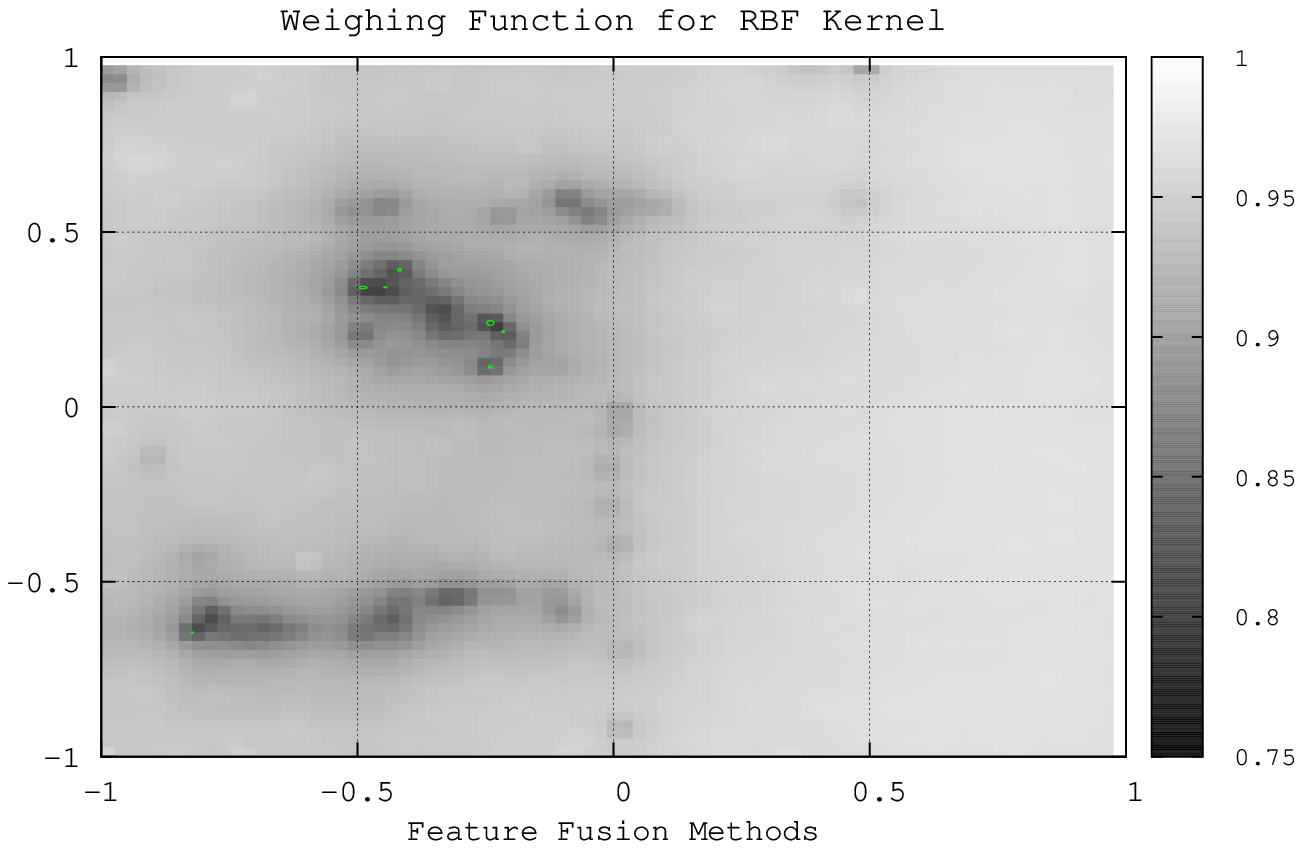}}
\centerline{(c)\hspace{\columnwidth}(d)}
\centerline{\includegraphics[width=1\columnwidth]{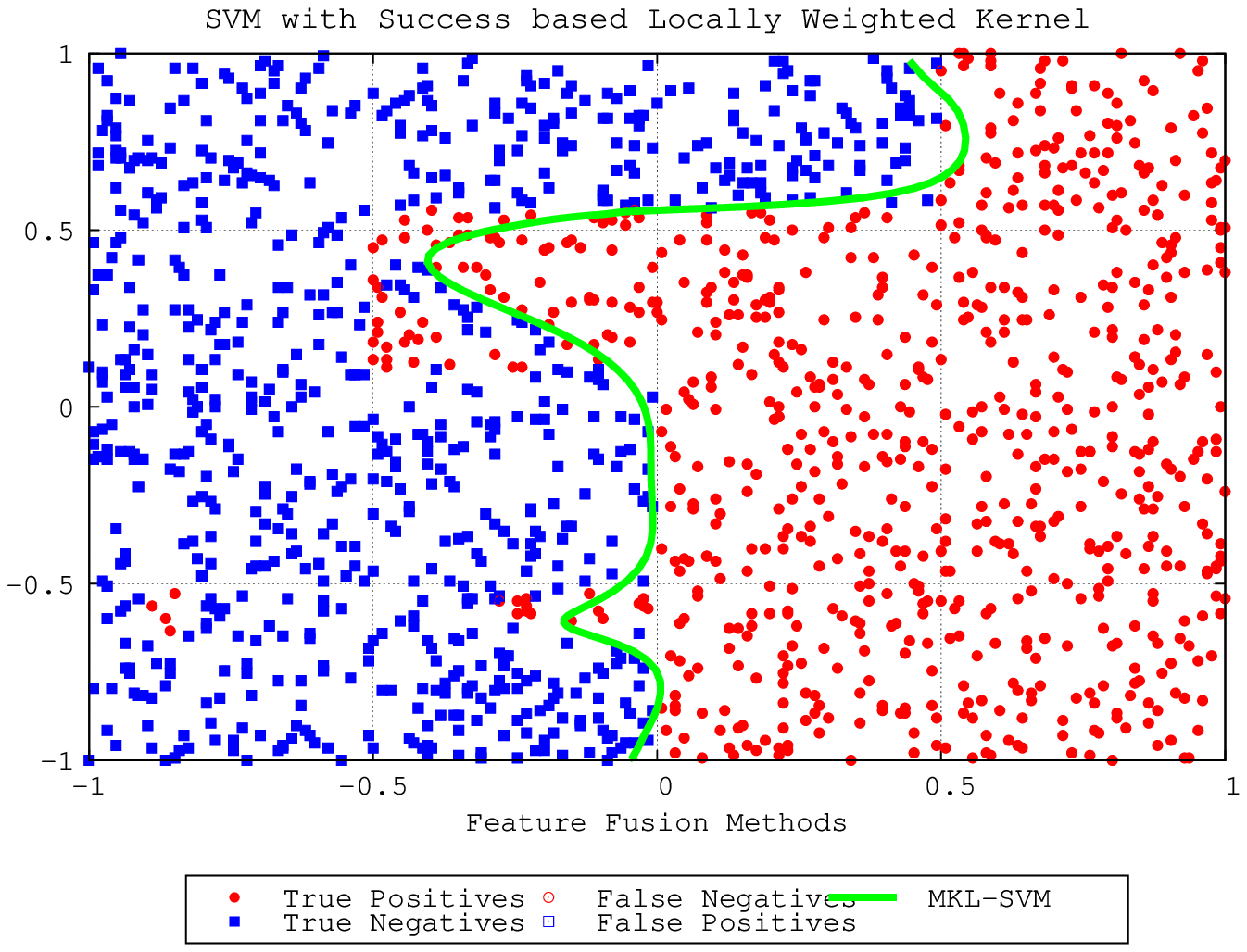}}
\centerline{(e)}
\caption{\small{Illustration of Our proposed Success based locally weighted MKL on toy dataset: For toy dataset we have two kernel functions ( Linear and RBF) each operating on a 2D feature. Hyperplanes obtained by training the classifiers with linear and RBF kernel functions are shown in Figures ~\ref{fig:toydata}(a) and ~\ref{fig:toydata}(b) respectively. Correctly classified data instances are represented by solid shapes while empty shapes represents the misclassified points. Next we try to estimate the success prediction surface using training data with objective of predicting high value ($1$) for successfully classified instances and low value($0$) for the  misclassified data instances. Figure ~\ref{fig:toydata}(c) and ~\ref{fig:toydata}(b) shows the success prediction function estimated by SVR for linear and RBF kernel functions respectively. The final discriminating hyperplane obtained using our proposed method is shown in Figure~\ref{fig:toydata}(e). This figure is best viewed in colour}}
\label{fig:toydata}
\end{figure*}


\begin{table*}[!t]
\small
\caption{\small{ Shot wise performance analysis of different methods on TV news commercials dataset. Our Proposed method S-MKL outperforms all baseline methods. standard deviations in result after repeating the experimentation are indicated in parenthesis }}
\begin{center}
{\tabulinesep=1.2mm
\begin{tabu}{ccccccccc}
\hline
Methods $\downarrow$ & \multicolumn{ 3}{c}{Commercials(Positive)} & & \multicolumn{ 3}{c}{Non Commercials(Negative)} & Support \\ \cline{ 2- 4} \cline{6-8}
& Precision &Recall& F-Measure && Precision& Recall & F-Measure &  Vectors \\ \hline
CONCAT & 0.94(0.0109) & 0.90(0.005) & 0.92(0.0001) && 0.93(0.0123) & 0.89(0.0124) & 0.91(0.0001) & 0.51(0.031) \\ 
F-EC & 0.91(0.0260) & 0.95(0.0126) & 0.93(0.0011) & & 0.92(0.0172) & 0.90(0.0246) & 0.91(0.001) & 0.47(0.0761) \\ 
SGMKL & 0.96(0.0159) & 0.83(0.12) & 0.89(0.009) & & 0.88(0.0221) & 0.94(0.0058) & 0.91(0.0001) & 0.57(0.0562) \\ 
L-MKL & 0.97(0.0013) & 0.95(0.0025) & 0.96(0.0001) & & 0.5(0.451) & 0.81(0.0055) & 0.62(0.0014) & 0.68(0.0902) \\ 
F-MKL & 0.94(0.0610) & 0.92(0.0038) & 0.93(0.0004) && 0.97(0.0049) & 0.95(0.0438) & 0.96(0.0004) & 0.6(0.0834) \\ 
S-MKL &  \bf0.99(0.0001) & \bf 0.99(0.0021) &  \bf0.99(0.0001) & &  \bf1(0.0003) &  \bf0.98(0.0039) &  \bf0.99(0.0002) &  \bf0.32(0.0057) \\ \hline
\end{tabu}}
\end{center}
\label{tab:commercialPerform}
\vspace{-2em}
\end{table*}
\begin{table*}[htbp]
\caption{\small{ Performance analysis of different methods on TV news commercials dataset based on duration of shots. Our Proposed method S-MKL outperforms all baseline methods. L-MKL has highest training time though it is fastest during training. Our method has moderate training and testing time}}
\begin{center}
{\tabulinesep=1.2mm
\begin{tabu}{cccccccccc}
\hline
Methods $\downarrow$ & \multicolumn{ 3}{c}{Commercials(Positive)} & & \multicolumn{ 3}{c}{Non Commercials(Negative)} & Avg. Training &Avg. Testing  \\ \cline{ 2- 4}\cline{6-8}
& Precision&Recall & F-Measure & & Precision & Recall &F-Measure &  time (Hr)& time (msec) \\ \hline
Concat & 0.822 & 0.853 & 0.837 && 0.89 & 0.881 & 0.885 & \bf18.4 & 19 \\ 
F-EC & 0.851 & 0.83 & 0.84 & & 0.88 & 0.867 & 0.873 & 38.6 & \bf14 \\ 
SGMKL & 0.819 & 0.835 & 0.827 && 0.856 & 0.864 & 0.86 & 67.8 & 45 \\ 
L-L-MKL & 0.834 & 0.848 & 0.841 && 0.623 & 0.721 & 0.668 & 75.1 & \bf14 \\
F-MKL & 0.918 & 0.893 & 0.905 & &0.908 & 0.91 & 0.909 & 43.1 & 28 \\ 
S-MKL & \bf0.987 & \bf0.989 &\bf 0.988 & &\bf0.996 & \bf0.986 & \bf0.991 & 48.6 & 27 \\ \hline
\end{tabu}}
\end{center}
\label{tab:timeAnalysis}
\vspace{-2em}
\end{table*}
\subsection{Toy Data Set}
\label{subsec:toydataset} 
%
%
%
Our proposed scheme is illustrated using a 2D synthetically created toy dataset consisting of $1500$ instances ( $750$ instances of each class)(Figure~\ref{fig:toydata}). We assume both dimension to be part of a feature and use linear and RBF kernel with it, hence for toy dataset  we have two kernel functions. Hyperplanes obtained by training the classifiers with linear and RBF kernel functions are shown in Figures ~\ref{fig:toydata}(a) and ~\ref{fig:toydata}(b) respectively. Correctly classified data instances are represented by solid shapes while empty shapes represents the misclassified points. Misclassification in case of linear kernel is in the regions of feature space where a nonlinear hyperplane is required for separating the data. While RBF kernel fails in the regions of feature space having significant data overlap due to over fitting.
Next we try to estimate the success prediction function using the training data with objective of predicting high values ($1$) for successfully classified instances and low values($0$) for the misclassified data instances. Success prediction function estimated by SVR are shown in Figure ~\ref{fig:toydata}(c) and ~\ref{fig:toydata}(d). SVM using weighted linear combination of linear and RBF kernels with weights decided by the success prediction function should combine the best of both individual classifiers. The final discriminating hyperplane obtained using our proposed method is shown in Figure~\ref{fig:toydata}(e). From ~\ref{fig:toydata}(c) and ~\ref{fig:toydata}(d) it is clear that linear kernel is selected where there is possibility of over-fitting by RBF kernel while RBF kernel is selected when a non-linear separating hyperplane is required, which is evident from Figure~\ref{fig:toydata}(e).

\subsection{TV News Commercials Dataset}
\label{subsec:toydataset}

We have benchmarked our method our own TV News commercials dataset (Section~\ref{sec:dataset}) (publically available) as no other commercial detection dataset is publically available. Particulars of the datasets are tabulated in Table~\ref{tab:dataset}. Each instance  of TV News Commercials dataset consists of $11$ audio visual features having $4117$ dimensions. We have used  $11$ linear kernels ( One kernel for each feature ), $11$ RBF kernels( One kernel for each feature ) and $4$ $\chi^{2}$ kernels ( one each with text distribution, motion distribution, frame difference distribution and audio bag of words). Hence for commercial detection we use SVM with a combination of $26$ kernel functions. The performance of classifiers trained with individual kernel functions is presented in Figure~\ref{fig:featAnal}. Out of these classifiers, Text distribution and MFCC bag of audio words with $\chi^{2}$ and RBF kernel classifier are turned out to be best performing classifiers.
The classification results of different methods on TV News Commercials dataset are tabulated in Table~\ref{tab:commercialPerform}. Our proposed method outperforms all other baseline methods.


The performance tabulated in Table~\ref{tab:commercialPerform} is not a fair evaluation from the point of view of TV News commercial detection due to the fact that even though TV commercials have more number of shots than non commercials, duration of commercial shots is much smaller compared to the duration of non commercial shots. Hence the cost of misclassifying a non-commercial shot is  more than the cost of misclassifying a commercial shot. Thus we present broadcast time wise analysis of commercial detection in Table~\ref{tab:timeAnalysis}. In terms of broadcast time, all baseline algorithms lags by a significant margin than our proposed method. In Table~\ref{tab:timeAnalysis} the average training and testing time for different methods are also reported. During training L-MKL turns out to be the most expansive approach followed by SG-MKL and S-MKL. L-MKL assumes the linear separability between the regions of influence of each kernel and locates these regions by gradient descent. The assumption on linear separability is not practical, hence convergence takes the extended time.  Simple concatenation with single SVM as expected took least training time.
L-MKL and F-EC are fastest during testing due to reduced number of kernel calculations during testing. In L-MKL kernel computations are reduced as theoretically for every support vector only single kernel is active. In F-EC kernel computations are reduced as individual classifiers are trained on features with small dimension.  Our proposed method stands third in terms of training and testing time. Comparatively long time taken by our proposed method may be attributed to number of classifiers and regressors involved. But longer training and testing time is justified by the gain in performance.

The results of experiments by varying the training data size are tabulated in Table~\ref{tab:commercialGen} and visualized in figure~\ref{fig:commercialGen}.  Intraclass variability preserved by CBO based data balancing is reflected in the consistent performance of classifiers even after varying training data size. All the methods except our proposed method(S-MKL) and localized MKL(LMKL) exhibit the consistent performance over varying training data sizes.  S-MKL becomes consistent after sufficient data is available for training. While L-MKL shows consistence in F-measure for positive class only resulting in highly biased classifier. Poor performance of our proposed method on smaller datasets may be attributed to the imperfect learning of success prediction function due to in sufficient data (SVRs have large MSE for small training data sizes).

\begin{table}[!h]
\vspace{-1em}
\scriptsize
\caption{Generalization Performance of different methods on Commercial Dataset: \small{Table shows the generalization performance of different methods on TV News Commercials Dataset. All the methods except our proposed method(S-MKL) and localized MKL(LMKL)  exhibit the consistent performance over varying training data sizes. S-MKL becomes consistent after sufficient data is available for training. While L-MKL shows consistence in F-measure for positive class only resulting in highly biased classifier.}}
\vspace{-1em}
\begin{center}
{\tabulinesep=1.5mm
	
	\begin{tabu}{ccccccccccccc}
\hline
     \\
\multirow{-2}{*}{\rotatebox[origin=c]{90}{Data}}&\multirow{-2}{*}{\rotatebox[origin=c]{90}{($\%$)}}& \multirow{-2}{*}{10} & \multirow{-2}{*}{20} & \multirow{-2}{*}{30} & \multirow{-2}{*}{40} & \multirow{-2}{*}{50} & \multirow{-2}{*}{60 }& \multirow{-2}{*}{70} & \multirow{-2}{*}{80} &\multirow{-2}{*}{90} \\\hline

& \rotatebox[origin=c]{90}{F+}  & 0.89 & 0.88 & 0.91 & 0.92 & 0.91 & 0.92 & 0.92 & 0.9 & 0.92 \\ 
\multirow{-2}{*}{\rotatebox[origin=c]{90}{Concat}} & \rotatebox[origin=c]{90}{F-}  & 0.9 & 0.91 & 0.91 & 0.91 & 0.92 & 0.91 & 0.9 & 0.91 & 0.92 \\ \hline
& \rotatebox[origin=c]{90}{F+}  & 0.88 & 0.85 & 0.92 & 0.93 & 0.92 & 0.93 & 0.91 & 0.93 & 0.92 \\ 
\multirow{-2}{*}{\rotatebox[origin=c]{90}{F-EC}} & \rotatebox[origin=c]{90}{F-} & 0.89 & 0.88 & 0.9 & 0.91 & 0.9 & 0.91 & 0.9 & 0.9 & 0.9 \\ \hline
& \rotatebox[origin=c]{90}{F+}  & 0.73 & 0.8 & 0.86 & 0.88 & 0.88 & 0.89 & 0.89 & 0.88 & 0.88 \\ 
\multirow{-2}{*}{\rotatebox[origin=c]{90}{SG-MKL}} & \rotatebox[origin=c]{90}{F-}  & 0.69 & 0.79 & 0.8 & 0.86 & 0.75 & 0.91 & 0.89 & 0.9 & 0.91 \\ \hline
& \rotatebox[origin=c]{90}{F+}  & 0.79 & 0.76 & 0.86 & 0.88 & 0.89 & 0.96 & 0.95 & 0.94 & 0.96 \\ 
 \multirow{-2}{*}{\rotatebox[origin=c]{90}{L-MKL}} & \rotatebox[origin=c]{90}{F-}  & 0.72 & 0.74 & 0.73 & 0.78 & 0.7 & 0.62 & 0.65 & 0.69 & 0.7 \\ \hline
& \rotatebox[origin=c]{90}{F+}  & 0.87 & 0.88 & 0.86 & 0.9 & 0.92 & 0.93 & 0.91 & 0.92 & 0.89 \\ 
\multirow{-2}{*}{\rotatebox[origin=c]{90}{F-MKL}} & \rotatebox[origin=c]{90}{F-}  & 0.89 & 0.94 & 0.94 & 0.94 & 0.93 & 0.96 & 0.95 & 0.94 & 0.93 \\ \hline
&\rotatebox[origin=c]{90}{F+}  & 0.66 & 0.78 & 0.83 & 0.89 & 0.92 & 0.99 & 0.99 & 1 & 0.99 \\ 
\multirow{-2}{*}{\rotatebox[origin=c]{90}{S-MKL}} & \rotatebox[origin=c]{90}{F-} & 0.59 & 0.81 & 0.86 & 0.93 & 0.95 & 0.99 & 0.98 & 0.99 & 0.98 \\ \hline
\end{tabu}
}
\end{center}
\label{tab:commercialGen}
\end{table}
\begin{figure*}[!t]
\centerline{\includegraphics[width=0.5\textwidth]{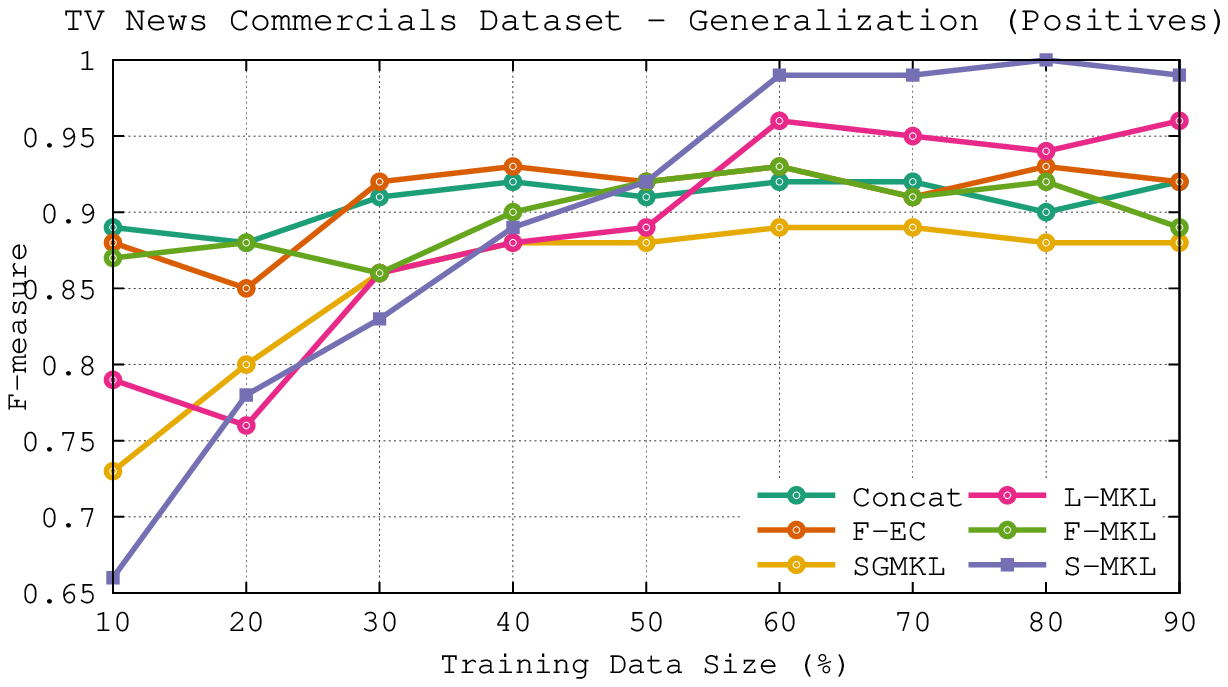}
\includegraphics[width=0.5\textwidth]{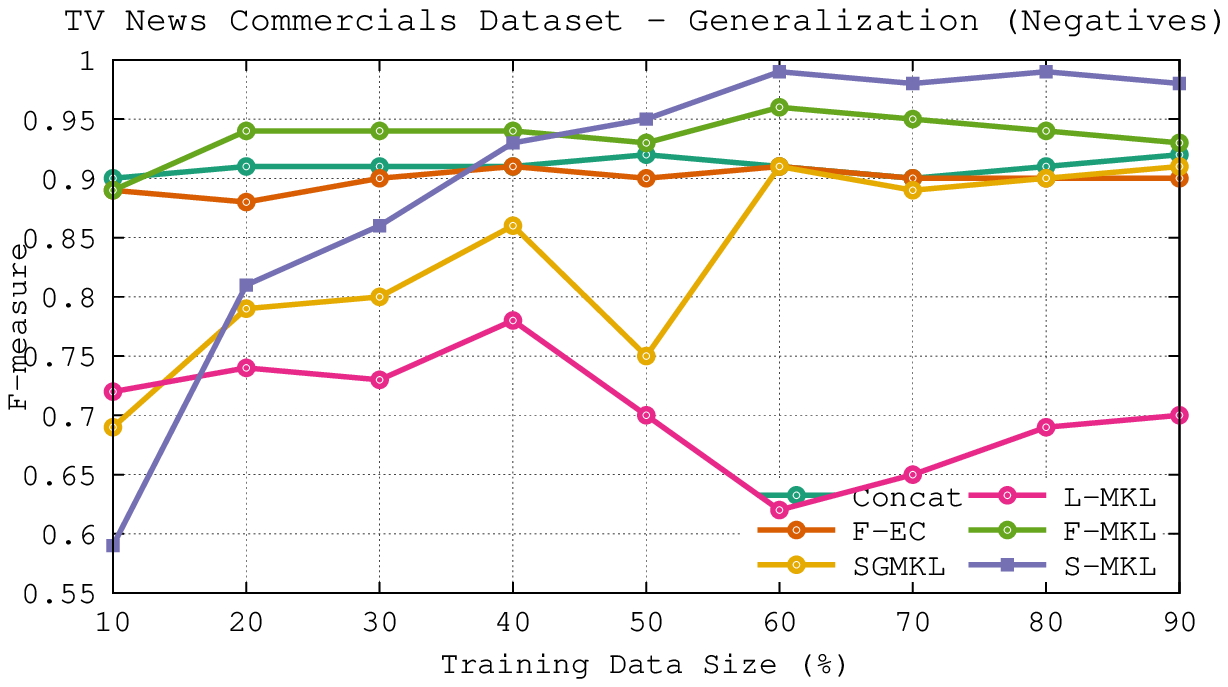}}
\vspace{-1em}
\caption{\small{Visualization of generalization performance data presented in Table~\ref{tab:commercialGen}. The variations of f-measures for (a) positive and (b) negative categories are presented with respect to changing training set size.}}
\label{fig:commercialGen}
\vspace{-1em}
\end{figure*}

\subsection{Benchmark Datasets}
\label{subsec:otherdata}

Most of the previous works on TV News Commercials detection including current state of art work by Liu et.al. \cite{evat} experimented and benchmarked the results on their own datasets. These datasets are not available in public domain. Moreover most of the works have used channel specific or country specific heuristics for extracting the features( e.g. presence of blank frame before commercials), designing classifiers and for post processing which are not true in general. Hence it is very difficult to benchmark the performance of our proposed method for commercial detection with  current state of art.
 To demonstrate the performance of proposed S-MKL, we have benched marked our results on $8$ publicly available datasets. Out of these $8$ datasets S-MKL outperforms other baseline methods on $6$ out of $8$ datasets. The results are tabulated in table~\ref{tab:performAll} and particulars of datasets are given in Appendix. Moreover in our method results for positive and negative classes are more or less balanced. This might be due to the fact that the success based weighing functions were learned for successful prediction of both the positive and negative categories. Performance of our method suffers drastically on smaller datasets. One of the possible reasons for failure of our method on two datasets is due to smaller dataset size which hampers the regression model for estimating the success prediction function ( trained SVRs had high MSEs on small datasets ). On smaller datasets L-MKL and MKL performs quite well but their performance decreases sharply as the data size increases. This may be due to violations of assumptions for these methods on larger datasets. In almost all the cases our method produces balanced output while other methods are showing strong bias towards either of the classes. Also, in our method the number of support vectors are significantly less compared to other methods.

\subsection{Discussions}
\label{subsec:Dis}

In our proposed method we have used a weighted linear combination of the kernels for training SVM instead of predefined single kernel.  The weights for the kernels are adaptively estimated from the data. S-MKL out of existing methods most closely relates to F-MKL \cite{tanabe:2008} and L-MKL\cite{alpaydin:2008}. In F-MKL kernels have fixed weights throughout the feature space. While L-MKL uses weights which are locally varying. In \cite{alpaydin:2008} it was reported for some datasets that L-MKL performs better than F-MKL. But in our experimentation we have observed that in most cases F-MKL have outperformed L-MKL. This reduction in performance in L-MKL  may be justified by the fact that L-MKL assumes linear separability of ``regions of use of kernels" \cite{alpaydin:2008} and hence, theoretically only one kernel should be active for any given data instance. However, this assumption fails in most practical cases unless an arbitrarily large number of kernels are used hence leading to misclassification.
On the other hand  our proposed S-MKL does not make any assumptions on linear separability of regions of use of kernels hence beats L-MKL.

SG-MKL and F-MKL both have fixed set of weights for the entire feature space  but we have observed that for large datasets SG-MKL tends to over fit (Evident from the bias towards either of the classes). Hence SG-MKL and F-MKL have comparable performance on smaller datasets but success based F-MKL outperforms on larger datasets.

Different kernel functions may provide different views of the data but these views may represent the redundant information.  The redundancy in information suppress  the complementary views and hence  redundant information in favor of misclassification hampers the performance of an ensemble classifier. It may be noted that out of baseline methods F-MKL( intermediate fusion) and F-EC(late fusion) use same weighing function (F-measure) to select among different classifiers but F-MKL has comparatively unbiased and better performance.  Hence it may be concluded that  F-MKL takes care of redundant information to an extent. Moreover in our proposed scheme  only successful kernel functions will have sufficient weight to contribute to the final decision. This weighing scheme ensures that even if the kernels have redundant information, it is in favor of correct decision. Moreover, success weighing also ensures that fewer number of correct classifiers won't be dominated by larger number of failed ones. We have observed that in most cases, even with a single successful kernel function, S-MKL could predicted correct labels. This indicates that in our scheme redundancy most of the time is in the favor of correct classification and not otherwise.    

\begin{table*}[t]
\scriptsize
\centering
\caption{ \small{ Table shows the performance analysis of all methods on benchmark datasets. Performance analysis shows the F measure of positive (Commercials) (F+) and negative ( non commercials )(F-) class along with fraction of data points which are chosen as support vectors(SV) on all datasets. It is clear from the table that our method ( S-MKL) out performs all other methods except on two datasets( Liver Disorder and Ionosphere) which are very small in size. The figures in bracket are the standard  deviations in values when experimentation is repeated.}  }
{\tabulinesep=1.5mm
	\begin{tabu}{ccccccccccc}
\hline
&&Liver  & Ionosphere & Breast & Diabetes & German & Mushrooms & COD & Adult \\  
\multicolumn{2}{c}{\multirow{-2}{*}{\rotatebox[origin=c]{90}{\parbox{0.3cm}{Data sets}} }}& Disorder &  & Cancer &  & Numeric &  & RNA &    \\ \hline
& F+ & 0.52  (0.04 ) & 0.68 (0.0053) & 0.69 (0.0236) & 0.75 (0.021) & 0.67 (0.0010) & 0.49 (0.0076) & 0.76 (0.004) & 0.28 (0.02)  \\ 
 \multirow{-2}{*}{\rotatebox[origin=c]{90}{Concat}} & F- & 0.71 (0.02) & 0.77 (0.034) & 0.87 (0.0113) & 0.49 (0.0038) & 0.43 (0.0008) & 0.56 (0.081) & 0.64 (0.0123) & 0.82 (0.14) \\ \hline
& F+ & 0.34 (0.1) & 0.59 (0.0189) & 0.71 (0.0613) & 0.78 (0.019) & 0.65 (0.0923) & 0.3 (0.0046) & 0.79 (0.024) & 0.2 (0.102)  \\
\multirow{-2}{*}{\rotatebox[origin=c]{90}{F-EC}} & F- &\bf0.81 (0.012) & 0.62 (0.0124) & 0.76 (0.0904) & 0.34 (0.0021) & 0.63 (0.021) & 0.79 (0.023) & 0.71 (0.0011) & 0.79 (0.012)  \\ \hline
& F+ & 0.62 (0.0053) & 0.72 (0.1041) & 0.74 (0.0019) & \bf0.81 (0.0051) & 0.71 (0.052) & 0.52 (0.046) & 0.62 (0.0001) & 0.58 (0.0018)  \\ 
 \multirow{-2}{*}{\rotatebox[origin=c]{90}{SGMKL}} & F- & 0.76 (0.009) & 0.79 (0.012) & 0.69 (0.0089) & 0.58 (0.0019) & 0.69 (0.01) & 0.69 (0.0234) & 0.54 (0.0074) & 0.49 (0.0001)  \\ \hline
& F+ & \bf0.63 (0.5) & \bf0.94 (0.0001) & 0.69 (0.0078) & 0.72 (0.0701) & \bf0.79 (0.0081) & 0.52 (0.083) & 0.4 (0.0009) & 0.6 (0.0025)  \\ 
 \multirow{-2}{*}{\rotatebox[origin=c]{90}{L-MKL}}& F- & 0.75 (0.091) & \bf0.87 (0.009) & 0.79 (0.012) & 0.69 (0.0101) &\bf 0.78 (0.0064) & 0.72 (0.0001) & 0.51 (0.0083) & 0.3 (0.5) \\ \hline
& F+ & 0.58 (0.0541) & 0.82 (0.0114) & 0.74 (0.023) & 0.71 (0.0109) & 0.71 (0.0005) & 0.73 (0.0131) & 0.79 (0.014) & 0.58 (0.019)  \\ 
 \multirow{-2}{*}{\rotatebox[origin=c]{90}{F-MKL} }& F- & 0.56 (0.0029) & 0.86 (0.0121) & 0.86 (0.008) & 0.79 (0.081) & 0.69 (0.0093) & 0.75 (0.01) & 0.82 (0.015) & 0.62 (0.0001)  \\\hline
& F+ & 0.54 (0.0874) & 0.65 (0.1534) & \bf0.89 (0.0071) & 0.79 (0.0067) & 0.71 (0.0053) & \bf0.87 (0.029) & \bf0.9 (0.0141) & \bf0.79 (0.015)  \\ 
\multirow{-2}{*}{\rotatebox[origin=c]{90}{S-MKL}} & F- & 0.51 (0.0809) & 0.69 (0.0729) & \bf0.94 (0.130) & \bf0.82 (0.0091) & 0.76 (0.0054) & \bf0.83 (0.059) & \bf0.89 (0.0157) & \bf0.84 (0.010)  \\ \hline
\end{tabu}}
\label{tab:performAll}
\end{table*}
\section{Conclusion}
\label{sec:conc}

We have proposed a ``Success based Local Weighing'' scheme for the selection of kernel functions in the context of commercial detection in news broadcast videos. The video shots are characterized by $11$ different (existing) audio-visual features like shot length, motion and scene text distribution, ZCR, STE, spectral features, fundamental frequency and MFCC Bag of Audio Words. We have trained SVM based classifiers with linear and RBF kernel for all the features and $\chi^{2}$ kernels (for distribution like features only) resulting in a total of $26$ feature classifier combinations. Our first proposition involves using a weighted linear combination of kernels instead of single kernel in SVM where the weighing functions are estimated (using support vector regression with RBF kernel) from the zones of success of the classifiers trained with individual kernels. Success prediction functions are designed to have values closer to $1.0$ where the corresponding kernel functions had success in the training data set and $0.0$ otherwise. Our proposed approach outperformed all baseline methods. We have created a TV News commercial dataset of $150$ hours from $5$ different channels which will be made available publically. We have verified the performance improvements of the proposed classifier on $8$ standard data sets along with our own TV News Commercials dataset

In the present work, we have proposed a single stage weight prediction algorithm from multiple kernel combination. However, we have not experimented with the possibilities of kernel combinations in the support vector regression stage and have only used the RBF kernel. We believe that the simultaneous estimation of weighing functions for kernel combinations in both classifier and regressors will require a reformulation of the problem involving stages of iterative optimization. Also, in this work, we have only contributed in the classifier stage while using existing features. This work can be extended further to include text/audio content and style as features whose combination with the proposed classifier will definitely lead to better performances.

\scriptsize{
	\bibliographystyle{IEEEbib}
\bibliography{ACM_MM_arxiv}
}

\appendix[Supplementary Material]
\label{app:supMat}

In this report, we present the supplementary material for our paper on ``TV News Commercials Detection using Success based Locally Weighted Kernel Combination''. We have experimented with $8$ different standard datasets viz. \emph{Liver Disorder}, \emph{Ionosphere}, \emph{Breast Cancer}, \emph{Diabetes}, \emph{German Numeric}, \emph{Mushrooms}, \emph{COD-RNA}, \emph{Adult} along with our own \emph{TV News Commercials} dataset. The comparative results of performance analysis are presented using Precision, Recall and F-Measure obtained on $9$ different datasets for $5$ different algorithms. Due to space constraint, it was not possible to present the detailed experimental results in the limited space of the main paper. Here, we have presented the results for our  proposed algorithm \emph{S-MKL}) along with the $5$ baseline approaches viz-- \emph{CONCAT}, \emph{F-EC} \cite{Rokach2010}, \emph{SGMKL}\cite{shogun:2010}, \emph{L-MKL}\cite{alpaydin:2008} and \emph{F-MKL}\cite{tanabe:2008}.

The experimental results on the $9$ datasets are presented in Sub-Sections~\ref{subsec:liver} to \ref{subsec:commercial}.For each data set, we have reported the following sets of results.

\begin{itemize}

\item \textbf{(a)} Tabulation and Visualization of precision, recall and f-measures for both positive and negative category using SVMs learned with different feature-kernel combinations. From the given dataset, $60\%$ of the labeled data are randomly drawn to form the training dataset and the learned classifier is tested over the remaining $40\%$ samples. This experiment is repeated $10$ times and the average performance measures are reported to indicate the success rates of each feature-kernel combination.

\item \textbf{(b)} Tabulation and Visualization of the generalization performance of $7$ different algorithms. The size of training set is varied from $10\%$ to $90\%$ (in steps of $10\%$) of the given dataset size. For each training data set size, the experiment is repeated $10$ times and the average F-measures obtained from the corresponding test data set for both positive and negative category are reported.

\item \textbf{(c)} Tabulation and Visualization of the comparative performance analysis of the $7$ different classification approaches. Classifiers for each method are learned from $60\%$ (training data set) of the given dataset and are tested on the remaining $40\%$ of samples (test data set). This experiment is repeated $10$ times. We have reported the average and standard deviation of the performance measures i.e. precision, recall and f-measure. We have also reported the fraction of the data set size used by the algorithm as support vectors.

\end{itemize}

\begin{sidewaystable*}[!h]
\small
\centering
\caption{ Table shows the dataset particulars and performance analysis of all methods on different datasets. Total Number of Kernel-Feature combinations trained on each dataset and their break up is shown in the first half of the table. Performance analysis shows the F measure of positive (F+) and negative(F-) class along with fraction of data points which are chosen as support vectors(SV) on all datasets. It is clear from the table that our method ( S-MKL) out performs all other methods except on two datasets( Liver Disorder and Ionosphere) which are very small in size. The figures in bracket are the std deviation in values when experimentation is repeated.}
{\tabulinesep=1.5mm
	\begin{tabu}{cccccccccccc}
\hline
\multicolumn{ 2}{c}{Dataset} &Liver  & Ionosphere & Breast & Diabetes & German & Mushrooms & COD & Adult & Commercial \\  \cline{ 1- 2}
\multicolumn{ 2}{c}{Particulars} & Disorder &  & Cancer &  & Numeric &  & RNA &  &  \\ \hline
\multicolumn{ 2}{c}{Size}& 345 & 351 & 683 & 768 & 1000 & 8124 & 244109 & 270000 & 129676 \\
\multicolumn{ 2}{c}{Positives  (\%)}& 42.09 & 64.1 & 34.99 & 65.1 & 30 & 64.1 & 33.33 & 24.84 & 64 \\
\multicolumn{ 2}{c}{Dimension}& 6 & 34 & 10 & 8 & 24 & 21 & 8 & 123 & 11 \\
\multicolumn{ 2}{c}{Features}& 6 & 17 & 10 & 8 & 24 & 121 & 8 & 14 & 4117 \\
\multicolumn{ 2}{c}{Feature+LK} & 6 & 17 & 10 & 8 & 24 & 21 & 8 & 14 & 11 \\
\multicolumn{ 2}{c}{Feature+RK}&6 & 17 & 10 & 8 & 24 & 21 & 8 & 14 & 11 \\
\multicolumn{ 2}{c}{Feature+XK}& 0 & 0 & 0 & 0 & 0 & 0 & 0 & 0 & 4 \\
\multicolumn{ 2}{c}{Feature + Kernel}& 12 & 34 & 20 & 16 & 48 & 42 & 16 & 28 & 26 \\ \hline
\multicolumn{ 11}{c}{Performance Analysis} \\ \hline
\multicolumn{ 1}{c}{} & F+ & 0.52  (0.04 ) & 0.68 (0.0053) & 0.69 (0.0236) & 0.75 (0.021) & 0.67 (0.0010) & 0.49 (0.0076) & 0.76 (0.004) & 0.28 (0.02) & 0.92 (0.0001) \\
\multicolumn{ 1}{c}{Concat} & F- & 0.71 (0.02) & 0.77 (0.034) & 0.87 (0.0113) & 0.49 (0.0038) & 0.43 (0.0008) & 0.56 (0.081) & 0.64 (0.0123) & 0.82 (0.14) & 0.91 (0.0001) \\
\multicolumn{ 1}{c}{} & SV & 0.73 (0.0003) & 0.76 (0.0001) & 0.49 (0.1890) & 0.62 (0.0726) & 0.77 (0.0001) & 0.82 (0.0174) & 0.73 (0.2804) & 0.79 (0.0333) & 0.51 (0.031) \\ \hline
\multicolumn{ 1}{c}{} & F+ & 0.34 (0.1) & 0.59 (0.0189) & 0.71 (0.0613) & 0.78 (0.019) & 0.65 (0.0923) & 0.3 (0.0046) & 0.79 (0.024) & 0.2 (0.102) & 0.93 (0.0011) \\
\multicolumn{ 1}{c}{F-EC} & F- &\bf0.81 (0.012) & 0.62 (0.0124) & 0.76 (0.0904) & 0.34 (0.0021) & 0.63 (0.021) & 0.79 (0.023) & 0.71 (0.0011) & 0.79 (0.012) & 0.91 (0.001) \\
\multicolumn{ 1}{c}{} & SV & 0.68 (0.0013) & 0.71 (0.0001) & 0.64 (0.0135) & 0.75 (0.0923) & 0.79 (0.0001) & 0.47 (0.0341) & 0.55 (0.104) & 0.8 (0.2104) & 0.47 (0.0761) \\ \hline
\multicolumn{ 1}{c}{} & F+ & 0.42 (0.801) & 0.66 (0.0129) & 0.79 (0.0112) & 0.79 (0.07) & 0.69 (0.1101) & 0.67 (0.017) & 0.83 (0.019) & 0.5 (0.0016) & 0.97 (0.0007) \\
\multicolumn{ 1}{c}{S-EC} & F- & 0.46 (0.0080) & 0.68 (0.0411) & 0.78 (0.0871) & 0.78 (0.024) & 0.71 (0.0812) & 0.69 (0.0125) & 0.84 (0.0125) & 0.72 (0.018) & 0.98 (0.0008) \\
\multicolumn{ 1}{c}{} & SV & 0.68 (0.091) & 0.71 (0.0001) & 0.63 (0.0135) & 0.75 (0.0923) & 0.79 (0.0001) & 0.47 (0.0341) & 0.55 (0.104) & 0.8 (0.2104) & 0.47 (0.0761) \\ \hline
\multicolumn{ 1}{c}{} & F+ & 0.62 (0.0053) & 0.72 (0.1041) & 0.74 (0.0019) & \bf0.81 (0.0051) & 0.71 (0.052) & 0.52 (0.046) & 0.62 (0.0001) & 0.58 (0.0018) & 0.89 (0.009) \\
\multicolumn{ 1}{c}{SGMKL} & F- & 0.76 (0.009) & 0.79 (0.012) & 0.69 (0.0089) & 0.58 (0.0019) & 0.69 (0.01) & 0.69 (0.0234) & 0.54 (0.0074) & 0.49 (0.0001) & 0.91 (0.0001) \\
\multicolumn{ 1}{c}{} & SV &\bf 0.6 (0.081) & 0.59 (0.081) & 0.65 (0.0089) & 0.52 (0.0521) & 0.62 (0.0762) & 0.61 (0.0341) & 0.8 (0.0099) & 0.62 (0.0053) & 0.57 (0.0562) \\ \hline
\multicolumn{ 1}{c}{} & F+ & \bf0.63 (0.5) & \bf0.94 (0.0001) & 0.69 (0.0078) & 0.72 (0.0701) & \bf0.79 (0.0081) & 0.52 (0.083) & 0.4 (0.0009) & 0.6 (0.0025) & 0.96 (0.0001) \\
\multicolumn{ 1}{c}{L-MKL} & F- & 0.75 (0.091) & \bf0.87 (0.009) & 0.79 (0.012) & 0.69 (0.0101) &\bf 0.78 (0.0064) & 0.72 (0.0001) & 0.51 (0.0083) & 0.3 (0.5) & 0.62 (0.0014) \\
\multicolumn{ 1}{c}{} & SV & 0.78 (0.0921) & 0.61 (0.081) & 0.5 (0.0023) & 0.49 (0.0801) & \bf0.49 (0.0023) & \bf0.42 (0.192) & 0.7 (0.0921) & 0.56 (0.0187) & 0.68 (0.0902) \\ \hline
\multicolumn{ 1}{c}{} & F+ & 0.58 (0.0541) & 0.82 (0.0114) & 0.74 (0.023) & 0.71 (0.0109) & 0.71 (0.0005) & 0.73 (0.0131) & 0.79 (0.014) & 0.58 (0.019) & 0.93 (0.0004) \\
\multicolumn{ 1}{c}{F-MKL} & F- & 0.56 (0.0029) & 0.86 (0.0121) & 0.86 (0.008) & 0.79 (0.081) & 0.69 (0.0093) & 0.75 (0.01) & 0.82 (0.015) & 0.62 (0.0001) & 0.96 (0.0004) \\
\multicolumn{ 1}{c}{} & SV & 0.62 (0.012) & \bf0.43 (0.0801) & 0.43 (0.0081) & 0.45 (0.091) & 0.52 (0.0289) & 0.6 (0.0821) & 0.49 (0.0921) & 0.54 (0.0076) & 0.6 (0.0834) \\ \hline
\multicolumn{ 1}{c}{} & F+ & 0.54 (0.0874) & 0.65 (0.1534) & \bf0.89 (0.0071) & 0.79 (0.0067) & 0.71 (0.0053) & \bf0.87 (0.029) & \bf0.9 (0.0141) & \bf0.79 (0.015) & \bf0.99 (0.0001) \\
\multicolumn{ 1}{c}{S-MKL} & F- & 0.51 (0.0809) & 0.69 (0.0729) & \bf0.94 (0.130) & \bf0.82 (0.0091) & 0.76 (0.0054) & \bf0.83 (0.059) & \bf0.89 (0.0157) & \bf0.84 (0.010) & \bf0.99 (0.0002) \\
\multicolumn{ 1}{c}{}& SV & 0.7 (0.1098) & 0.69 (0.0009) &\bf 0.35 (0.00724 & \bf0.44 (0.01) & 0.67 (0.0431) & 0.46 (0.2130) & \bf0.29 (0.1067) & \bf0.31 (0.0025) & \bf0.32 (0.0057) \\ \hline
\end{tabu}
 }
\label{tab:performAna}
\end{sidewaystable*}

\subsection{Liver Disorder Database}
\label{subsec:liver}

\begin{sidewaysfigure*}[htbp]
\centerline{\includegraphics[width=1\textheight]{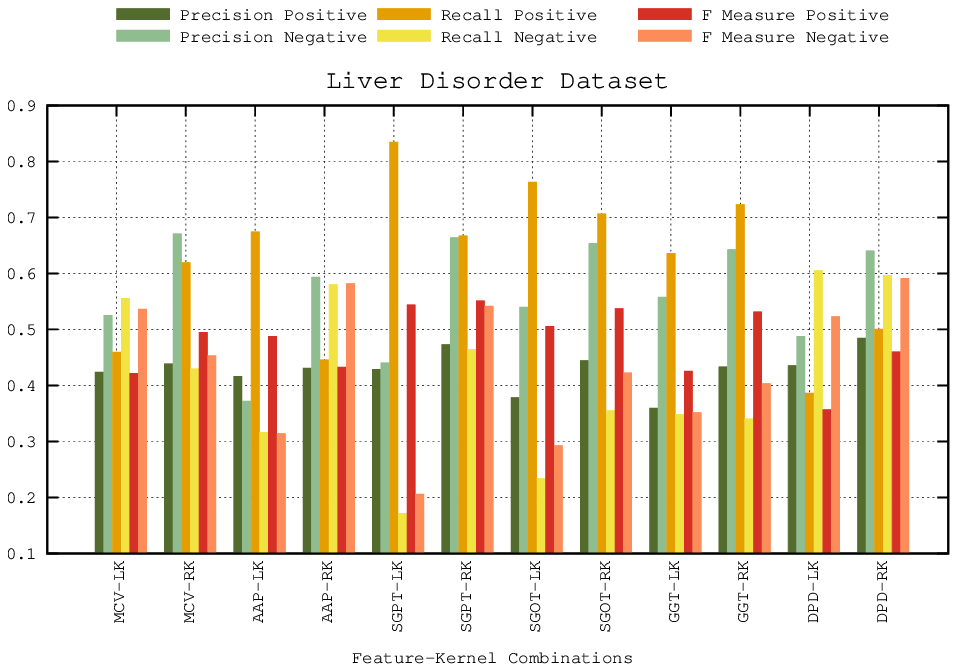}}
\caption{\small{Visualization of the performance analysis data presented in Table~\ref{tab:liverFeature}. The precision, recall and f-measures for different feature kernel combinations are shown for the Liver Disorder dataset.}}
\label{fig:liverFeature}
\end{sidewaysfigure*}

\begin{figure*}
\centerline{
\includegraphics[width=0.5\textwidth]{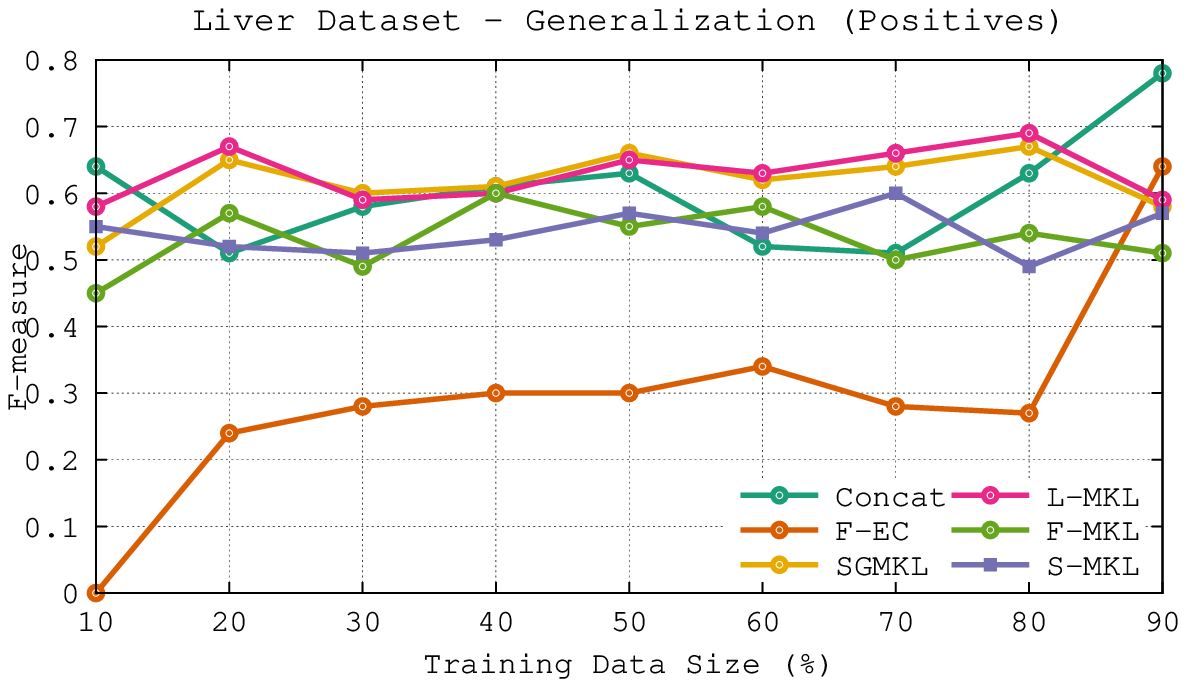}
\includegraphics[width=0.5\textwidth]{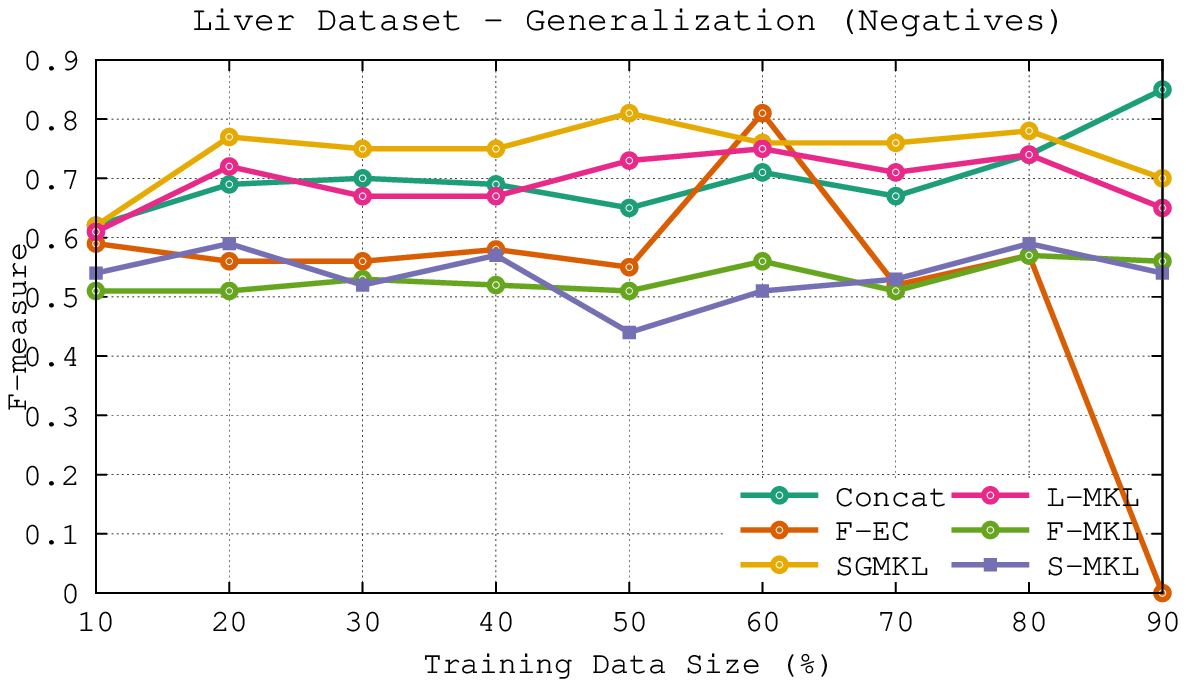}
}
\centerline{(a) \hspace{0.5\textwidth} (b)}
\caption{\small{Visualization of generalization performance data presented in Table~\ref{tab:liverGen}. The variations of f-measures for (a) positive and (b) negative categories are presented with respect to changing training set size.}}
\label{fig:liverGen}
\end{figure*}

\begin{figure*}
\centerline{\includegraphics[width=1\textwidth]{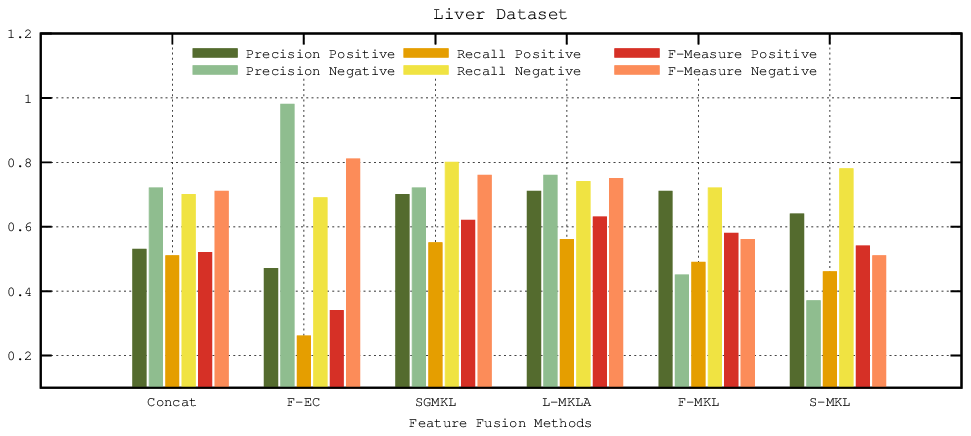}}
\caption{\small{Visualization of the performance analysis data presented in Table~\ref{tab:liverPerform}}}
\label{fig:liverPerform}
\end{figure*}

The Liver Disorder dataset consists of $345$ samples with $42.09\%$ Positive sample. Each sample is represented by $6$ single continuous valued attributes --  viz. Mean Corpuscular Volume (MCV),  Alkphos Alkaline Phosphotase ( AAP ), sgpt alamine aminotransferase (SGPT), sgot aspartate aminotransferase (SGOT) , gammagt gamma-glutamyl transpeptidase (GGT) and number of half-pint equivalents of alcoholic beverages drunk per day (DPD). We have used Linear (LK) and RBF (RK) kernels with each attribute resulting in a total of $12$ feature-kernel combinations.  Performance of individual feature kernel combinations are tabulated in table~\ref{tab:liverFeature} and are visualized in figure~\ref{fig:liverFeature}. Table~\ref{tab:liverGen} and Figure~\ref{fig:liverGen} shows the Generalization  performance of different classifiers on Liver Disorders while Table~\ref{tab:liverPerform} and Figure~\ref{fig:liverPerform} presents the detailed performance analysis of different classifiers when trained on $60\%$ of total available data.

\begin{table*}[htbp]
\caption{Feature performance Analysis of Liver Disorder dataset}
\begin{center}
{\tabulinesep=1.5mm                                           \begin{tabu}{cccccccc}
\hline
\multicolumn{1}{c}{Features} & \multicolumn{3}{c}{Positive}  & & \multicolumn{3}{c}{Negative} \\ \cline{ 2-4}\cline{6-8}
&Precision&Recall&F Measure&&Precision&Recall&F Measure \\ \hline
MCV-LK & 0.42368 & 0.459092 & 0.421284 && 0.52507 & 0.555038 & 0.536074 \\
MCV-RK & 0.438128 & 0.618804 & 0.494224 &&\bf 0.670361 & 0.429232 & 0.452681 \\
AAP-LK & 0.415749 & 0.673579 & 0.487407 && 0.371859 & 0.315939 & 0.313507 \\
AAP-RK & 0.430263 & 0.445382 & 0.432371 && 0.593162 & 0.579905 & 0.581749 \\
SGPT-LK & 0.428257 & \bf0.833881 & 0.543668 && 0.440176 & 0.171336 & 0.205765 \\
SGPT-RK & 0.472466 & 0.66664 & \bf0.550533 && 0.663817 & 0.464059 & 0.54131 \\
SGOT-LK & 0.37804 & 0.762776 & 0.504994 && 0.539381 & 0.233135 & 0.292341 \\
SGOT-RK & 0.443979 & 0.705945 & 0.536724 && 0.653168 & 0.354788 & 0.422216 \\
GGT-LK & 0.359207 & 0.635488 & 0.425339 && 0.557056 & 0.347778 & 0.351379 \\
GGT-RK & 0.432986 & 0.722473 & 0.531231 && 0.642164 & 0.340201 & 0.402715 \\
DPD-LK & 0.434913 & 0.385872 & 0.356303 && 0.487472 & \bf0.604935 & 0.522403 \\
DPD-RK & \bf0.484239 & 0.5004 & 0.459422 && 0.64 & 0.596141 & \bf0.590562 \\ \hline
\end{tabu}}
\end{center}
\label{tab:liverFeature}
\end{table*}

\begin{table*}[htbp]
\caption{Generalization Performance of different algorithms on Liver Disorder dataset.}
\begin{center}
{\tabulinesep=1.5mm                                           \begin{tabu}{ccccccccccr}
\hline
\multicolumn{ 2}{|c|}{\backslashbox{Methods$\downarrow$}{Data Size$\rightarrow$}}&10 & 20 & 30 & 40 & 50 & 60 & 70 & 80 & 90 \\ \hline
\multirow{2}{*}{Concat} & F+  &0.64 & 0.51 & 0.58 & 0.61 & 0.63 & 0.52 & 0.51 & 0.63 & 0.78 \\
\multicolumn{ 1}{c}{} & F-     	&0.62 & 0.69 & 0.7 & 0.69 & 0.65 & 0.71 & 0.67 & 0.74 & 0.85 \\ \hline
\multirow{2}{*}{F-EC}& F+  	&0 & 0.24 & 0.28 & 0.3 & 0.3 & 0.34 & 0.28 & 0.27 & 0.64 \\
\multicolumn{ 1}{c}{} & F-     	&0.59 & 0.56 & 0.56 & 0.58 & 0.55 & 0.81 & 0.52 & 0.57 & 0 \\ \hline
\multirow{2}{*}{SG-MKL}&F+ 	&0.52 & 0.65 & 0.6 & 0.61 & 0.66 & 0.62 & 0.64 & 0.67 & 0.58 \\
\multicolumn{ 1}{c}{} & F-  		&0.62 & 0.77 & 0.75 & 0.75 & 0.81 & 0.76 & 0.76 & 0.78 & 0.7 \\ \hline
\multirow{2}{*}{L-MKL}& F+ 	&0.58 & 0.67 & 0.59 & 0.6 & 0.65 & 0.63 & 0.66 & 0.69 & 0.59 \\
\multicolumn{ 1}{c}{} & F-  		&0.61 & 0.72 & 0.67 & 0.67 & 0.73 & 0.75 & 0.71 & 0.74 & 0.65 \\ \hline
\multirow{2}{*}{F-MKL}& F+ 	&0.45 & 0.57 & 0.49 & 0.6 & 0.55 & 0.58 & 0.5 & 0.54 & 0.51 \\
\multicolumn{ 1}{c}{} & F- 		&0.51 & 0.51 & 0.53 & 0.52 & 0.51 & 0.56 & 0.51 & 0.57 & 0.56 \\ \hline
\multirow{2}{*}{S-MKL}& F+ 	&0.55 & 0.52 & 0.51 & 0.53 & 0.57 & 0.54 & 0.6 & 0.49 & 0.57 \\
\multicolumn{ 1}{c}{} & F- 		&0.54 & 0.59 & 0.52 & 0.57 & 0.44 & 0.51 & 0.53 & 0.59 & 0.54 \\ \hline
\end{tabu}}
\end{center}
\label{tab:liverGen}
\end{table*}

\begin{table*}[htbp]
\caption{\small{The averages and standard deviations (in braces) of performances of different classifiers on Liver Disorder dataset when trained with $60\%$ of available data and the experiments are repeated $10$ times. It may be noted that SG-MKL and L-MKL out performs all other classifiers though biased. While inferior performance of S-EC , F-EC and S-MKL may be attributed to the insufficient data.}}
\begin{center}
{\tabulinesep=1.5mm                                           \begin{tabu}{ccccccccc}
\hline
\multicolumn{ 1}{c}{Methods $\downarrow$} & \multicolumn{ 3}{c}{Positive} & & \multicolumn{ 3}{c}{Negative} & \multicolumn{ 1}{c}{Support} \\ \cline{ 2- 4} \cline{6-8}
\multicolumn{ 1}{l}{} & \multicolumn{1}{l}{Precision} & \multicolumn{1}{l}{Recall} & \multicolumn{1}{l}{F-Measure} && \multicolumn{1}{l}{Precision} & \multicolumn{1}{l}{Recall} & \multicolumn{1}{l}{F-Measure} & \multicolumn{ 1}{c}{ Vectors} \\ \hline
CONCAT & 0.53(0.012) & 0.51(0.0162) & 0.52 (0.04 )& & 0.72(0.031) & 0.7(0.02) & 0.71(0.02) & 0.73(0.0003) \\
F-EC & 0.47(0.0022) & 0.26(0.0081) & 0.34(0.1) & &\bf 0.98(0.0031) & 0.69(0.015) & \bf 0.81(0.012) & 0.68(0.0013) \\
SGMKL & 0.7(0.057) & 0.55(0.307) & 0.62(0.0053) & &0.72(0.0513) & 0.8(0.0579) & 0.76(0.009) &  \bf0.6(0.081) \\
L-MKL &  \bf0.71(0.032) &  \bf0.56(0.0802) &  \bf0.63(0.5) && 0.76(0.0413) & 0.74(0.0482) & 0.75(0.091) & 0.78(0.0921) \\
F-MKL & 0.71(0.391) & 0.49(0.473) & 0.58(0.0541) & &0.45(0.057) & 0.72(0.0301) & 0.56(0.0029) & 0.62(0.012) \\
S-MKL & 0.64(0.0713) & 0.46(0.015) & 0.54(0.0874) & &0.37(0.0104) & 0.78(0.1082) & 0.51(0.0809) & 0.7(0.1098) \\ \hline
\end{tabu}}
\end{center}
\label{tab:liverPerform}
\end{table*}

\clearpage

\subsection{Ionosphere Dataset}
\begin{sidewaysfigure*}[htbp]
\centerline{\includegraphics[width=1\textheight]{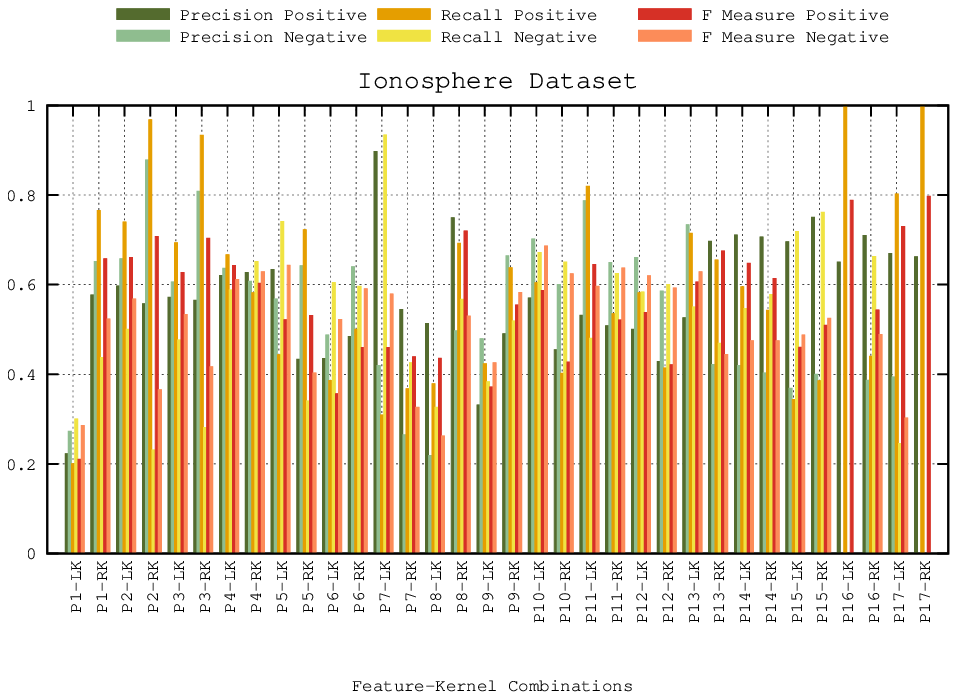}}
\caption{\small{Visualization of the performance analysis data presented in Table~\ref{tab:ionosphereFeature}. The precision, recall and f-measures for different feature kernel combinations are shown for the Ionosphere dataset.}}
\label{fig:ionosphereFeature}
\end{sidewaysfigure*}

\begin{figure*}
\centerline{\includegraphics[width=0.5\textwidth]{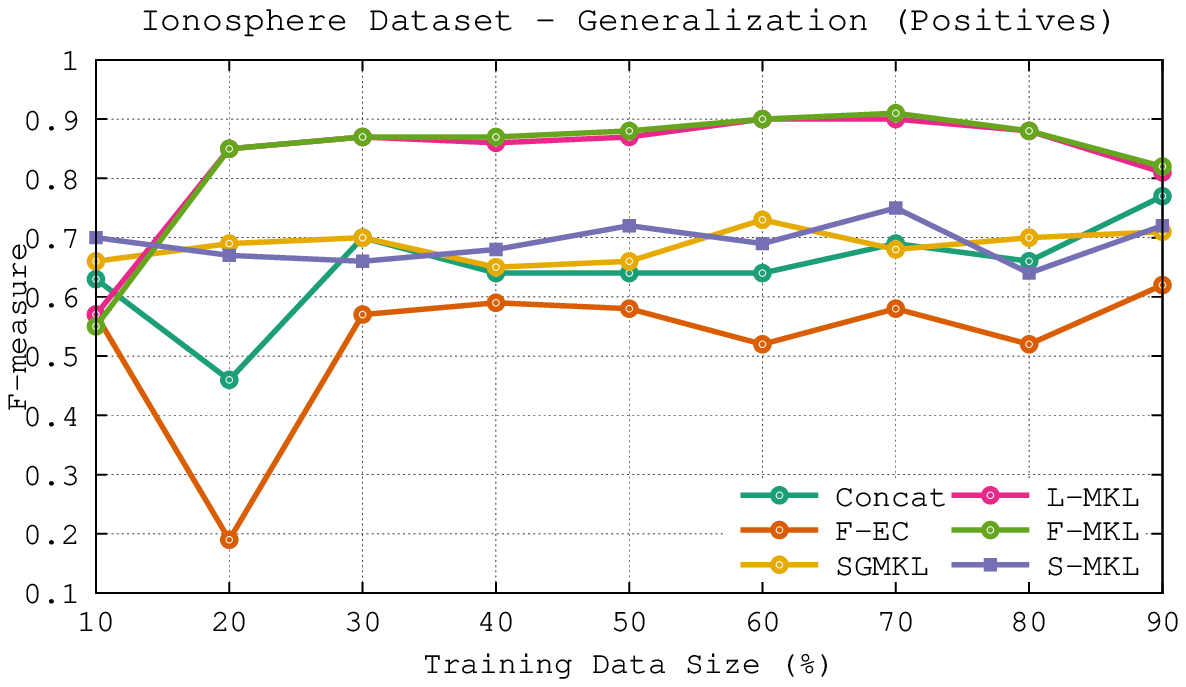}
\includegraphics[width=0.5\textwidth]{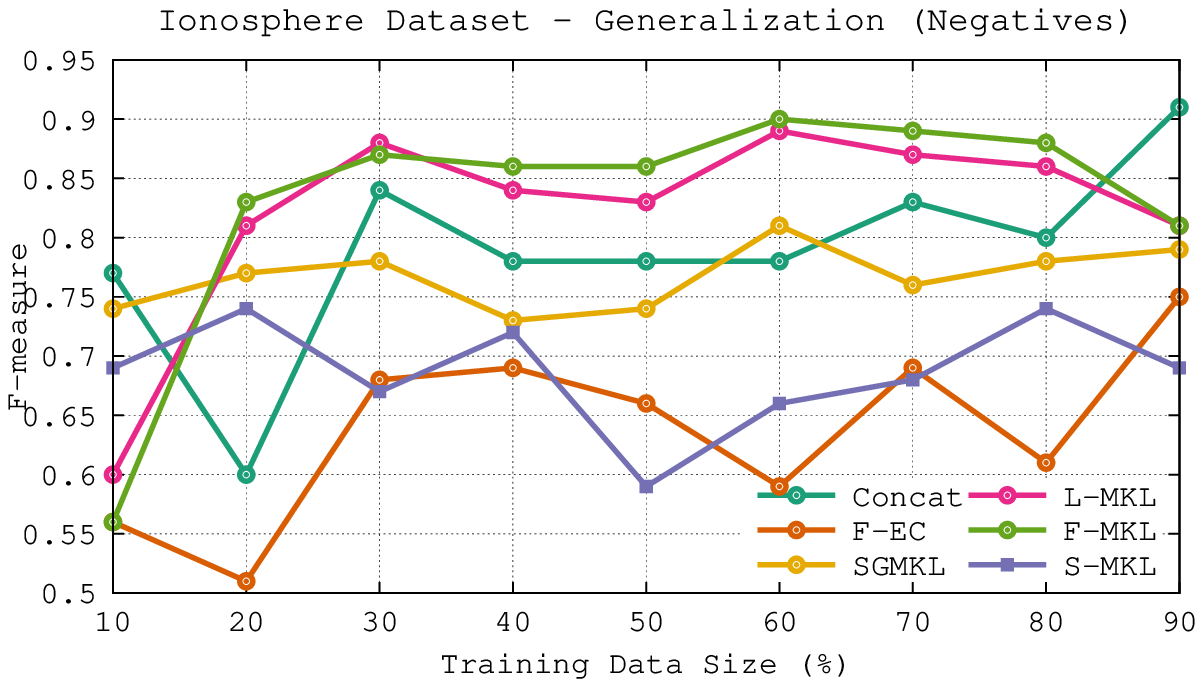}}
\caption{\small{Visualization of generalization performance data presented in Table~\ref{tab:ionosphereGen}. The variations of f-measures for (a) positive and (b) negative categories are presented with respect to changing training set size.}}
\label{fig:ionosphereGen}
\end{figure*}

\begin{figure*}[htbp]
\centerline{\includegraphics[width=1\textwidth]{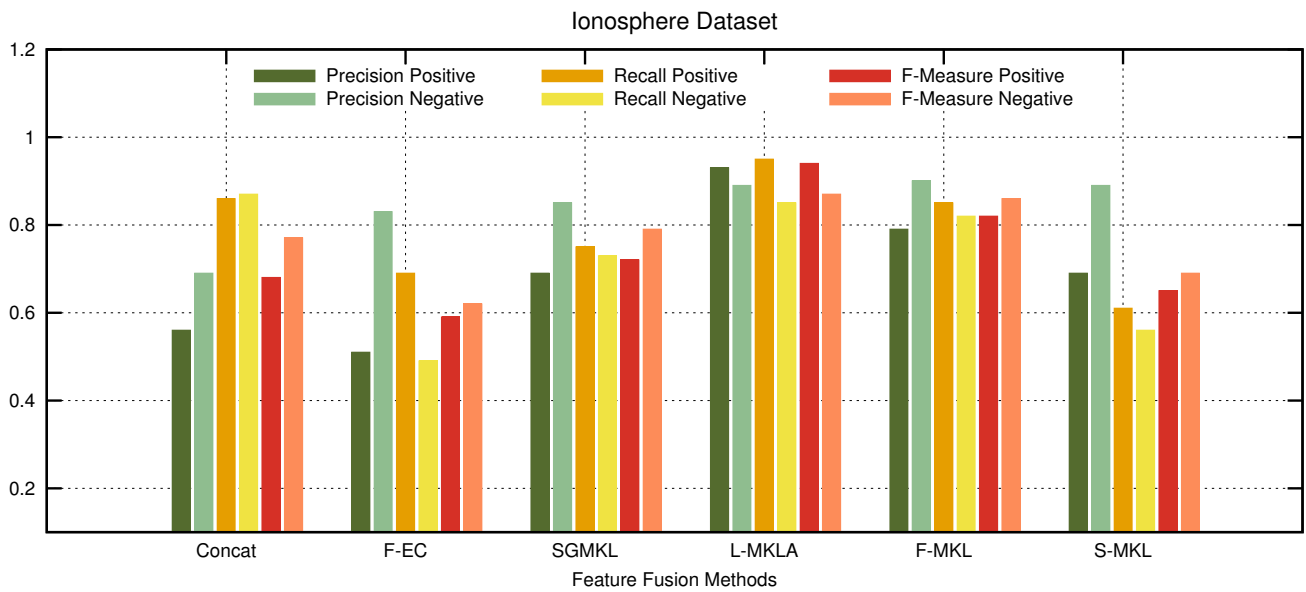}}
\caption{\small{Visualization of the performance analysis data presented in Table~\ref{tab:ionospherePerform}}}
\label{fig:ionospherePerform}
\end{figure*}

The Ionosphere dataset consists of $351$ samples with $64.1\%$ Positive sample. Each sample is a combination of $17$ distinct $2$ dimensional features.( represented by P1 through P17). We have used Linear (LK) and RBF (RK) kernels with each attribute resulting in a total of $34$ feature-kernel combinations.  Performance of individual feature kernel combinations are tabulated in table~\ref{tab:ionosphereFeature} and are visualized in figure~\ref{fig:ionosphereFeature}. Table~\ref{tab:ionosphereGen} and Figure~\ref{fig:ionosphereGen} shows the Generalization  performance of different algorithms on Ionosphere dataset while Table~\ref{tab:ionospherePerform} and Figure~\ref{fig:ionospherePerform} presents the detailed performance analysis of different classifiers when trained on $60\%$ of total available data.
\label{subsec:ionosphere}
\begin{table*}[htbp]
\caption{Feature performance Analysis of Ionosphere dataset}
\begin{center}
{\tabulinesep=1.5mm                                           \begin{tabu}{cccccccc}
\hline

\multicolumn{1}{c}{Features} & \multicolumn{3}{c}{Positivve}  && \multicolumn{3}{c}{Negative} \\ \cline{ 2-4} \cline{6-8}
&Precision&Recall&F Measure&&Precision&Recall&F Measure \\ \hline
P1-LK & 0.222222 & 0.2 & 0.210526 && 0.272727 & 0.3 & 0.285714 \\
P1-RK & 0.576471 & 0.765625 & 0.657718 && 0.651163 & 0.4375 & 0.523364 \\
P2-LK & 0.596639 & 0.739583 & 0.660465 && 0.657534 & 0.5 & 0.568047 \\
P2-RK & 0.557196 & 0.967949 & 0.70726 &&\bf0.878049 & 0.230769 & 0.365482 \\
P3-LK & 0.571984 & 0.693396 & 0.626866 && 0.606061 & 0.47619 & 0.533333 \\
P3-RK & 0.564841 & 0.933333 & 0.70377 && 0.808219 & 0.280952 & 0.416961 \\
P4-LK & 0.620155 & 0.666667 & 0.64257 && 0.636364 & 0.588235 & 0.611354 \\
P4-RK & 0.626761 & 0.581699 & 0.60339 && 0.607362 & 0.651316 & 0.628571 \\
P5-LK & 0.633508 & 0.443223 & 0.521552 && 0.568182 & 0.740741 & 0.643087 \\
P5-RK & 0.432986 & 0.722473 & 0.531231 && 0.642164 & 0.340201 & 0.402715 \\
P6-LK & 0.434913 & 0.385872 & 0.356303 && 0.487472 & 0.604935 & 0.522403 \\
P6-RK & 0.484239 & 0.5004 & 0.459422 && 0.64 & 0.596141 & 0.590562 \\
P7-LK & \bf 0.896774 & 0.308889 & 0.459504 && 0.419776 &  \bf0.93361 & 0.579151 \\
P7-RK & 0.544444 & 0.3675 & 0.438806 && 0.264535 & 0.425234 & 0.326165 \\
P8-LK & 0.513514 & 0.378917 & 0.436066 && 0.218638 & 0.326203 & 0.261803 \\
P8-RK & 0.748918 & 0.692 & 0.719335 && 0.496732 & 0.567164 & 0.529617 \\
P9-LK & 0.331325 & 0.423077 & 0.371622 && 0.479167 & 0.383333 & 0.425926 \\
P9-RK & 0.490066 & 0.637931 & 0.554307 && 0.664 & 0.51875 & 0.582456 \\
P10-LK & 0.570093 & 0.60396 & 0.586538 && 0.701493 & 0.671429 &  \bf0.686131 \\
P10-RK & 0.454545 & 0.402299 & 0.426829 && 0.6 & 0.65 & 0.624 \\
P11-LK & 0.531532 & 0.819444 & 0.644809 && 0.786885 & 0.48 & 0.596273 \\
P11-RK & 0.508197 & 0.534483 & 0.521008 && 0.649351 & 0.625 & 0.636943 \\
P12-LK & 0.5 & 0.581395 & 0.537634 && 0.660377 & 0.583333 & 0.619469 \\
P12-RK & 0.428571 & 0.413793 & 0.421053 && 0.585366 & 0.6 & 0.592593 \\
P13-LK & 0.526316 & 0.714286 & 0.606061 && 0.733333 & 0.55 & 0.628571 \\
P13-RK & 0.696682 & 0.654788 & 0.675086 && 0.421642 & 0.46888 & 0.444008 \\
P14-LK & 0.710448 & 0.595 & 0.647619 && 0.419355 & 0.546729 & 0.474645 \\
P14-RK & 0.70632 & 0.542857 & 0.613893 && 0.402985 & 0.57754 & 0.474725 \\
P15-LK & 0.695946 & 0.343333 & 0.459821 && 0.36859 & 0.71875 & 0.487288 \\
P15-RK & 0.75 & 0.385542 & 0.509284 && 0.4 & 0.761194 & 0.524422 \\
P16-LK & 0.650327 &  \bf1 & 0.788119 && 0 & 0 & 0 \\
P16-RK & 0.709677 & 0.44 & 0.54321 && 0.386861 & 0.6625 & 0.488479 \\
P17-LK & 0.669421 & 0.80198 & 0.72973 && 0.393939 & 0.245283 & 0.302326 \\
P17-RK & 0.662338 & \bf 1 & \bf0.796875 && 0 & 0 & 0 \\ \hline
\end{tabu}}
\end{center}
\label{tab:ionosphereFeature}
\end{table*}

\begin{table*}[htbp]
\caption{Generalization Performance of different algorithms on Ionosphere dataset.}
\begin{center}
{\tabulinesep=1.5mm                                           \begin{tabu}{ccccccccccr}
\hline
\multicolumn{ 2}{|c|}{\backslashbox{Methods$\downarrow$}{Data Size$\rightarrow$}}&10 & 20 & 30 & 40 & 50 & 60 & 70 & 80 & 90 \\ \hline
\multirow{ 2}{*}{Concat}& F+  &0.63 & 0.46 & 0.7 & 0.64 & 0.64 & 0.68 & 0.69 & 0.66 & 0.77 \\
\multicolumn{ 1}{c}{} & F-     	&0.77 & 0.6 & 0.84 & 0.78 & 0.78 & 0.77 & 0.83 & 0.8 & 0.91 \\ \hline
\multirow{2}{*}{F-EC}& F+  	&0.57 & 0.19 & 0.57 & 0.59 & 0.58 & 0.59 & 0.58 & 0.52 & 0.62 \\
\multicolumn{ 1}{c}{} & F-     	&0.56 & 0.51 & 0.68 & 0.69 & 0.66 & 0.62 & 0.69 & 0.61 & 0.75 \\ \hline
\multicolumn{ 1}{c}{S-EC}& F+  	&0.66 & 0.59 & 0.66 & 0.68 & 0.67 & 0.66 & 0.67 & 0.61 & 0.71 \\
\multicolumn{ 1}{c}{} & F-     	&0.65 & 0.6 & 0.77 & 0.78 & 0.75 & 0.68 & 0.78 & 0.7 & 0.84 \\ \hline
\multirow{2}{*}{SG-MKL}&F+ 	&0.66 & 0.69 & 0.7 & 0.65 & 0.66 & 0.72 & 0.68 & 0.7 & 0.71 \\
\multicolumn{ 1}{c}{} & F-  		&0.74 & 0.77 & 0.78 & 0.73 & 0.74 & 0.79 & 0.76 & 0.78 & 0.79 \\ \hline
\multirow{2}{*}{L-MKL}& F+ 	&0.57 & 0.85 & 0.87 & 0.86 & 0.87 & 0.94 & 0.9 & 0.88 & 0.81 \\
\multicolumn{ 1}{c}{} & F-  		&0.6 & 0.81 & 0.88 & 0.84 & 0.83 & 0.87 & 0.87 & 0.86 & 0.81 \\ \hline
\multirow{2}{*}{F-MKL}& F+ 	&0.55 & 0.85 & 0.87 & 0.87 & 0.88 & 0.82 & 0.91 & 0.88 & 0.82 \\
\multicolumn{ 1}{c}{} & F- 		&0.56 & 0.83 & 0.87 & 0.86 & 0.86 & 0.86 & 0.89 & 0.88 & 0.81 \\ \hline
\multirow{2}{*}{S-MKL}& F+ 	&0.7 & 0.67 & 0.66 & 0.68 & 0.72 & 0.65 & 0.75 & 0.64 & 0.72 \\
\multicolumn{ 1}{c}{} & F- 		&0.69 & 0.74 & 0.67 & 0.72 & 0.59 & 0.69 & 0.68 & 0.74 & 0.69 \\ \hline
\end{tabu}}
\end{center}
\label{tab:ionosphereGen}
\end{table*}

\begin{table*}[htbp]
\caption{\small{The averages and standard deviations (in braces) of performances of different classifiers on Ionosphere dataset when trained with $60\%$ of available data and the experiments are repeated $10$ times. It may be noted that SG-MKL and L-MKL out performs all other classifiers though biased. While inferior performance of S-EC , F-EC and S-MKL may be attributed to the insufficient data.}}
\begin{center}
{\tabulinesep=1.5mm                                           \begin{tabu}{ccccccccc}
\hline
\multicolumn{ 1}{c}{Methods $\downarrow$} & \multicolumn{ 3}{c}{Positive} & & \multicolumn{ 3}{c}{Negative} & \multicolumn{ 1}{c}{Support} \\ \cline{ 2- 4} \cline{6-8}
\multicolumn{ 1}{l}{} & \multicolumn{1}{l}{Precision} & \multicolumn{1}{l}{Recall} & \multicolumn{1}{l}{F-Measure} && \multicolumn{1}{l}{Precision} & \multicolumn{1}{l}{Recall} & \multicolumn{1}{l}{F-Measure} & \multicolumn{ 1}{c}{ Vectors} \\ \hline
CONCAT & 0.56(0.0059) & 0.86(0.0099) & 0.68(0.0053) & & 0.69(0.0703) &  \bf0.87(0.0207) & 0.77(0.034) & 0.76(0.0001) \\ \hline
F-EC & 0.51(0.0087) & 0.69(0.0609) & 0.59(0.0189) & & 0.83(0.0771) & 0.49(0.0120) & 0.62(0.0124) & 0.71(0.0001) \\ \hline
SGMKL & 0.69(0.1059) & 0.75(0.0262) & 0.72(0.1041)  && 0.85(0.1578) & 0.73(0.0026) & 0.79(0.012) & 0.59(0.081) \\ \hline
L-MKL &  \bf0.93(0.0037) &  \bf0.95(0.009) &  \bf0.94(0.0001)  && 0.89(0.0067) & 0.85(0.014) &  \bf0.87(0.009) & 0.61(0.081) \\ \hline
F-MKL & 0.79(0.0759) & 0.85(0.0018) & 0.82(0.0114)  && \bf 0.9(0.0081) & 0.82(0.051) & 0.86(0.0121) &  \bf0.43(0.0801) \\ \hline
S-MKL & 0.69(0.0928) & 0.61(0.0702) & 0.65(0.1534)  && 0.89(0.0005) & 0.56(0.0243) & 0.69(0.0729) & 0.69(0.0009) \\ \hline
\end{tabu}}
\end{center}
\label{tab:ionospherePerform}
\end{table*}

\clearpage

\subsection{Breast Cancer Dataset}
\label{subsec:cancer}

\begin{sidewaysfigure*}[htbp]
\centerline{\includegraphics[width=1\textheight]{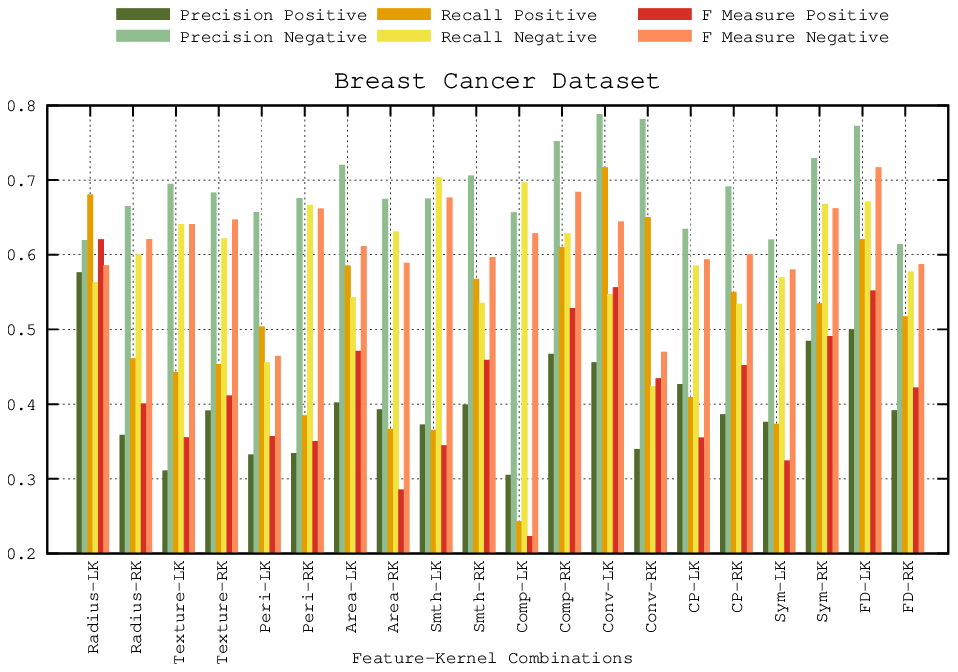}}
\caption{\small{Visualization of the performance analysis data presented in Table~\ref{tab:cancerFeature}. The precision, recall and f-measures for different feature kernel combinations are shown for the Breast Cancer dataset.}}
\label{fig:cancerFeature}
\end{sidewaysfigure*}

\begin{figure*}
\centerline{\includegraphics[width=0.5\textwidth]{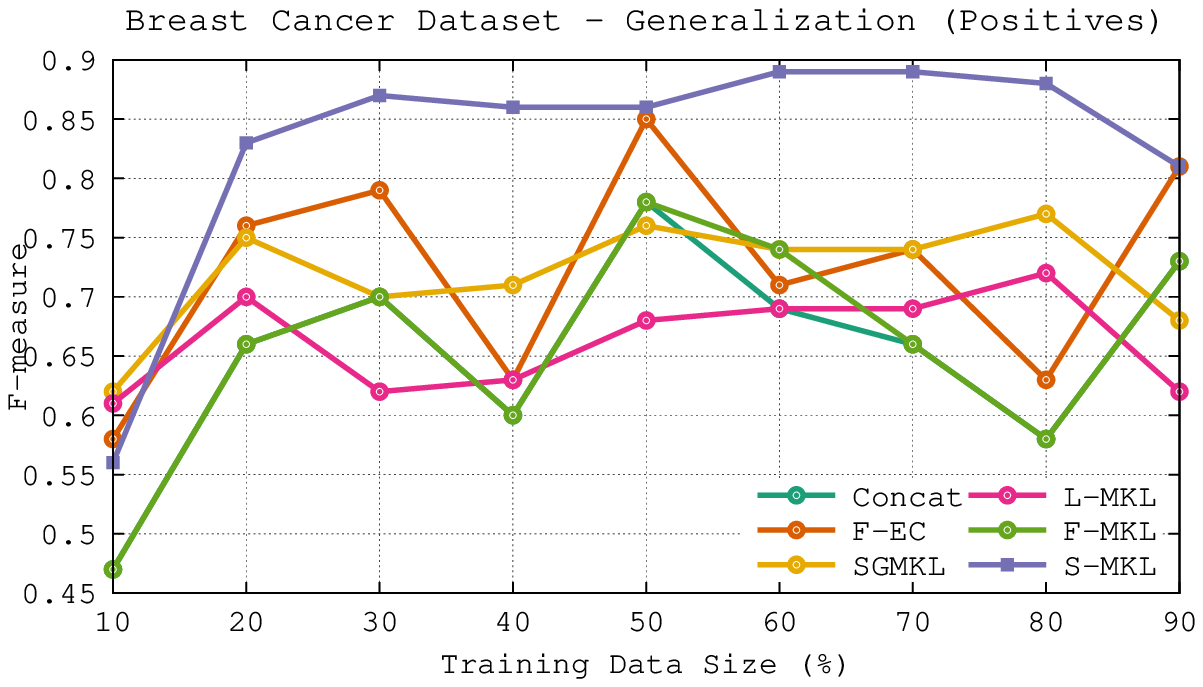}
\includegraphics[width=0.5\textwidth]{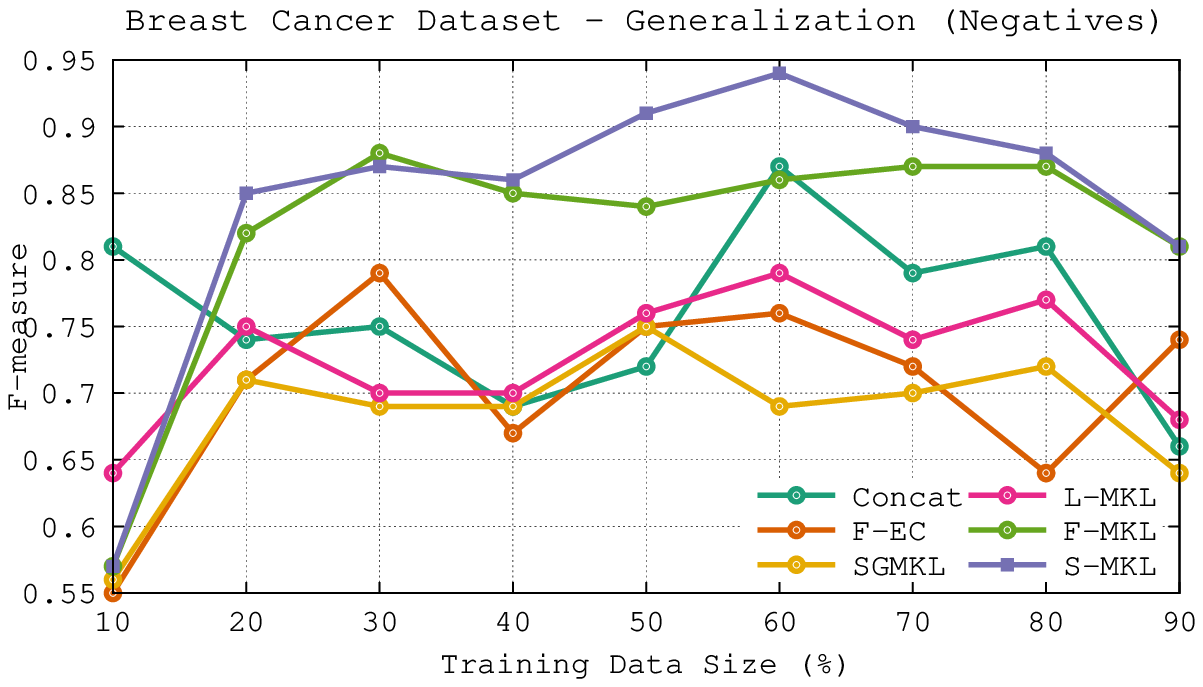}}
\caption{\small{Visualization of generalization performance data presented in Table~\ref{tab:cancerGen}. The variations of f-measures for (a) positive and (b) negative categories are presented with respect to changing training set size.}}
\label{fig:cancerGen}
\end{figure*}

\begin{figure*}[htbp]
\centerline{\includegraphics[width=1\textwidth]{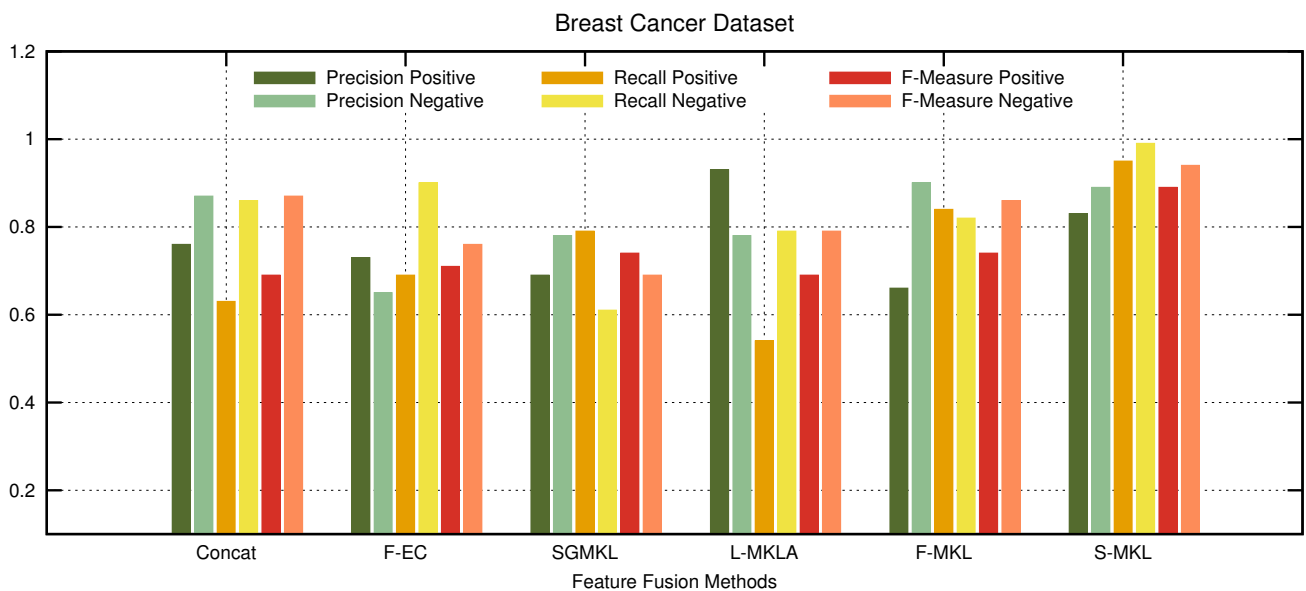}}
\caption{\small{Visualization of the performance analysis data presented in Table~\ref{tab:cancerPerform}}}
\label{fig:cancerPerform}
\end{figure*}

The Breast Cancer dataset consists of $683$ samples with $34.99\%$ Positive sample. Each sample is represented by $10$ single continuous valued attributes --  viz. Mean of distances from center to points on the perimeter (Radius),  Texture (standard deviation of gray-scale values), perimeter (Peri), Area, smoothness (local variation in radius lengths)(Smth),compactness (Comp), concavity (severity of concave portions of the contour)(Conv),concave points (number of concave portions of the contour)(CP), symmetry(Sym) and fractal dimension ("coastline approximation" - 1)(FD). We have used Linear (LK) and RBF (RK) kernels with each attribute resulting in a total of $20$ feature-kernel combinations.  Performance of individual feature kernel combinations are tabulated in table~\ref{tab:cancerFeature} and are visualized in figure~\ref{fig:cancerFeature}. Table~\ref{tab:cancerGen} and Figure~\ref{fig:cancerGen} shows the Generalization  performance of different classifiers on Breast Cancer dataset while Table~\ref{tab:cancerPerform} and Figure~\ref{fig:cancerPerform} presents the detailed performance analysis of different classifiers when trained on $60\%$ of total available data.
\begin{table*}[htbp]
\caption{Feature Performance Analysis of Breast Cancer dataset}
\begin{center}
{\tabulinesep=1.5mm                                           \begin{tabu}{cccccccc}
\hline

\multicolumn{1}{c}{Features} & \multicolumn{3}{c}{Positivve}  && \multicolumn{3}{c}{Negative} \\ \cline{ 2-4} \cline{6-8}
&Precision&Recall&F Measure&&Precision&Recall&F Measure \\ \hline
Radius-LK &  \bf0.575843 & 0.679942 &  \bf0.620116  && 0.61881 & 0.562549 & 0.585236 \\
Radius-RK & 0.358298 & 0.46064 & 0.400244  && 0.664583 & 0.599974 & 0.620562 \\
Texture-LK & 0.310375 & 0.44232 & 0.355099  && 0.694573 & 0.640389 & 0.640404 \\
Texture-RK & 0.390645 & 0.452835 & 0.410879  && 0.682653 & 0.621174 & 0.646566 \\
Peri-LK & 0.331731 & 0.503465 & 0.356626  && 0.656649 & 0.455214 & 0.46397 \\
Peri-RK & 0.33387 & 0.384269 & 0.349865  && 0.675208 & 0.666217 & 0.661113 \\
Area-LK & 0.401502 & 0.584811 & 0.47062 & & 0.719853 & 0.542596 & 0.611108 \\
Area-RK & 0.392282 & 0.366113 & 0.284813  && 0.67416 & 0.630207 & 0.588556 \\
Smth-LK & 0.372135 & 0.364853 & 0.344443  && 0.674702 &  \bf0.703392 & 0.676147 \\
Smth-RK & 0.39873 & 0.56741 & 0.458666  && 0.705299 & 0.535057 & 0.59635 \\
Comp-LK & 0.304817 & 0.242383 & 0.22284  && 0.656172 & 0.696269 & 0.628304 \\
Comp-RK & 0.466912 & 0.609577 & 0.528002  && 0.751752 & 0.628203 & 0.68367 \\
Conv-LK & 0.455532 &  \bf0.716436 & 0.555755  &&  \bf0.79221 & 0.546826 & 0.643782 \\
Conv-RK & 0.339006 & 0.649939 & 0.433862  && 0.780864 & 0.423623 & 0.469333 \\
CP-LK & 0.426094 & 0.408868 & 0.354523  & & 0.634324 & 0.584691 & 0.593063 \\
CP-RK & 0.385767 & 0.549764 & 0.451743  && 0.690697 & 0.533478 & 0.599916 \\
Sym-LK & 0.375772 & 0.372904 & 0.323609  && 0.619693 & 0.569145 & 0.579481 \\
Sym-RK & 0.484009 & 0.534213 & 0.490603  && 0.72898 & 0.667763 & 0.661767 \\
FD-LK & 0.499404 & 0.620618 & 0.551432  && 0.771952 & 0.670763 &  \bf0.71628 \\
FD-RK & 0.391119 & 0.516801 & 0.421552  && 0.613849 & 0.576895 & 0.586759 \\
\end{tabu}}
\end{center}
\label{tab:cancerFeature}
\end{table*}

\begin{table*}[htbp]
\caption{Generalization Performance of different algorithms on Breast Cancer dataset.}
\begin{center}
{\tabulinesep=1.5mm                                           \begin{tabu}{ccccccccccr}
\hline
\multicolumn{ 2}{|c|}{\backslashbox{Methods$\downarrow$}{Data Size$\rightarrow$}}&10 & 20 & 30 & 40 & 50 & 60 & 70 & 80 & 90 \\ \hline
\multirow{ 2}{*}{Concat}& F+  &0.47 & 0.66 & 0.7 & 0.6 & 0.78 & 0.69 & 0.66 & 0.58 & 0.73 \\
\multicolumn{ 1}{c}{} & F-     	&0.81 & 0.74 & 0.75 & 0.69 & 0.72 & 0.87 & 0.79 & 0.81 & 0.66 \\ \hline
\multirow{2}{*}{F-EC}& F+  	&0.58 & 0.76 & 0.79 & 0.63 & 0.85 & 0.71 & 0.74 & 0.63 & 0.81 \\
\multicolumn{ 1}{c}{} & F-     	&0.55 & 0.71 & 0.79 & 0.67 & 0.75 & 0.76 & 0.72 & 0.64 & 0.74 \\ \hline
\multirow{2}{*}{SG-MKL}&F+ 	&0.62 & 0.75 & 0.7 & 0.71 & 0.76 & 0.74 & 0.74 & 0.77 & 0.68 \\
\multicolumn{ 1}{c}{} & F-  		&0.56 & 0.71 & 0.69 & 0.69 & 0.75 & 0.69 & 0.7 & 0.72 & 0.64 \\ \hline
\multirow{2}{*}{L-MKL}& F+ 	&0.61 & 0.7 & 0.62 & 0.63 & 0.68 & 0.69 & 0.69 & 0.72 & 0.62 \\
\multicolumn{ 1}{c}{} & F-  		&0.64 & 0.75 & 0.7 & 0.7 & 0.76 & 0.79 & 0.74 & 0.77 & 0.68 \\ \hline
\multirow{2}{*}{F-MKL}& F+ 	&0.47 & 0.66 & 0.7 & 0.6 & 0.78 & 0.74 & 0.66 & 0.58 & 0.73 \\
\multicolumn{ 1}{c}{} & F- 		&0.57 & 0.82 & 0.88 & 0.85 & 0.84 & 0.86 & 0.87 & 0.87 & 0.81 \\ \hline
\multirow{2}{*}{S-MKL}& F+ 	&0.56 & 0.83 & 0.87 & 0.86 & 0.86 & 0.89 & 0.89 & 0.88 & 0.81 \\
\multicolumn{ 1}{c}{} & F- 		&0.57 & 0.85 & 0.87 & 0.86 & 0.91 & 0.94 & 0.9 & 0.88 & 0.81 \\ \hline
\end{tabu}}
\end{center}
\label{tab:cancerGen}
\end{table*}

\begin{table*}[htbp]
\caption{\small{The averages and standard deviations (in braces) of performances of different classifiers on Breast Cancer dataset when trained with $60\%$ of available data and the experiments are repeated $10$ times. }}
\begin{center}
{\tabulinesep=1.5mm                                           \begin{tabu}{ccccccccc}
\hline
\multicolumn{ 1}{c}{Methods $\downarrow$} & \multicolumn{ 3}{c}{Positive} & & \multicolumn{ 3}{c}{Negative} & \multicolumn{ 1}{c}{Support} \\ \cline{ 2- 4} \cline{6-8}
\multicolumn{ 1}{l}{} & \multicolumn{1}{l}{Precision} & \multicolumn{1}{l}{Recall} & \multicolumn{1}{l}{F-Measure} && \multicolumn{1}{l}{Precision} & \multicolumn{1}{l}{Recall} & \multicolumn{1}{l}{F-Measure} & \multicolumn{ 1}{c}{ Vectors} \\ \hline
Concat & 0.76(0.1098) & 0.63(0.081) & 0.69(0.0236) && 0.87(0.0045) & 0.86(0.001) & 0.87(0.0113) & 0.49(0.1890) \\
F-EC & 0.73(0.0991) & 0.69(0.4140) & 0.71(0.0613) && 0.65(0.0267) & 0.9(0.0029) & 0.76(0.0904) & 0.64(0.0135) \\
SGMKL & 0.69(0.0193) & 0.79(0.012) & 0.74(0.0019) && 0.78(0.0045) & 0.61(0.0712) & 0.69(0.0089) & 0.65(0.0089) \\
L-MKL & 0.93(0.0019) & 0.54(0.0027) & 0.69(0.0078) && 0.78(0.0031) & 0.79(0.0182) & 0.79(0.012) & 0.5(0.0023) \\
F-MKL & 0.66(0.021) & 0.84(0.0319) & 0.74(0.023) && \bf0.9(0.0901) & 0.82(0.0013) & 0.86(0.008) & 0.43(0.0081) \\
S-MKL & 0.83(0.0176) & \bf0.95(0.0076) & \bf0.89(0.0071) && 0.89(0.0028) & \bf0.99(0.0009) & \bf0.94(0.130) & \bf0.35(0.00724 \\ \hline
\end{tabu}}
\end{center}
\label{tab:cancerPerform}
\end{table*}

\clearpage

\subsection{Diabetes Dataset}
\begin{sidewaysfigure*}[htbp]
\centerline{\includegraphics[width=1\textheight]{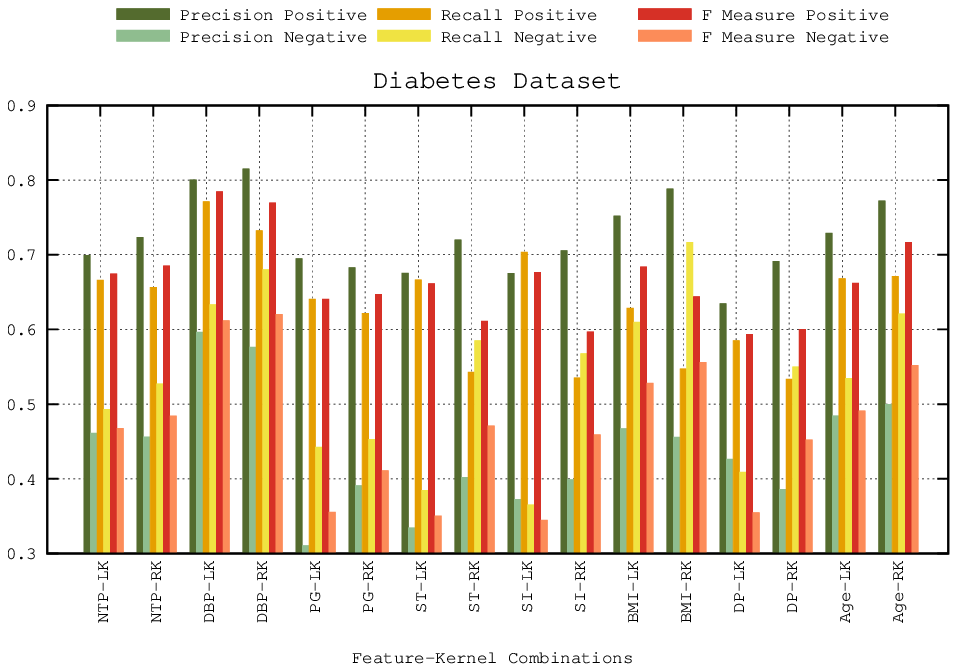}}
\caption{\small{Visualization of the performance analysis data presented in Table~\ref{tab:diabetesFeature}. The precision, recall and f-measures for different feature kernel combinations are shown for the Diabetes dataset.}}
\label{fig:diabetesFeature}
\end{sidewaysfigure*}

\begin{figure*}
\centerline{\includegraphics[width=0.5\textwidth]{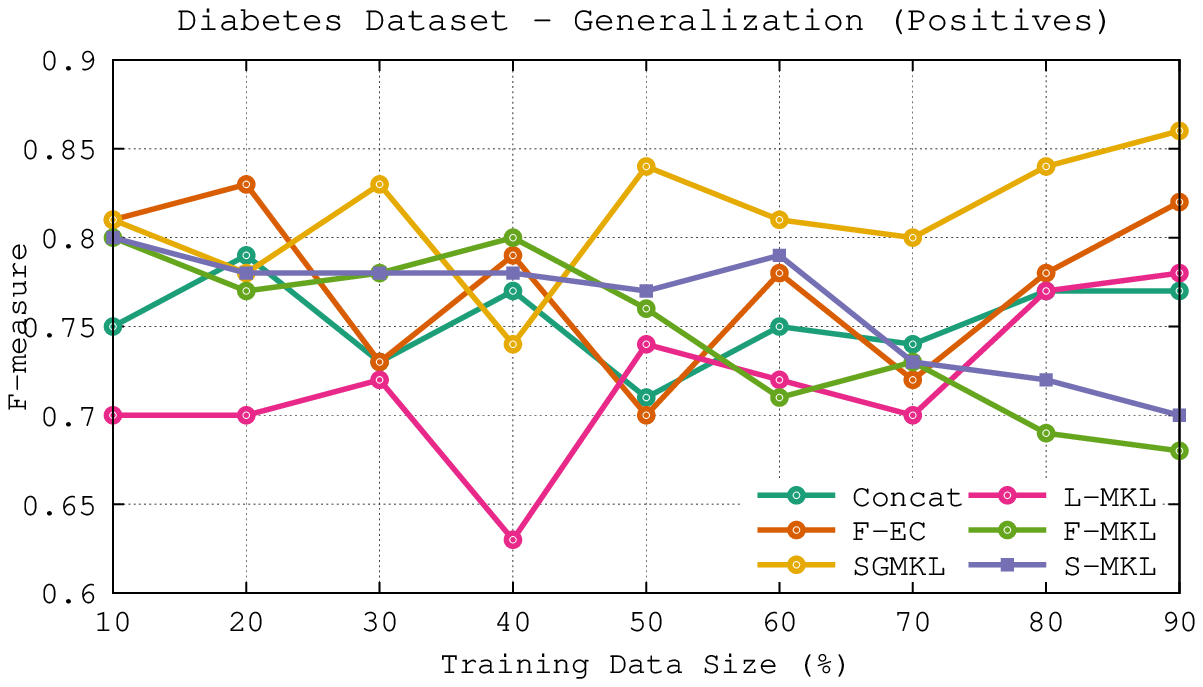}
\includegraphics[width=0.5\textwidth]{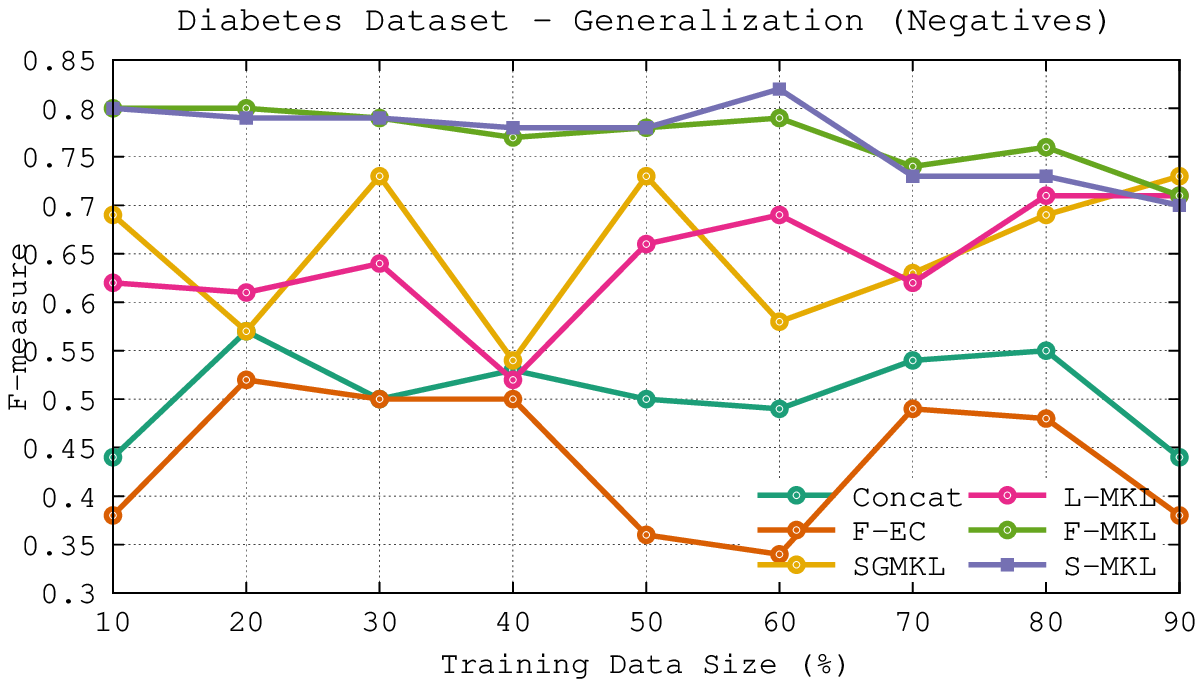}}
\caption{\small{Visualization of generalization performance data presented in Table~\ref{tab:diabetesGen}. The variations of f-measures for (a) positive and (b) negative categories are presented with respect to changing training set size.}}
\label{fig:diabetesGen}
\end{figure*}

\begin{figure*}[htbp]
\centerline{\includegraphics[width=1\textwidth]{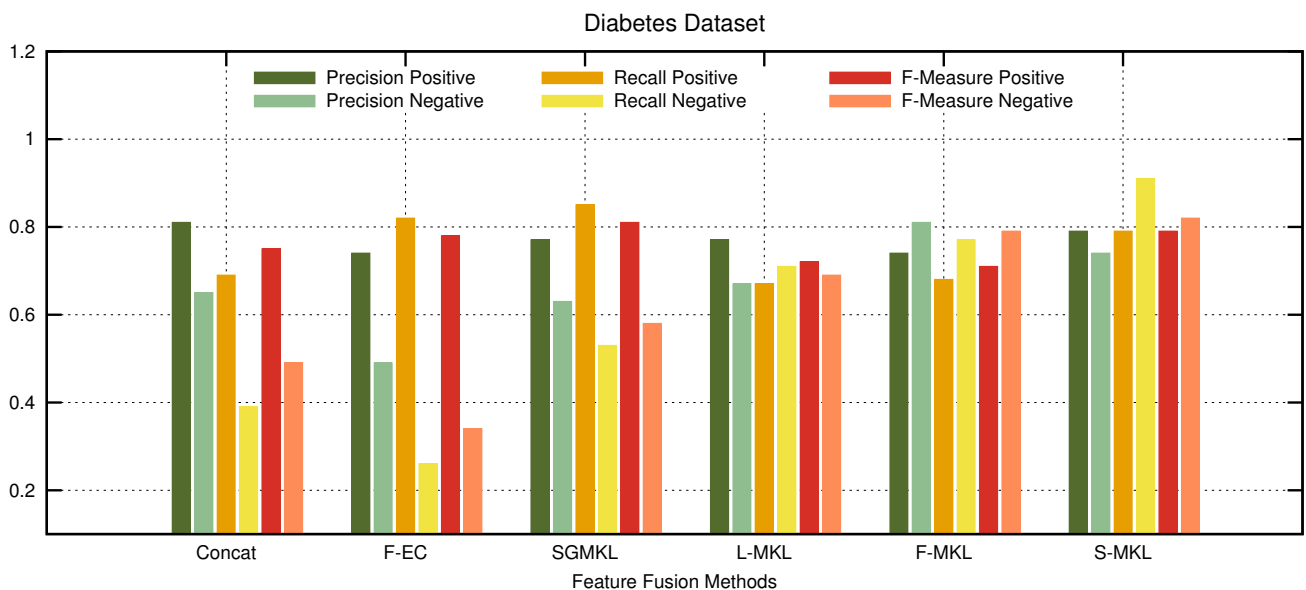}}
\caption{\small{Visualization of the performance analysis data presented in Table~\ref{tab:diabetesPerform}}}
\label{fig:diabetesPerform}
\end{figure*}

The Diabetes dataset consists of $768$ samples with $65.1\%$ Positive sample. Each sample is represented by $8$ single continuous valued attributes --  viz. Number of times pregnant(NTP), Plasma glucose concentration a 2 hours in an oral glucose tolerance test (PG) , Diastolic blood pressure (mm Hg)(DBP), Triceps skin fold thickness (mm)(ST), 2-Hour serum insulin (mu U/ml)(SI), Body mass index (BMI), Diabetes pedigree function (DP) and Age. We have used Linear (LK) and RBF (RK) kernels with each attribute resulting in a total of $16$ feature-kernel combinations.  Performance of individual feature kernel combinations are tabulated in table~\ref{tab:diabetesFeature} and are visualized in figure~\ref{fig:diabetesFeature}. Table~\ref{tab:diabetesGen} and Figure~\ref{fig:diabetesGen} shows the Generalization  performance of different classifiers on Breast Cancer dataset while Table~\ref{tab:diabetesPerform} and Figure~\ref{fig:diabetesPerform} presents the detailed performance analysis of different classifiers when trained on $60\%$ of total available data.
\label{subsec:diabetes}
\begin{table*}[htbp]
\caption{Feature performance Analysis of Diabetes dataset}
\begin{center}
{\tabulinesep=1.5mm                                           \begin{tabu}{cccccccc}
\hline

\multicolumn{1}{c}{Features} & \multicolumn{3}{c}{Positive}  & \multicolumn{3}{c}{Negative} \\ \cline{ 2-4} \cline{6-8}
&Precision&Recall&F Measure&&Precision&Recall&F Measure \\ \hline
NTP-LK & 0.698807 & 0.665722 & 0.674274 && 0.46096 & 0.492702 & 0.467352 \\
NTP-RK & 0.723 & 0.656164 & 0.684975 && 0.455736 & 0.526967 & 0.483911 \\
DBP-LK & 0.79998 & \bf 0.770933 &  \bf0.784133 &&  \bf0.596206 & 0.633064 & 0.61155 \\
DBP-RK &  \bf0.814627 & 0.732131 & 0.769109 && 0.575843 & 0.679942 &  \bf0.620116 \\
PG-LK & 0.694573 & 0.640389 & 0.640404 && 0.310375 & 0.44232 & 0.355099 \\
PG-RK & 0.682653 & 0.621174 & 0.646566 && 0.390645 & 0.452835 & 0.410879 \\
ST-LK & 0.675208 & 0.666217 & 0.661113 && 0.33387 & 0.384269 & 0.349865 \\
ST-RK & 0.719853 & 0.542596 & 0.611108 && 0.401502 & 0.584811 & 0.47062 \\
SI-LK & 0.674702 & 0.703392 & 0.676147 && 0.372135 & 0.364853 & 0.344443 \\
SI-RK & 0.705299 & 0.535057 & 0.59635 && 0.39873 & 0.56741 & 0.458666 \\
BMI-LK & 0.751752 & 0.628203 & 0.68367 && 0.466912 & 0.609577 & 0.528002 \\
BMI-RK & 0.787882 & 0.546826 & 0.643782 && 0.455532 & \bf 0.716436 & 0.555755 \\
DP-LK & 0.634324 & 0.584691 & 0.593063 && 0.426094 & 0.408868 & 0.354523 \\
DP-RK & 0.690697 & 0.533478 & 0.599916 && 0.385767 & 0.549764 & 0.451743 \\
Age-LK & 0.72898 & 0.667763 & 0.661767 && 0.484009 & 0.534213 & 0.490603 \\
Age-RK & 0.771952 & 0.670763 & 0.71628 && 0.499404 & 0.620618 & 0.551432 \\ \hline
\end{tabu}}
\end{center}
\label{tab:diabetesFeature}
\end{table*}

\begin{table*}[htbp]
\caption{Generalization Performance of different algorithms on Diabetes dataset.}
\begin{center}
{\tabulinesep=1.5mm                                           \begin{tabu}{ccccccccccr}
\hline
\multicolumn{ 2}{|c|}{\backslashbox{Methods$\downarrow$}{Data Size$\rightarrow$}}&10 & 20 & 30 & 40 & 50 & 60 & 70 & 80 & 90 \\ \hline
\multirow{ 2}{*}{Concat}& F+  &0.75 & 0.79 & 0.73 & 0.77 & 0.71 & 0.75 & 0.74 & 0.77 & 0.77 \\
\multicolumn{ 1}{c}{} & F-       	&0.44 & 0.57 & 0.5 & 0.53 & 0.5 & 0.49 & 0.54 & 0.55 & 0.44 \\ \hline
\multirow{2}{*}{F-EC}& F+    	&0.81 & 0.83 & 0.73 & 0.79 & 0.7 & 0.78 & 0.72 & 0.78 & 0.82 \\
\multicolumn{ 1}{c}{} & F-       	&0.38 & 0.52 & 0.5 & 0.5 & 0.36 & 0.34 & 0.49 & 0.48 & 0.38 \\ \hline
\multirow{2}{*}{SG-MKL}&F+    	&0.81 & 0.78 & 0.83 & 0.74 & 0.84 & 0.81 & 0.8 & 0.84 & 0.86 \\
\multicolumn{ 1}{c}{} & F-  	    	&0.69 & 0.57 & 0.73 & 0.54 & 0.73 & 0.58 & 0.63 & 0.69 & 0.73 \\ \hline
\multirow{2}{*}{L-MKL}& F+  	&0.7 & 0.7 & 0.72 & 0.63 & 0.74 & 0.72 & 0.7 & 0.77 & 0.78 \\
\multicolumn{ 1}{c}{} & F-  	    	&0.62 & 0.61 & 0.64 & 0.52 & 0.66 & 0.69 & 0.62 & 0.71 & 0.71 \\ \hline
\multirow{2}{*}{F-MKL}& F+   	&0.8 & 0.77 & 0.78 & 0.8 & 0.76 & 0.71 & 0.73 & 0.69 & 0.68 \\
\multicolumn{ 1}{c}{} & F- 	     	&0.8 & 0.8 & 0.79 & 0.77 & 0.78 & 0.79 & 0.74 & 0.76 & 0.71 \\ \hline
\multirow{2}{*}{S-MKL}& F+   	&0.8 & 0.78 & 0.78 & 0.78 & 0.77 & 0.79 & 0.73 & 0.72 & 0.7 \\
\multicolumn{ 1}{c}{} & F- 	     	&0.8 & 0.79 & 0.79 & 0.78 & 0.78 & 0.82 & 0.73 & 0.73 & 0.7 \\ \hline
\end{tabu}}
\end{center}
\label{tab:diabetesGen}
\end{table*}

\begin{table*}[htbp]
\caption{\small{The averages and standard deviations (in braces) of performances of different classifiers on Diabetes dataset when trained with $60\%$ of available data and the experiments are repeated $10$ times. }}
\begin{center}
{\tabulinesep=1.5mm                                           \begin{tabu}{ccccccccc}
\hline
\multicolumn{ 1}{c}{Methods $\downarrow$} & \multicolumn{ 3}{c}{Positive} & & \multicolumn{ 3}{c}{Negative} & \multicolumn{ 1}{c}{Support} \\ \cline{ 2- 4} \cline{6-8}
\multicolumn{ 1}{l}{} & \multicolumn{1}{l}{Precision} & \multicolumn{1}{l}{Recall} & \multicolumn{1}{l}{F-Measure} && \multicolumn{1}{l}{Precision} & \multicolumn{1}{l}{Recall} & \multicolumn{1}{l}{F-Measure} & \multicolumn{ 1}{c}{ Vectors} \\ \hline
CONCAT & \bf 0.81(0.104) & 0.69(0.0807) & 0.75(0.021) && 0.65(0.0391) & 0.39(0.0102) & 0.49(0.0038) & 0.62(0.0726) \\
F-EC & 0.74(0.1082) & 0.82(0.1101) & 0.78(0.019) && 0.49(0.0472) & 0.26(0.0313) & 0.34(0.0021) & 0.75(0.0923) \\
SGMKL & 0.77(0.0921) &  \bf0.85(0.0813) &  \bf0.81(0.0051) && 0.63(0.0932) & 0.53(0.0414) & 0.58(0.0019) & 0.52(0.0521) \\
L-MKL & 0.77(0.0332) & 0.67(0.0642) & 0.72(0.0701) && 0.67(0.01) & 0.71(0.0092) & 0.69(0.0101) & 0.49(0.0801) \\
F-MKL & 0.74(0.0192) & 0.68(0.0304) & 0.71(0.0109) &&  \bf0.81(0.0591) & 0.77(0.0012) & 0.79(0.081) & 0.45(0.091) \\
S-MKL & 0.79(0.0204) & 0.79(0.0028) & 0.79(0.0067) && 0.74(0.0134) &  \bf0.91(0.1009) &  \bf0.82(0.0091) & \bf 0.44(0.01) \\ \hline
\end{tabu}}
\end{center}
\label{tab:diabetesPerform}
\end{table*}

\clearpage
\subsection{German Numeric Dataset}
\label{subsec:german}
\begin{sidewaysfigure*}[htbp]
\centerline{\includegraphics[width=1\textheight]{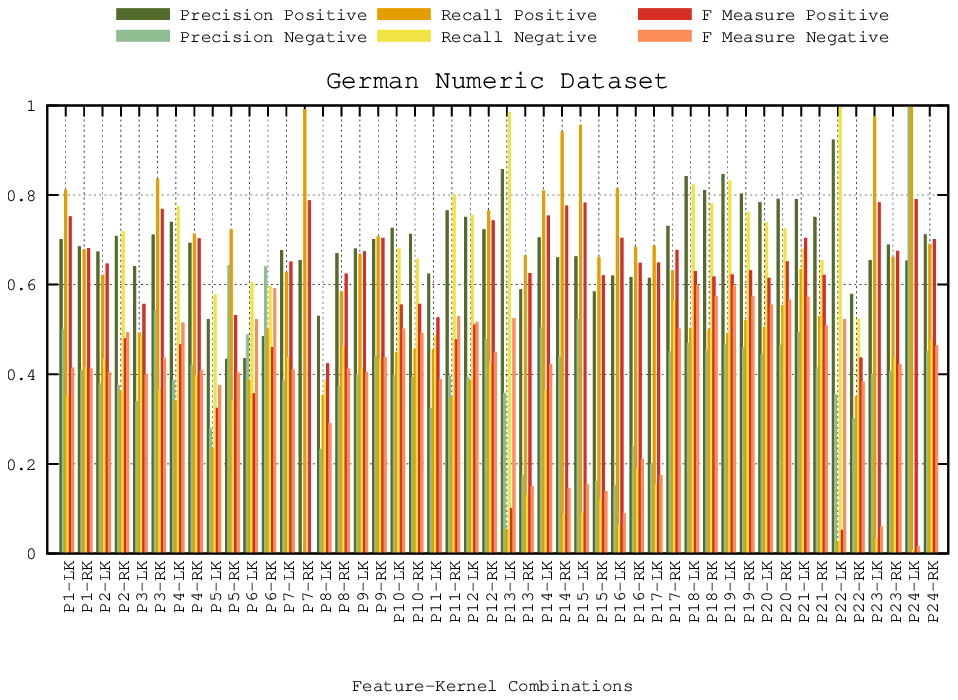}}
\caption{\small{Visualization of the performance analysis data presented in Table~\ref{tab:germanFeature}. The precision, recall and f-measures for different feature kernel combinations are shown for the German Numeric dataset.}}
\label{fig:germanFeature}
\end{sidewaysfigure*}

\begin{figure*}
\centerline{\includegraphics[width=0.5\textwidth]{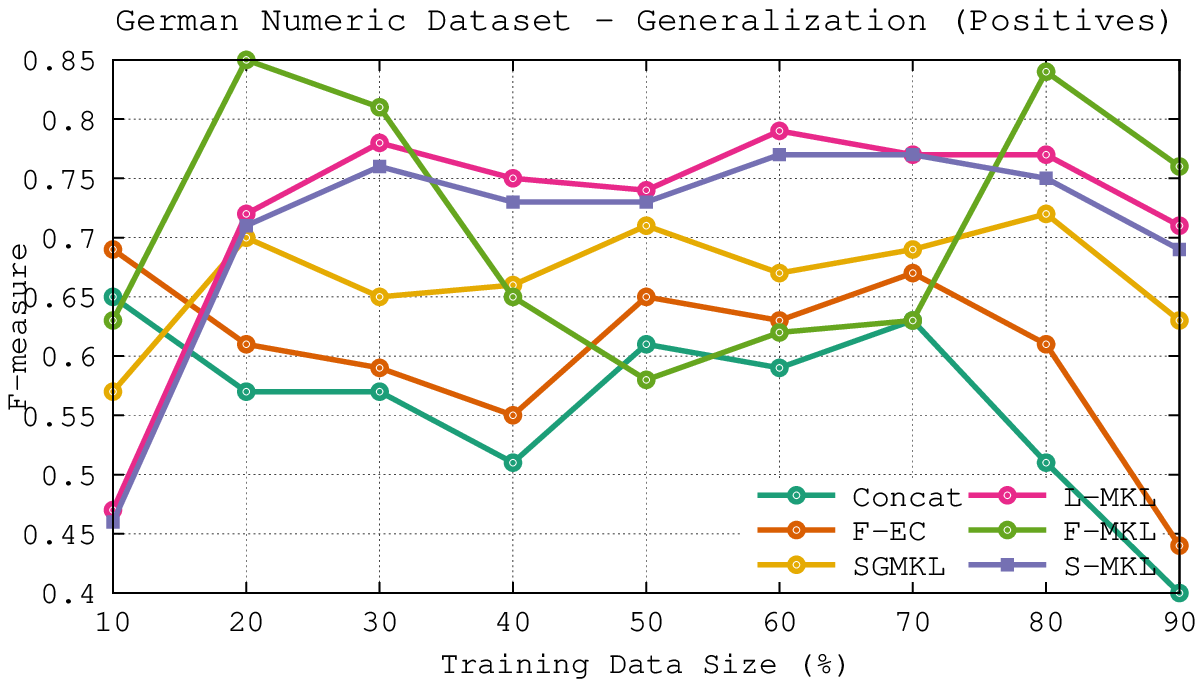}
\includegraphics[width=0.5\textwidth]{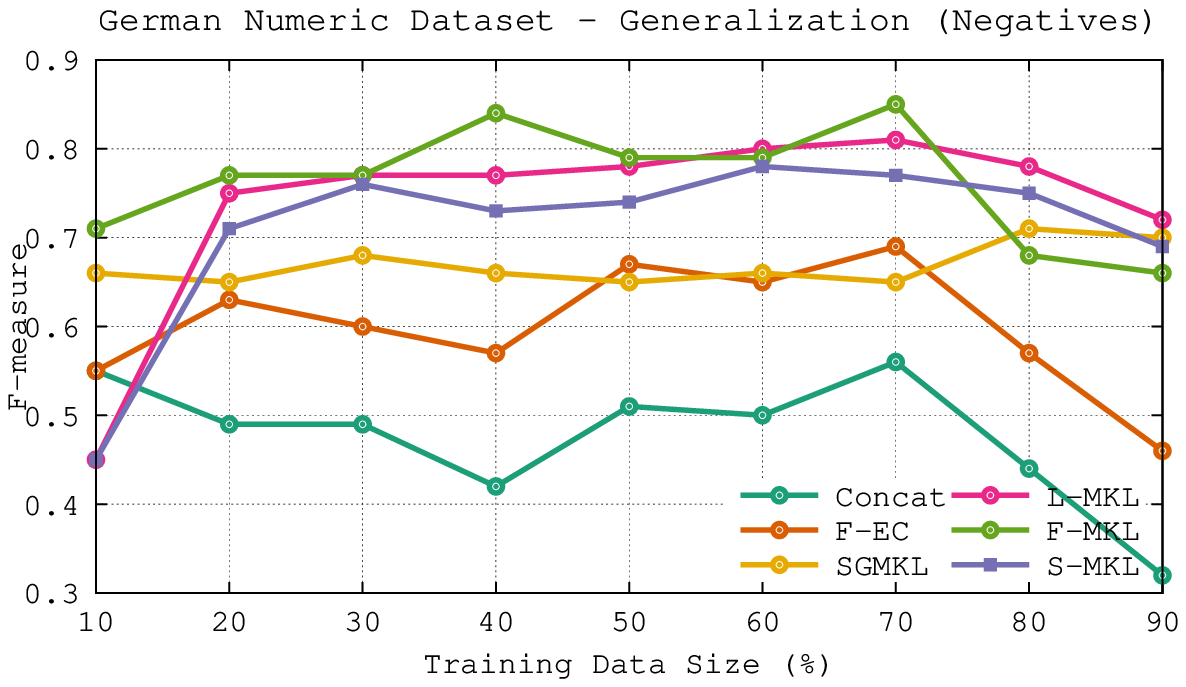}}
\caption{\small{Visualization of generalization performance data presented in Table~\ref{tab:germanGen}. The variations of f-measures for (a) positive and (b) negative categories are presented with respect to changing training set size.}}
\label{fig:germanGen}
\end{figure*}

\begin{figure*}[htbp]
\centerline{\includegraphics[width=1\textwidth]{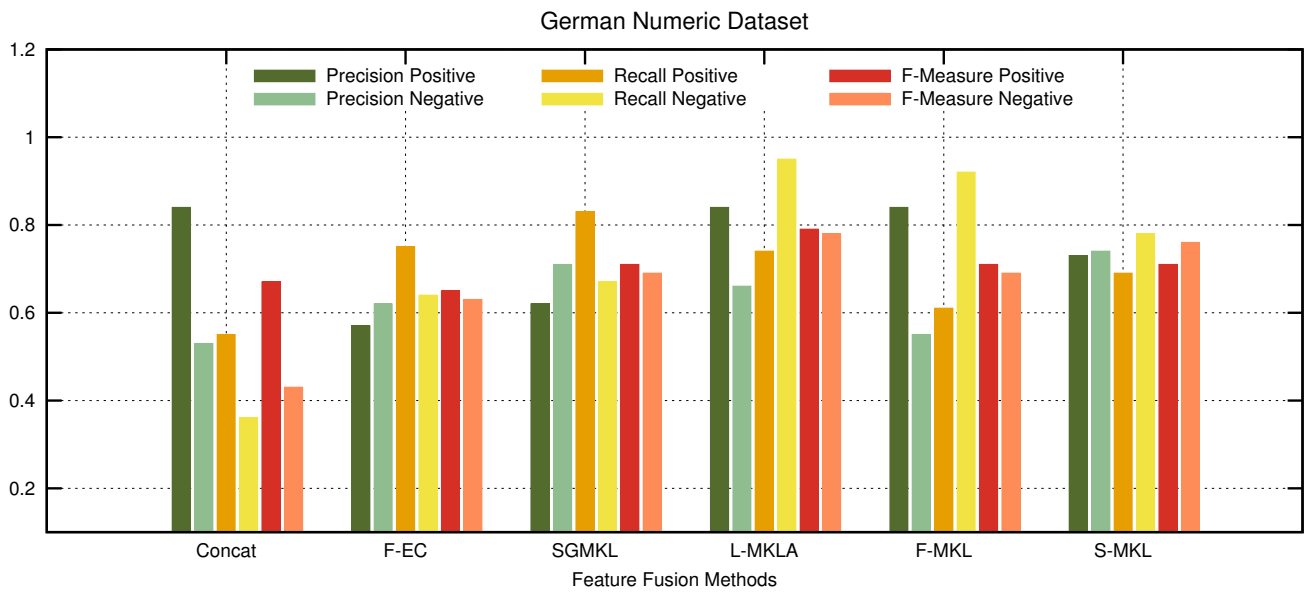}}
\caption{\small{Visualization of the performance analysis data presented in Table~\ref{tab:germanPerform}}}
\label{fig:germanPerform}
\end{figure*}

The German Numeric dataset consists of $1000$ samples with $30\%$ Positive sample. Each sample is represented by $24$ single continuous valued attributes, represented by P1 through P24. We have used Linear (LK) and RBF (RK) kernels with each attribute resulting in a total of $48$ feature-kernel combinations.  Performance of individual feature kernel combinations are tabulated in table~\ref{tab:germanFeature} and are visualized in figure~\ref{fig:germanFeature}. Table~\ref{tab:germanGen} and Figure~\ref{fig:germanGen} shows the Generalization  performance of different classifiers on Breast Cancer dataset while Table~\ref{tab:germanPerform} and Figure~\ref{fig:germanPerform} presents the detailed performance analysis of different classifiers when trained on $60\%$ of total available data.

\begin{table*}[htbp]
\caption{Feature performance Analysis of German Numeric dataset}
\begin{center}
{\tabulinesep=1.5mm                                           \begin{tabu}{cccccccc}
\hline

\multicolumn{1}{c}{Features} & \multicolumn{3}{c}{Positive}  & \multicolumn{3}{c}{Negative} \\ \cline{ 2-4} \cline{6-8}
&Precision&Recall&F Measure&&Precision&Recall&F Measure \\ \hline
P1-LK & 0.700576 & 0.811111 & 0.751802 && 0.5 & 0.352697 & 0.413625 \\
P1-RK & 0.684343 & 0.6775 & 0.680905 & &0.408257 & 0.415888 & 0.412037 \\
P2-LK & 0.67284 & 0.621083 & 0.645926 && 0.378505 & 0.433155 & 0.40399 \\
P2-RK & 0.707792 & 0.362126 & 0.479121 && 0.374593 & 0.71875 & 0.492505 \\
P3-LK & 0.640625 & 0.492 & 0.556561 && 0.338542 & 0.485075 & 0.398773 \\
P3-RK & 0.710638 & 0.835 & 0.767816 && 0.541667 & 0.364486 & 0.435754 \\
P4-LK & 0.73913 & 0.34 & 0.465753 && 0.385093 & 0.775 & 0.514523 \\
P4-RK & 0.692308 & 0.712871 & 0.702439 && 0.42 & 0.396226 & 0.407767 \\
P5-LK & 0.521739 & 0.235294 & 0.324324 && 0.277778 & 0.576923 & 0.375 \\
P5-RK & 0.432986 & 0.722473 & 0.531231 && 0.642164 & 0.340201 & 0.402715 \\
P6-LK & 0.434913 & 0.385872 & 0.356303 && 0.487472 & 0.604935 & 0.522403 \\
P6-RK & 0.484239 & 0.5004 & 0.459422 && 0.64 & 0.596141 & 0.590562 \\
P7-LK & 0.676259 & 0.626667 & 0.650519 && 0.384615 & 0.4375 & 0.409357 \\
P7-RK & 0.653595 & 0.990099 & 0.787402 && 0 & 0 & 0 \\
P8-LK & 0.529412 & 0.352941 & 0.423529 && 0.232558 & 0.384615 & 0.289855 \\
P8-RK & 0.669211 & 0.584444 & 0.623962 && 0.372483 & 0.460581 & 0.411874 \\
P9-LK & 0.679389 & 0.6675 & 0.673392 && 0.39819 & 0.411215 & 0.404598 \\
P9-RK & 0.700565 & 0.706553 & 0.703546 && 0.440217 & 0.433155 & 0.436658 \\
P10-LK & 0.725806 & 0.448505 & 0.554415 && 0.396364 & 0.68125 & 0.501149 \\
P10-RK & 0.7125 & 0.456 & 0.556098 && 0.392857 & 0.656716 & 0.49162 \\
P11-LK & 0.623288 & 0.455 & 0.526012 && 0.322981 & 0.485981 & 0.38806 \\
P11-RK & 0.764706 & 0.346667 & 0.477064 && 0.395062 & 0.8 & 0.528926 \\
P12-LK & 0.75 & 0.386139 & 0.509804 && 0.392157 & 0.754717 & 0.516129 \\
P12-RK & 0.722222 & 0.764706 & 0.742857 && 0.478261 & 0.423077 & 0.44898 \\
P13-LK & 0.857143 & 0.053333 & 0.100418 && 0.357466 & 0.983402 & 0.524336 \\
P13-RK & 0.588496 & 0.665 & 0.624413 && 0.17284 & 0.130841 & 0.148936 \\
P14-LK & 0.704715 & 0.809117 & 0.753316 && 0.503704 & 0.363636 & 0.42236 \\
P14-RK & 0.659674 & 0.940199 & 0.775342 && 0.4375 & 0.0875 & 0.145833 \\
P15-LK & 0.66205 & 0.956 & 0.782324 && 0.521739 & 0.089552 & 0.152866 \\
P15-RK & 0.584071 & 0.66 & 0.619718 && 0.160494 & 0.121495 & 0.138298 \\
P16-LK & 0.619289 & 0.813333 & 0.70317 && 0.151515 & 0.0625 & 0.088496 \\
P16-RK & 0.616071 & 0.683168 & 0.647887 && 0.238095 & 0.188679 & 0.210526 \\
P17-LK & 0.614035 & 0.686275 & 0.648148 && 0.2 & 0.153846 & 0.173913 \\
P17-RK & 0.730077 & 0.631111 & 0.676996 && 0.450331 & 0.564315 & 0.500921 \\
P18-LK & 0.841004 & 0.5025 & 0.629108 && 0.469333 & 0.82243 &  \bf0.597623 \\
P18-RK & 0.810185 & 0.498575 & 0.617284 && 0.453416 & 0.780749 & 0.573674 \\
P19-LK & 0.845714 & 0.491694 & 0.621849 && 0.465035 & 0.83125 & 0.596413 \\
P19-RK & 0.802469 & 0.52 & 0.631068 &&0.459459 & 0.761194 & 0.573034 \\
P20-LK & 0.782946 & 0.505 & 0.613982 && 0.44382 & 0.738318 & 0.554386 \\
P20-RK & 0.790476 & 0.553333 & 0.65098 && 0.464 & 0.725 & 0.565854 \\
P21-LK & 0.790123 & 0.633663 & 0.703297 && 0.493151 & 0.679245 & 0.571429 \\
P21-RK & 0.75 & 0.529412 & 0.62069 && 0.414634 & 0.653846 & 0.507463 \\
P22-LK &  \bf0.923077 & 0.026667 & 0.051836 && 0.353982 &  \bf0.995851 & 0.522307 \\
P22-RK & 0.578512 & 0.35 & 0.436137 && 0.301075 & 0.523364 & 0.382253 \\
P23-LK & 0.65392 & 0.974359 & 0.782609 && 0.4 & 0.032086 & 0.059406 \\
P23-RK & 0.688581 & 0.66113 & 0.674576 && 0.406977 & 0.4375 & 0.421687 \\
P24-LK & 0.652742 &  \bf1 &  \bf0.789889 &&  \bf1 & 0.007463 & 0.014815 \\
P24-RK & 0.71134 & 0.69 & 0.700508 && 0.451327 & 0.476636 & 0.463636 \\ \hline
\end{tabu}}
\end{center}
\label{tab:germanFeature}
\end{table*}

\begin{table*}[htbp]
\caption{Generalization Performance of different algorithms on German Numeric dataset.}
\begin{center}
{\tabulinesep=1.5mm                                           \begin{tabu}{ccccccccccr}
\hline
\multicolumn{ 2}{|c|}{\backslashbox{Methods$\downarrow$}{Data Size$\rightarrow$}}&10 & 20 & 30 & 40 & 50 & 60 & 70 & 80 & 90 \\ \hline
\multirow{ 2}{*}{Concat}& F+  &0.65 & 0.57 & 0.57 & 0.51 & 0.61 & 0.67 & 0.63 & 0.51 & 0.4 \\
\multicolumn{ 1}{c}{} & F-     	&0.55 & 0.49 & 0.49 & 0.42 & 0.51 & 0.43 & 0.56 & 0.44 & 0.32 \\
\multirow{2}{*}{F-EC}& F+  	&0.69 & 0.61 & 0.59 & 0.55 & 0.65 & 0.65 & 0.67 & 0.61 & 0.44 \\
\multicolumn{ 1}{c}{} & F-     	&0.55 & 0.63 & 0.6 & 0.57 & 0.67 & 0.63 & 0.69 & 0.57 & 0.46 \\
\multirow{2}{*}{SG-MKL}&F+ 	&0.57 & 0.7 & 0.65 & 0.66 & 0.71 & 0.71 & 0.69 & 0.72 & 0.63 \\
\multicolumn{ 1}{c}{} & F-  		&0.66 & 0.65 & 0.68 & 0.66 & 0.65 & 0.69 & 0.65 & 0.71 & 0.7 \\
\multirow{2}{*}{L-MKL}& F+ 	&0.47 & 0.72 & 0.78 & 0.75 & 0.74 & 0.79 & 0.77 & 0.77 & 0.71 \\
\multicolumn{ 1}{c}{} & F-  		&0.45 & 0.75 & 0.77 & 0.77 & 0.78 & 0.78 & 0.81 & 0.78 & 0.72 \\
\multirow{2}{*}{F-MKL}& F+ 	&0.63 & 0.85 & 0.81 & 0.65 & 0.58 & 0.71 & 0.63 & 0.84 & 0.76 \\
\multicolumn{ 1}{c}{} & F- 		&0.71 & 0.77 & 0.77 & 0.84 & 0.79 & 0.69 & 0.85 & 0.68 & 0.66 \\
\multirow{2}{*}{S-MKL}& F+ 	&0.46 & 0.71 & 0.76 & 0.73 & 0.73 & 0.71 & 0.77 & 0.75 & 0.69 \\
\multicolumn{ 1}{c}{} & F- 		&0.45 & 0.71 & 0.76 & 0.73 & 0.74 & 0.76 & 0.77 & 0.75 & 0.69 \\ \hline
\end{tabu}}
\end{center}
\label{tab:germanGen}
\end{table*}

\begin{table*}[htbp]
\caption{\small{The averages and standard deviations (in braces) of performances of different classifiers on German Numeric dataset when trained with $60\%$ of available data and the experiments are repeated $10$ times. }}
\begin{center}
{\tabulinesep=1.5mm                                           \begin{tabu}{ccccccccc}
\hline
\multicolumn{ 1}{c}{Methods $\downarrow$} & \multicolumn{ 3}{c}{Positive} & & \multicolumn{ 3}{c}{Negative} & \multicolumn{ 1}{c}{Support} \\ \cline{ 2- 4} \cline{6-8}
\multicolumn{ 1}{l}{} & \multicolumn{1}{l}{Precision} & \multicolumn{1}{l}{Recall} & \multicolumn{1}{l}{F-Measure} && \multicolumn{1}{l}{Precision} & \multicolumn{1}{l}{Recall} & \multicolumn{1}{l}{F-Measure} & \multicolumn{ 1}{c}{ Vectors} \\ \hline
CONCAT &  \bf0.84(0.0018) & 0.55(0.0541) & 0.67(0.0010) && 0.53(0.0221) & 0.36(0.0871) & 0.43(0.0008) & 0.77(0.0001) \\
F-EC & 0.57(0.0203) & 0.75(0.0312) & 0.65(0.0923) && 0.62(0.0412) & 0.64(0.0331) & 0.63(0.021) & 0.79(0.0001) \\
SGMKL & 0.62(0.0720) & \bf 0.83(0.0423) & 0.71(0.052) && 0.71(0.1840) & 0.67(0.0206) & 0.69(0.01) & 0.62(0.0762) \\
L-MKL & 0.84(0.0607) & 0.74(0.0156) &  \bf0.79(0.0081) && 0.66(0.0672) &  \bf0.95(0.0413) & \bf 0.78(0.0064) &  \bf0.49(0.0023) \\
F-MKL & 0.84(0.0191) & 0.61(0.3093) & 0.71(0.0005) && 0.55(0.0550) & 0.92(0.0641) & 0.69(0.0093) & 0.52(0.0289) \\
S-MKL & 0.73(0.0410) & 0.69(0.0097) & 0.71(0.0053) &&  \bf0.74(0.0030) & 0.78(0.0085) & 0.76(0.0054) & 0.67(0.0431) \\ \hline
\end{tabu}}
\end{center}
\label{tab:germanPerform}
\end{table*}

\clearpage
\subsection{Mushroom Dataset}
\label{subsec:mushroom}
\begin{sidewaysfigure*}[htbp]
\centerline{\includegraphics[width=1\textheight]{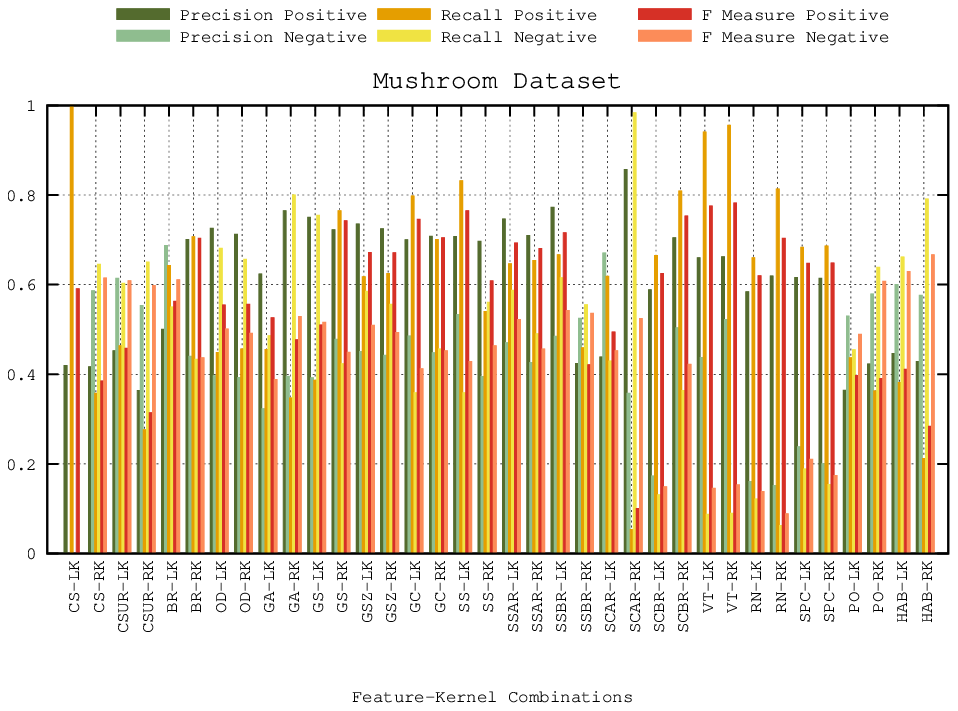}}
\caption{\small{Visualization of the performance analysis data presented in Table~\ref{tab:mushroomFeature}. The precision, recall and f-measures for different feature kernel combinations are shown for the Mushroom dataset.}}
\label{fig:mushroomFeature}
\end{sidewaysfigure*}

\begin{figure*}
\centerline{\includegraphics[width=0.5\textwidth]{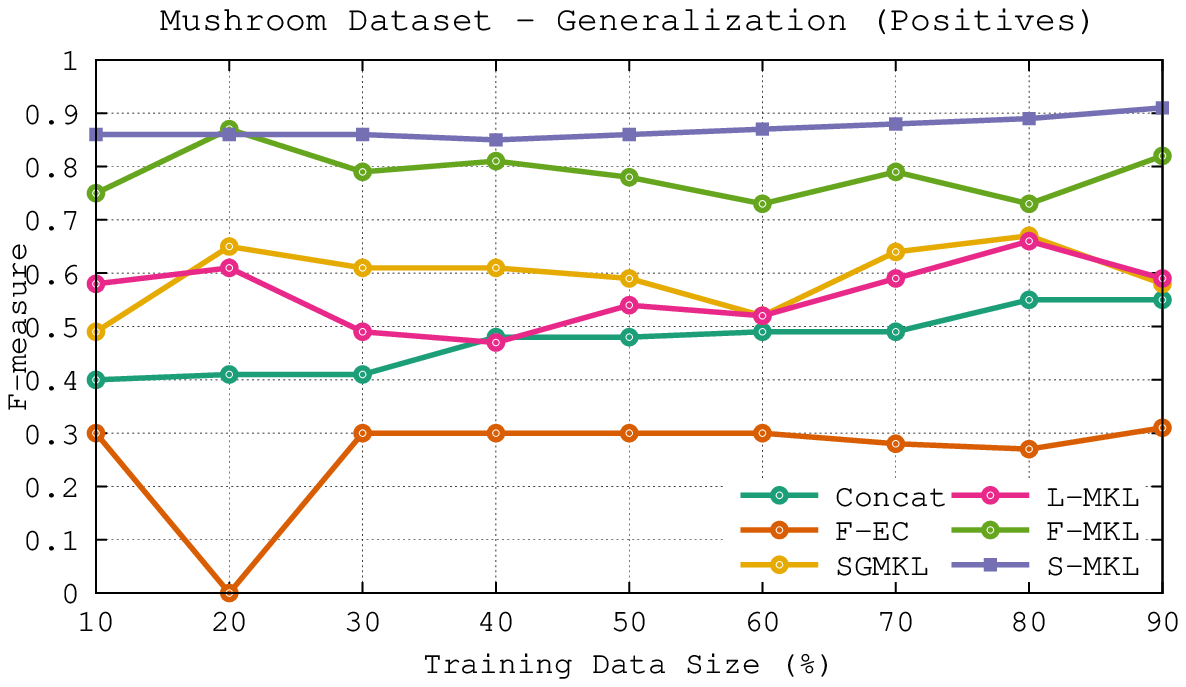}
\includegraphics[width=0.5\textwidth]{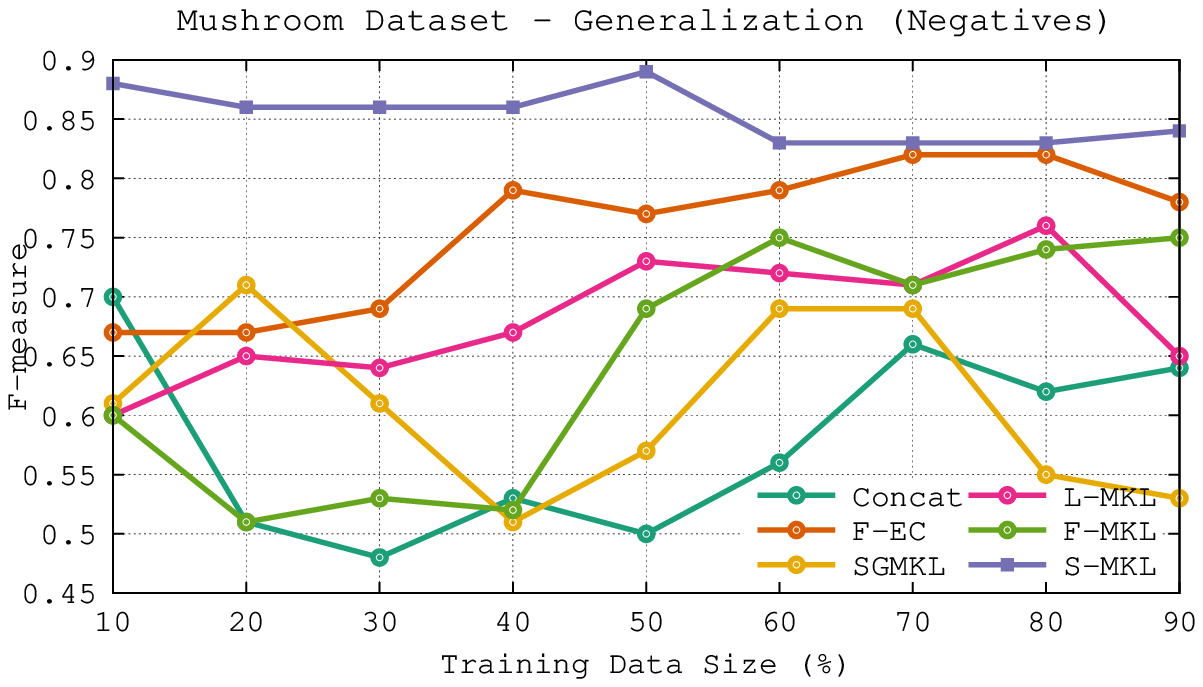}}
\caption{\small{Visualization of generalization performance data presented in Table~\ref{tab:mushroomGen}. The variations of f-measures for (a) positive and (b) negative categories are presented with respect to changing training set size.}}
\label{fig:mushroomGen}
\end{figure*}

\begin{figure*}[htbp]
\centerline{\includegraphics[width=1\textwidth]{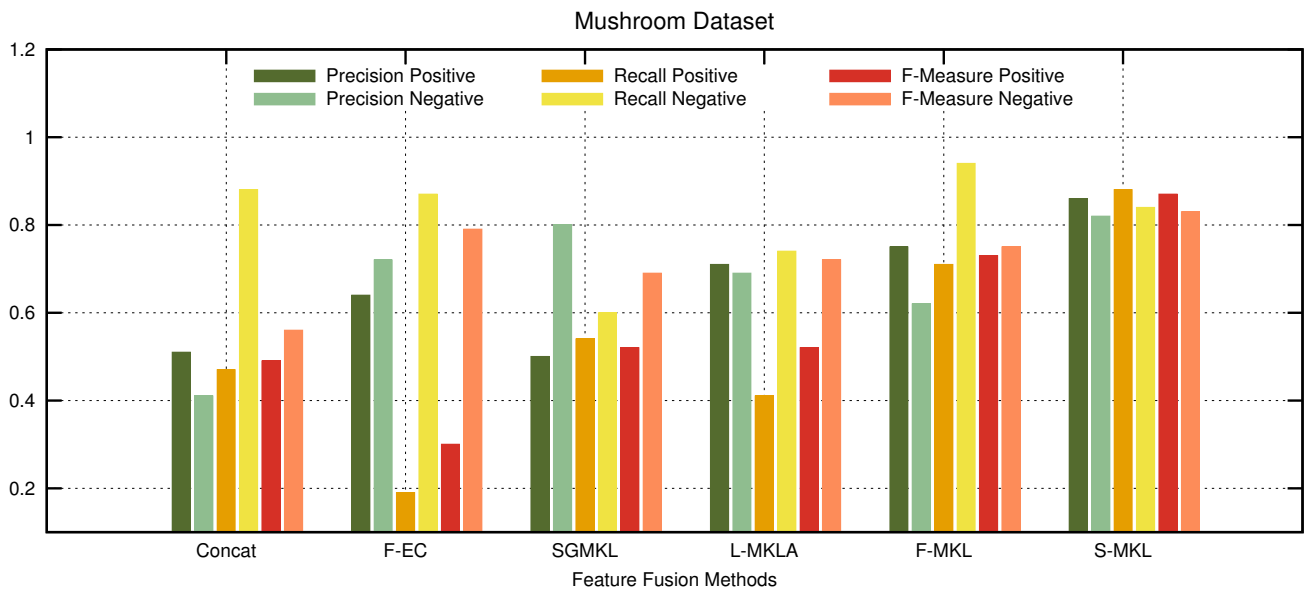}}
\caption{\small{Visualization of the performance analysis data presented in Table~\ref{tab:mushroomPerform}}}
\label{fig:mushroomPerform}
\end{figure*}

The Mushroom dataset consists of $8124$ samples with $64.1\%$ Positive sample. Each sample is represented by $123$ binary values representing $21$ different attributes viz.-- cap-shape(CS), cap-surface(CSUR),bruises(BR), odor(OD), gill-attachment(GA), gill-spacing(GS),gill-size(GSZ),gill-color(GC), stalk-shape(SS), stalk-surface-above-ring(SSAR),stalk-surface-below-ring(SSBR), stalk-color-above-ring(SCAR), stalk-color-below-ring(SCBR), veil-type(VT), veil-color(VC), ring-number(RN), ring-type(RT), spore-print-color(SPC), population(PO) and habitat(HAB).
We have used Linear (LK) and RBF (RK) kernels with each attribute resulting in a total of $42$ feature-kernel combinations.  Performance of individual feature kernel combinations are tabulated in table~\ref{tab:mushroomFeature} and are visualized in figure~\ref{fig:mushroomFeature}. Table~\ref{tab:mushroomGen} and Figure~\ref{fig:mushroomGen} shows the Generalization  performance of different classifiers on Mushroom dataset while Table~\ref{tab:mushroomPerform} and Figure~\ref{fig:mushroomPerform} presents the detailed performance analysis of different classifiers when trained on $60\%$ of total available data.
\begin{table*}[htbp]
\caption{Feature performance Analysis of Mushroom dataset}
\begin{center}
{\tabulinesep=1.5mm                                           \begin{tabu}{cccccccc}
\hline

\multicolumn{1}{c}{Features} & \multicolumn{3}{c}{Positive}  & \multicolumn{3}{c}{Negative} \\ \cline{ 2-4} \cline{6-8}
&Precision&Recall&F Measure&&Precision&Recall&F Measure \\ \hline
CS-LK & 0.419162 & \bf 1 & 0.590717 && 0 & 0 & 0 \\
CS-RK & 0.416667 & 0.357143 & 0.384615 && 0.586207 & 0.64557 & 0.614458 \\
CSUR-LK & 0.452381 & 0.463415 & 0.457831 && 0.614035 & 0.603448 & 0.608696 \\
CSUR-RK & 0.363636 & 0.275862 & 0.313725 && 0.553191 & 0.65 & 0.597701 \\
BR-LK & 0.5 & 0.642857 & 0.5625 & &0.6875 & 0.55 & 0.611111 \\
BR-RK & 0.700565 & 0.706553 & 0.703546 && 0.440217 & 0.433155 & 0.436658 \\
OD-LK & 0.725806 & 0.448505 & 0.554415 && 0.396364 & 0.68125 & 0.501149 \\
OD-RK & 0.7125 & 0.456 & 0.556098 && 0.392857 & 0.656716 & 0.49162 \\
GA-LK & 0.623288 & 0.455 & 0.526012 && 0.322981 & 0.485981 & 0.38806 \\
GA-RK & 0.764706 & 0.346667 & 0.477064 && 0.395062 & 0.8 & 0.528926 \\
GS-LK & 0.75 & 0.386139 & 0.509804 && 0.392157 & 0.754717 & 0.516129 \\
GS-RK & 0.722222 & 0.764706 & 0.742857 && 0.478261 & 0.423077 & 0.44898 \\
GSZ-LK & 0.73545 & 0.617778 & 0.671498 && 0.450479 & 0.585062 & 0.509025 \\
GSZ-RK & 0.724638 & 0.625 & 0.671141 && 0.442379 & 0.556075 & 0.492754 \\
GC-LK & 0.7 & 0.797721 & 0.745672& & 0.485507 & 0.358289 & 0.412308 \\
GC-RK & 0.708054 & 0.700997 & 0.704508 && 0.447853 & 0.45625 & 0.452012 \\
SS-LK & 0.707483 & 0.832 & 0.764706 && 0.533333 & 0.358209 & 0.428571 \\
SS-RK & 0.696774 & 0.54 & 0.608451 && 0.394737 & 0.560748 & 0.46332 \\
SSAR-LK & 0.746154 & 0.646667 & 0.692857 && 0.47 & 0.5875 & 0.522222 \\
SSAR-RK & 0.709677 & 0.653465 & 0.680412 && 0.42623 & 0.490566 & 0.45614 \\
SSBR-LK & 0.772727 & 0.666667 & 0.715789 && 0.484848 & 0.615385 & 0.542373 \\
SSBR-RK & 0.42368 & 0.459092 & 0.421284 && 0.52507 & 0.555038 & 0.536074 \\
SCAR-LK & 0.438128 & 0.618804 & 0.494224 &&  \bf0.670361 & 0.429232 & 0.452681 \\
SCAR-RK &  \bf0.857143 & 0.053333 & 0.100418 && 0.357466 &  \bf0.983402 & 0.524336 \\
SCBR-LK & 0.588496 & 0.665 & 0.624413& & 0.17284 & 0.130841 & 0.148936 \\
SCBR-RK & 0.704715 & 0.809117 & 0.753316 && 0.503704 & 0.363636 & 0.42236 \\
VT-LK & 0.659674 & 0.940199 & 0.775342 && 0.4375 & 0.0875 & 0.145833 \\
VT-RK & 0.66205 & 0.956 & \bf 0.782324 && 0.521739 & 0.089552 & 0.152866 \\
RN-LK & 0.584071 & 0.66 & 0.619718 && 0.160494 & 0.121495 & 0.138298 \\
RN-RK & 0.619289 & 0.813333 & 0.70317 && 0.151515 & 0.0625 & 0.088496 \\
SPC-LK & 0.616071 & 0.683168 & 0.647887 && 0.238095 & 0.188679 & 0.210526 \\
SPC-RK & 0.614035 & 0.686275 & 0.648148 && 0.2 & 0.153846 & 0.173913 \\
PO-LK & 0.364238 & 0.436508 & 0.397112 && 0.529801 & 0.454545 & 0.489297 \\
PO-RK & 0.42268 & 0.362832 & 0.390476 && 0.578947 & 0.63871 & 0.607362 \\
HAB-LK & 0.445783 & 0.381443 & 0.411111 && 0.6 & 0.661765 & 0.629371 \\
HAB-RK & 0.428571 & 0.211765 & 0.283465 && 0.575949 & 0.791304 &  \bf0.666667 \\ \hline
\end{tabu}}
\end{center}
\label{tab:mushroomFeature}
\end{table*}

\begin{table*}[htbp]
\caption{Generalization Performance of different algorithms on Mushroom dataset.}
\begin{center}
{\tabulinesep=1.5mm                                           \begin{tabu}{ccccccccccr}
\hline
\multicolumn{ 2}{|c|}{\backslashbox{Methods$\downarrow$}{Data Size$\rightarrow$}}&10 & 20 & 30 & 40 & 50 & 60 & 70 & 80 & 90 \\ \hline
\multirow{ 2}{*}{Concat}& F+  &0.4 & 0.41 & 0.41 & 0.48 & 0.48 & 0.49 & 0.49 & 0.55 & 0.55 \\
\multicolumn{ 1}{c}{} & F-     	&0.7 & 0.51 & 0.48 & 0.53 & 0.5 & 0.56 & 0.66 & 0.62 & 0.64 \\ \hline
\multirow{2}{*}{F-EC}& F+  	&0.3 & 0 & 0.3 & 0.3 & 0.3 & 0.3 & 0.28 & 0.27 & 0.31 \\
\multicolumn{ 1}{c}{} & F-     	&0.67 & 0.67 & 0.69 & 0.79 & 0.77 & 0.79 & 0.82 & 0.82 & 0.78 \\ \hline
\multirow{2}{*}{SG-MKL}&F+ 	&0.49 & 0.65 & 0.61 & 0.61 & 0.59 & 0.52 & 0.64 & 0.67 & 0.58 \\
\multicolumn{ 1}{c}{} & F-  		&0.61 & 0.71 & 0.61 & 0.51 & 0.57 & 0.69 & 0.69 & 0.55 & 0.53 \\ \hline
\multirow{2}{*}{L-MKL}& F+ 	&0.58 & 0.61 & 0.49 & 0.47 & 0.54 & 0.52 & 0.59 & 0.66 & 0.59 \\
\multicolumn{ 1}{c}{} & F-  		&0.6 & 0.65 & 0.64 & 0.67 & 0.73 & 0.72 & 0.71 & 0.76 & 0.65 \\ \hline
\multirow{2}{*}{F-MKL}& F+ 	&0.75 & 0.87 & 0.79 & 0.81 & 0.78 & 0.73 & 0.79 & 0.73 & 0.82 \\
\multicolumn{ 1}{c}{} & F- 		&0.6 & 0.51 & 0.53 & 0.52 & 0.69 & 0.75 & 0.71 & 0.74 & 0.75 \\ \hline
\multirow{2}{*}{S-MKL}& F+ 	&0.86 & 0.86 & 0.86 & 0.85 & 0.86 & 0.87 & 0.88 & 0.89 & 0.91 \\
\multicolumn{ 1}{c}{} & F- 		&0.88 & 0.86 & 0.86 & 0.86 & 0.89 & 0.83 & 0.83 & 0.83 & 0.84 \\ \hline
\end{tabu}}
\end{center}
\label{tab:mushroomGen}
\end{table*}

\begin{table*}[htbp]
\caption{\small{The averages and standard deviations (in braces) of performances of different classifiers on Mushroom dataset when trained with $60\%$ of available data and the experiments are repeated $10$ times.}}
\begin{center}
{\tabulinesep=1.5mm                                           \begin{tabu}{ccccccccc}
\hline
\multicolumn{ 1}{c}{Methods $\downarrow$} & \multicolumn{ 3}{c}{Positive} & & \multicolumn{ 3}{c}{Negative} & \multicolumn{ 1}{c}{Support} \\ \cline{ 2- 4} \cline{6-8}
\multicolumn{ 1}{l}{} & \multicolumn{1}{l}{Precision} & \multicolumn{1}{l}{Recall} & \multicolumn{1}{l}{F-Measure} && \multicolumn{1}{l}{Precision} & \multicolumn{1}{l}{Recall} & \multicolumn{1}{l}{F-Measure} & \multicolumn{ 1}{c}{ Vectors} \\ \hline
CONCAT & 0.51(0.0321) & 0.47(0.0084) & 0.49(0.0076) && 0.41(0.084) & 0.88(0.019) & 0.56(0.081) & 0.73(0.0003) \\
F-EC & 0.64(0.0121) & 0.19(0.374) & 0.3(0.0046) && 0.72(0.0421) & 0.87(0.0048) & 0.79(0.023) & 0.68(0.0013) \\
SGMKL & 0.5(0.0103) & 0.54(0.0178) & 0.52(0.046) && 0.8(0.0508) & 0.6(0.0045) & 0.69(0.0234) &  \bf0.6(0.081) \\
L-MKL & 0.71(0.0003) & 0.41(0.106) & 0.52(0.083) && 0.69(0.0014) & 0.74(0.0059) & 0.72(0.0001) & 0.78(0.0921) \\
F-MKL & 0.75(0.027) & 0.71(0.0068) & 0.73(0.0131) && 0.62(0.0451) &  \bf0.94(0.0011) & 0.75(0.01) & 0.62(0.012) \\
S-MKL &  \bf0.86(0.033) & \bf 0.88(0.0161) &  \bf0.87(0.029) &&  \bf0.82(0.0154) & 0.84(0.0021) &  \bf0.83(0.059) & 0.7(0.1098) \\ \hline
\end{tabu}}
\end{center}
\label{tab:mushroomPerform}
\end{table*}

\clearpage
\subsection{COD-RNA Dataset}
\label{subsec:cod}
\begin{sidewaysfigure*}[htbp]
\centerline{\includegraphics[width=1\textheight]{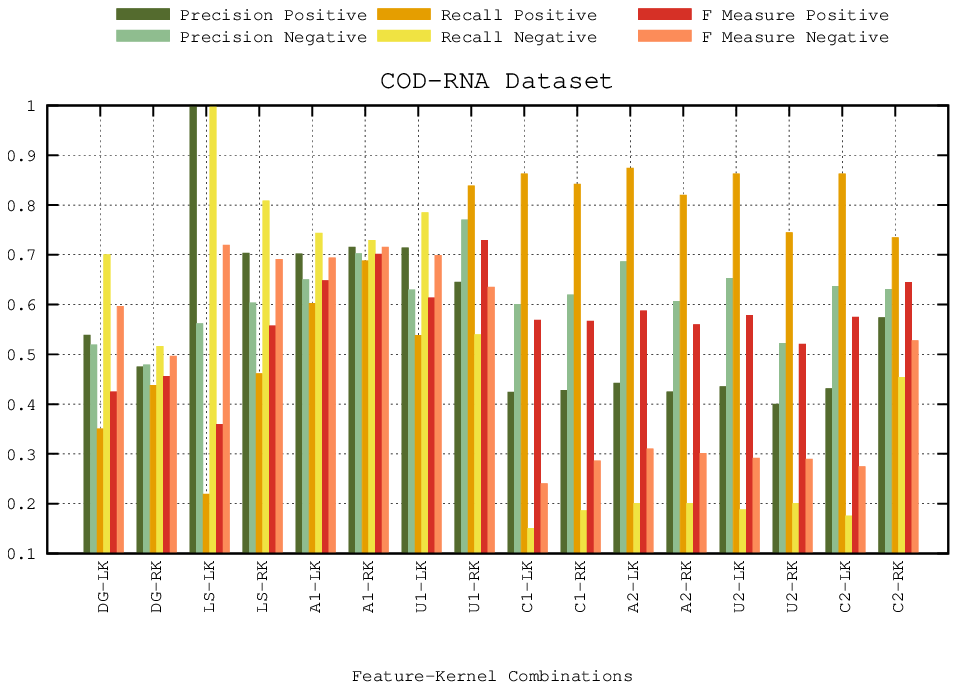}}
\caption{\small{Visualization of the performance analysis data presented in Table~\ref{tab:codFeature}. The precision, recall and f-measures for different feature kernel combinations are shown for the COD-RNA dataset.}}
\label{fig:codFeature}
\end{sidewaysfigure*}

\begin{figure*}
\centerline{\includegraphics[width=0.5\textwidth]{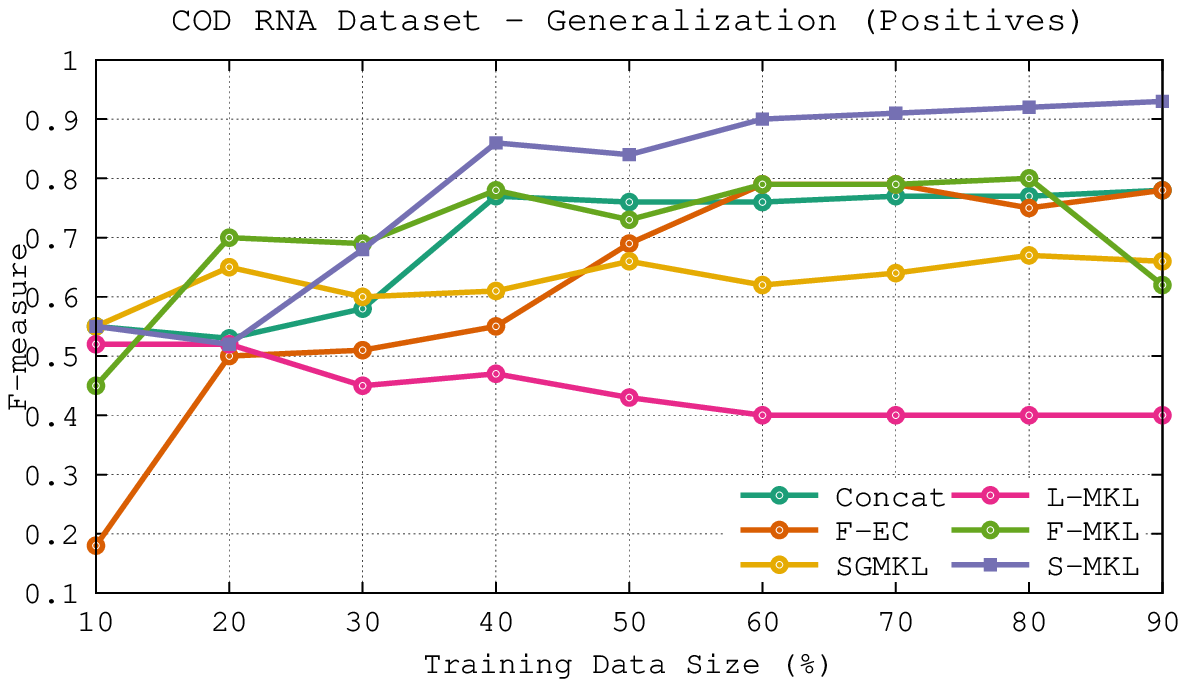}
\includegraphics[width=0.5\textwidth]{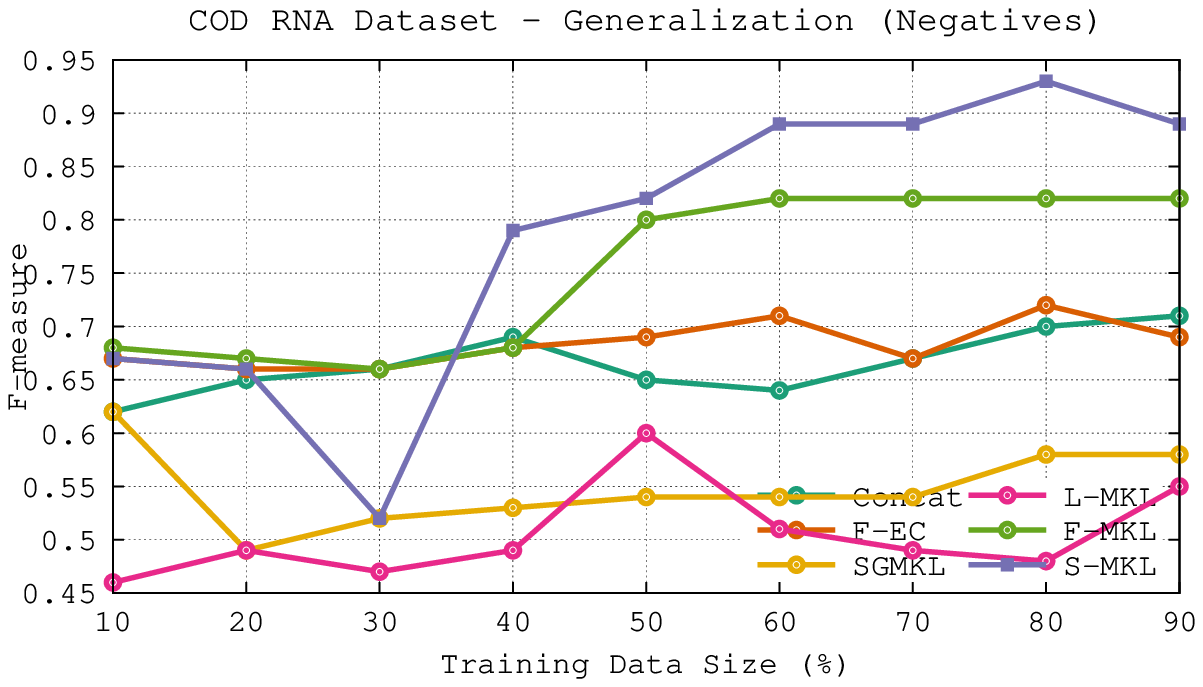}}
\caption{\small{Visualization of generalization performance data presented in Table~\ref{tab:codGen}. The variations of f-measures for (a) positive and (b) negative categories are presented with respect to changing training set size.}}
\label{fig:codGen}
\end{figure*}

\begin{figure*}[htbp]
\centerline{\includegraphics[width=1\textwidth]{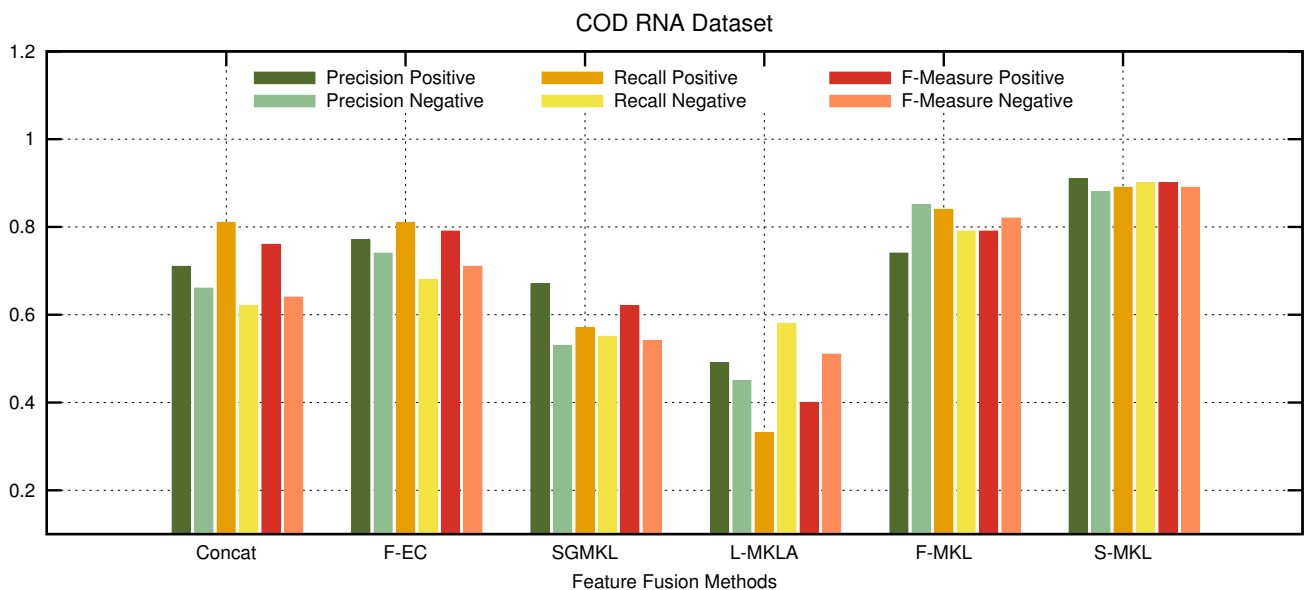}}
\caption{\small{Visualization of the performance analysis data presented in Table~\ref{tab:codPerform}}}
\label{fig:codPerform}
\end{figure*}

The COD-RNA dataset consists of $244109$ samples with $33.33\%$ Positive sample. Each sample is represented by $8$ single continuous valued attributes viz. -- Divide by 10 to get deltaG total value computed by the Dynalign algorithm (DG),  The length of shorter sequence(LS), 'A' frequencies of sequence 1(A1),'U' frequencies of sequence 1(U1),'C' frequencies of sequence 1(C1),'A' frequencies of sequence 2(A2), 'U' frequencies of sequence 2(U2), and 'C' frequencies of sequence 2(C2). We have used Linear (LK) and RBF (RK) kernels with each attribute resulting in a total of $16$ feature-kernel combinations.  Performance of individual feature kernel combinations are tabulated in table~\ref{tab:germanFeature} and are visualized in figure~\ref{fig:codFeature}. Table~\ref{tab:codGen} and Figure~\ref{fig:codGen} shows the Generalization  performance of different classifiers on COD-RNA dataset while Table~\ref{tab:codPerform} and Figure~\ref{fig:codPerform} presents the detailed performance analysis of different classifiers when trained on $60\%$ of total available data.
\begin{table*}[htbp]
\caption{Feature performance Analysis of COD-RNA dataset}
\begin{center}
{\tabulinesep=1.5mm                                           \begin{tabu}{cccccccc}
\hline

\multicolumn{1}{c}{Features} & \multicolumn{3}{c}{Positive}  & \multicolumn{3}{c}{Negative} \\ \cline{ 2-4} \cline{6-8}
&Precision&Recall&F Measure&&Precision&Recall&F Measure \\ \hline
DG-LK & 0.538462 & 0.35 & 0.424242 && 0.518519 & 0.7 & 0.595745 \\
DG-RK & 0.474576 & 0.4375 & 0.455285 && 0.478261 & 0.515625 & 0.496241 \\
LS-LK & 1 & 0.21875 & 0.358974 && 0.561404 & \bf 1 &  \bf0.719101 \\
LS-RK & 0.70297 & 0.461039 & 0.556863 && 0.602871 &  0.807692 & 0.690411 \\
A1-LK & 0.701657 & 0.601896 & 0.647959 && 0.65 & 0.742857 & 0.693333 \\
A1-RK &  \bf0.715 & 0.6875 & 0.70098 && 0.701835 & 0.728571 & 0.714953 \\
U1-LK & 0.713755 & 0.537815 & 0.613419 && 0.629213 & 0.784314 & 0.698254 \\
U1-RK & 0.64467 & 0.838284 &  \bf0.728838 &&  \bf0.769953 & 0.539474 & 0.634429 \\
C1-LK & 0.423729 & 0.862069 & 0.568182 && 0.6 & 0.15 & 0.24 \\
C1-RK & 0.427136 & 0.841584 & 0.566667 && 0.619048 & 0.185714 & 0.285714 \\
A2-LK & 0.44186 &  \bf0.873563 & 0.586873 && 0.685714 & 0.2 & 0.309677 \\
A2-RK & 0.42446 & 0.819444 & 0.559242 && 0.606061 & 0.2 & 0.300752 \\
U2-LK & 0.434783 & 0.862069 & 0.578035 && 0.652174 & 0.1875 & 0.291262 \\
U2-RK & 0.4 & 0.744186 & 0.520325 && 0.521739 & 0.2 & 0.289157 \\
C2-LK & 0.431034 & 0.862069 & 0.574713 && 0.636364 & 0.175 & 0.27451 \\
C2-RK & 0.573171 & 0.734375 & 0.643836 && 0.630435 & 0.453125 & 0.527273 \\ \hline
\end{tabu}}
\end{center}
\label{tab:codFeature}
\end{table*}

\begin{table*}[htbp]
\caption{Generalization Performance of different algorithms on COD-RNA dataset.}
\begin{center}
{\tabulinesep=1.5mm                                           \begin{tabu}{ccccccccccr}
\hline
\multicolumn{ 2}{|c|}{\backslashbox{Methods$\downarrow$}{Data Size$\rightarrow$}}&10 & 20 & 30 & 40 & 50 & 60 & 70 & 80 & 90 \\ \hline
\multirow{ 2}{*}{Concat}& F+  &0.55 & 0.53 & 0.58 & 0.77 & 0.76 & 0.76 & 0.77 & 0.77 & 0.78 \\
\multicolumn{ 1}{c}{} & F-     	&0.62 & 0.65 & 0.66 & 0.69 & 0.65 & 0.64 & 0.67 & 0.7 & 0.71 \\ \hline
\multirow{2}{*}{F-EC}& F+  	&0.18 & 0.5 & 0.51 & 0.55 & 0.69 & 0.79 & 0.79 & 0.75 & 0.78 \\
\multicolumn{ 1}{c}{} & F-     	&0.67 & 0.66 & 0.66 & 0.68 & 0.69 & 0.71 & 0.67 & 0.72 & 0.69 \\ \hline
\multirow{2}{*}{SG-MKL}&F+ 	&0.55 & 0.65 & 0.6 & 0.61 & 0.66 & 0.62 & 0.64 & 0.67 & 0.66 \\
\multicolumn{ 1}{c}{} & F-  		&0.62 & 0.49 & 0.52 & 0.53 & 0.54 & 0.54 & 0.54 & 0.58 & 0.58 \\ \hline
\multirow{2}{*}{L-MKL}& F+ 	&0.52 & 0.52 & 0.45 & 0.47 & 0.43 & 0.4 & 0.4 & 0.4 & 0.4 \\
\multicolumn{ 1}{c}{} & F-  		&0.46 & 0.49 & 0.47 & 0.49 & 0.6 & 0.51 & 0.49 & 0.48 & 0.55 \\ \hline
\multirow{2}{*}{F-MKL}& F+ 	&0.45 & 0.7 & 0.69 & 0.78 & 0.73 & 0.79 & 0.79 & 0.8 & 0.62 \\
\multicolumn{ 1}{c}{} & F- 		&0.68 & 0.67 & 0.66 & 0.68 & 0.8 & 0.82 & 0.82 & 0.82 & 0.82 \\ \hline
\multirow{2}{*}{S-MKL}& F+ 	&0.55 & 0.52 & 0.68 & 0.86 & 0.84 & 0.9 & 0.91 & 0.92 & 0.93 \\
\multicolumn{ 1}{c}{} & F- 		&0.67 & 0.66 & 0.52 & 0.79 & 0.82 & 0.89 & 0.89 & 0.93 & 0.89 \\ \hline
\end{tabu}}
\end{center}
\label{tab:codGen}
\end{table*}

\begin{table*}[htbp]
\caption{\small{The averages and standard deviations (in braces) of performances of different classifiers on COD-RNA dataset when trained with $60\%$ of available data and the experiments are repeated $10$ times. }}
\begin{center}
{\tabulinesep=1.5mm                                           \begin{tabu}{ccccccccc}
\hline
\multicolumn{ 1}{c}{Methods $\downarrow$} & \multicolumn{ 3}{c}{Positive} & & \multicolumn{ 3}{c}{Negative} & \multicolumn{ 1}{c}{Support} \\ \cline{ 2- 4} \cline{6-8}
\multicolumn{ 1}{l}{} & \multicolumn{1}{l}{Precision} & \multicolumn{1}{l}{Recall} & \multicolumn{1}{l}{F-Measure} && \multicolumn{1}{l}{Precision} & \multicolumn{1}{l}{Recall} & \multicolumn{1}{l}{F-Measure} & \multicolumn{ 1}{c}{ Vectors} \\ \hline
CONCAT & 0.71(0.0071) & 0.81(0.0047) & 0.76(0.004) && 0.66(0.0362) & 0.62(0.0069) & 0.64(0.0123) & 0.73(0.2804) \\
F-EC & 0.77(0.0084) & 0.81(0.0612) & 0.79(0.024) && 0.74(0.0607) & 0.68(0.0052) & 0.71(0.0011) & 0.55(0.104) \\
SGMKL & 0.67(0.0216) & 0.57(0.0063) & 0.62(0.0001) && 0.53(0.0353) & 0.55(0.0087) & 0.54(0.0074) & 0.8(0.0099) \\
L-MKL & 0.49(0.0046) & 0.33(0.0057) & 0.4(0.0009) && 0.45(0.0082) & 0.58(0.0147) & 0.51(0.0083) & 0.7(0.0921) \\
F-MKL & 0.74(0.0066) & 0.84(0.042) & 0.79(0.014) && 0.85(0.0001) & 0.79(0.176) & 0.82(0.015) & 0.49(0.0921) \\
S-MKL &  \bf0.91(0.0008) & \bf 0.89(0.0742) &  \bf0.9(0.0141) && \bf 0.88(0.0049) &  \bf0.9(0.0009) &  \bf0.89(0.0157) &  \bf0.29(0.1067) \\ \hline
\end{tabu}}
\end{center}
\label{tab:codPerform}
\end{table*}

\clearpage
\subsection{Adult Dataset}
\label{subsec:adult}
\begin{sidewaysfigure*}[htbp]
\centerline{\includegraphics[width=1\textheight]{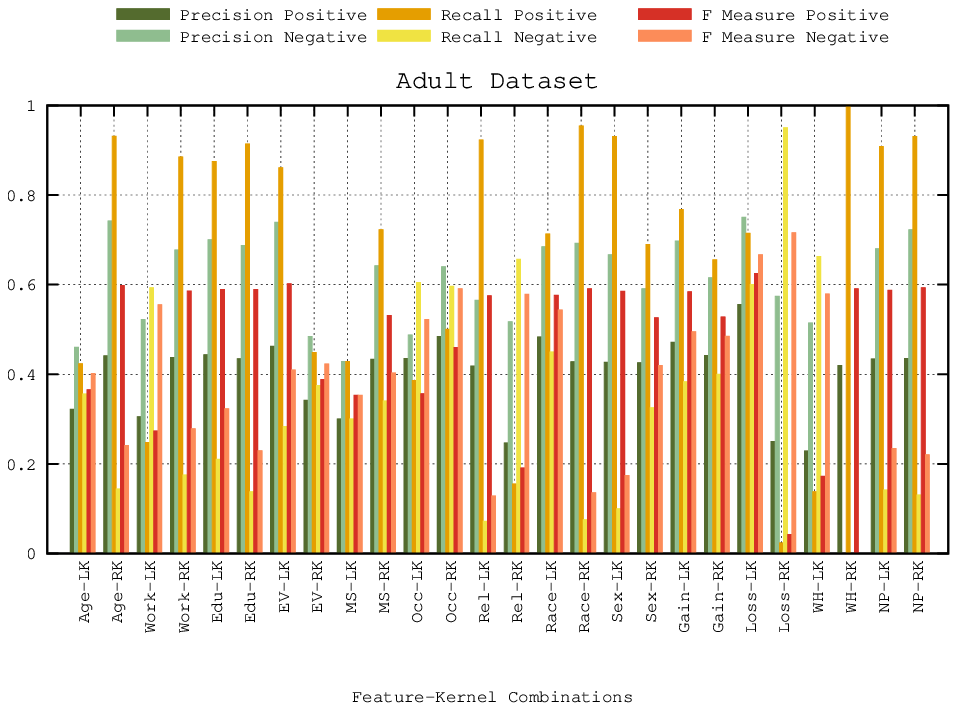}}
\caption{\small{Visualization of the performance analysis data presented in Table~\ref{tab:adultFeature}. The precision, recall and f-measures for different feature kernel combinations are shown for the Adult dataset.}}
\label{fig:adultFeature}
\end{sidewaysfigure*}

\begin{figure*}
\centerline{\includegraphics[width=0.5\textwidth]{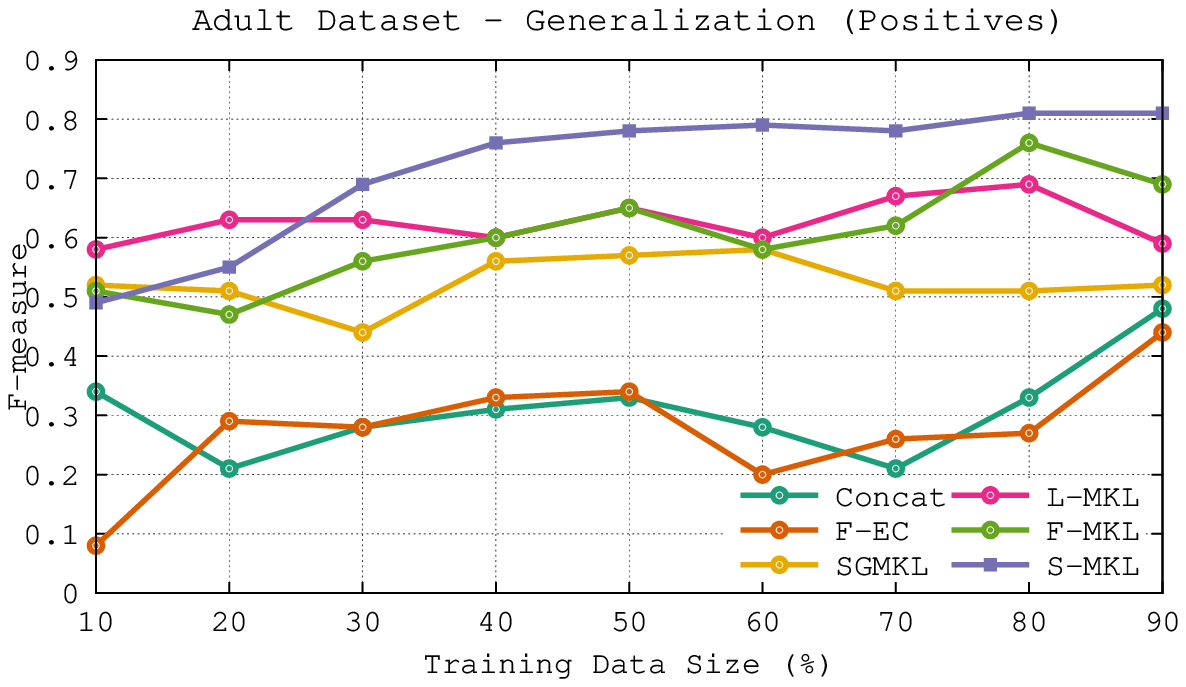}
\includegraphics[width=0.5\textwidth]{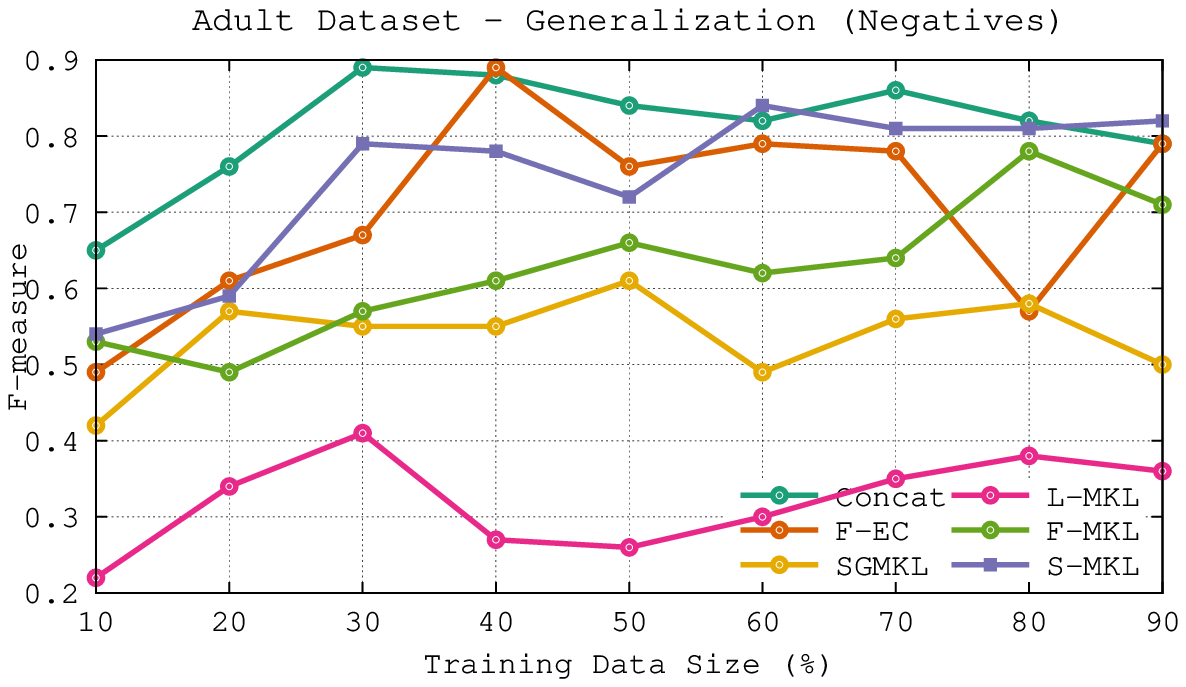}}
\caption{\small{Visualization of generalization performance data presented in Table~\ref{tab:adultGen}. The variations of f-measures for (a) positive and (b) negative categories are presented with respect to changing training set size.}}
\label{fig:adultGen}
\end{figure*}

\begin{figure*}[htbp]
\centerline{\includegraphics[width=1\textwidth]{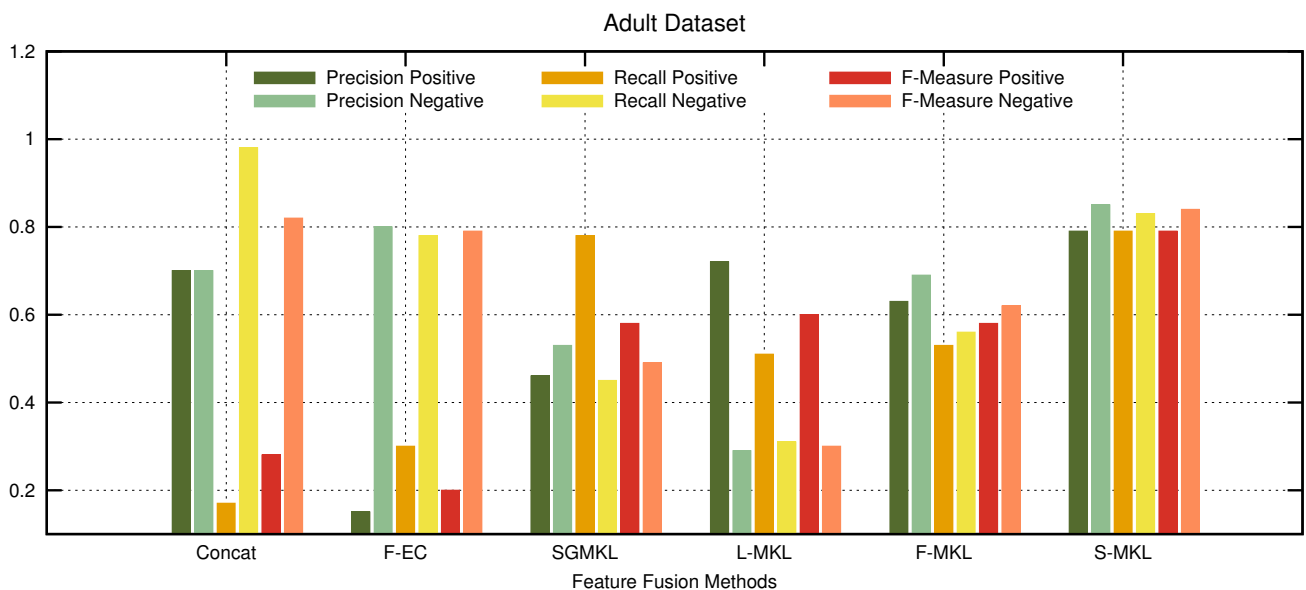}}
\caption{\small{Visualization of the performance analysis data presented in Table~\ref{tab:adultPerform}}}
\label{fig:adultPerform}
\end{figure*}

The Adult dataset consists of $270000$ samples with $24.84\%$ Positive sample. Each sample is represented by $123$ binary and continuous values representing $14$ distinct attributes viz. -- Age, Work, Weight(wgt), Education(Edu), Education Value(EV), Marital Status (MS) Occupation (Occ), Relation (Rel), Race, Sex, Gain , Loss , Work Hours (WH) and Native Place. We have used Linear (LK) and RBF (RK) kernels with each attribute resulting in a total of $28$ feature-kernel combinations.  Performance of individual feature kernel combinations are tabulated in table~\ref{tab:adultFeature} and are visualized in figure~\ref{fig:adultFeature}. Table~\ref{tab:adultGen} and Figure~\ref{fig:adultGen} shows the Generalization  performance of different classifiers on Adult dataset while Table~\ref{tab:adultPerform} and Figure~\ref{fig:adultPerform} presents the detailed performance analysis of different classifiers when trained on $60\%$ of total available data.
\begin{table*}[htbp]
\caption{Feature performance Analysis of Adult dataset}
\begin{center}
{\tabulinesep=1.5mm                                           \begin{tabu}{cccccccc}
\hline

\multicolumn{1}{c}{Features} & \multicolumn{3}{c}{Positive}  & \multicolumn{3}{c}{Negative} \\ \cline{ 2-4} \cline{6-8}
&Precision&Recall&F Measure&&Precision&Recall&F Measure \\ \hline
Age-LK & 0.321637 & 0.423077 & 0.365449 && 0.460432 & 0.355556 & 0.401254 \\
Age-RK & 0.440816 &  \bf0.931034 & 0.598338 &&  \bf0.741935 & 0.14375 & 0.240838 \\
Work-LK & 0.304878 & 0.247525 & 0.273224& & 0.522013 & 0.592857 & 0.555184 \\
Work-RK & 0.4375 & 0.885057 & 0.585551 && 0.677419 & 0.175 & 0.278146 \\
Edu-LK & 0.443662 & 0.875 & 0.588785 && 0.7 & 0.21 & 0.323077 \\
Edu-RK & 0.434426 & 0.913793 & 0.588889 && 0.6875 & 0.1375 & 0.229167 \\
EV-LK & 0.4625 & 0.860465 &  \bf0.601626 && 0.73913 & 0.283333 & 0.409639 \\
EV-RK & 0.342105 & 0.448276 & 0.38806 && 0.483871 & 0.375 & 0.422535 \\
MS-LK & 0.3 & 0.428571 & 0.352941 && 0.428571 & 0.3 & 0.352941 \\
MS-RK & 0.432986 & 0.722473 & 0.531231 && 0.642164 & 0.340201 & 0.402715 \\
Occ-LK & 0.434913 & 0.385872 & 0.356303& & 0.487472 & 0.604935 & 0.522403 \\
Occ-RK & 0.484239 & 0.5004 & 0.459422 && 0.64 & 0.596141 & 0.590562 \\
Rel-LK & 0.418118 & 0.923077 & 0.57554 && 0.565217 & 0.072222 & 0.128079 \\
Rel-RK & 0.246575 & 0.155172 & 0.190476 && 0.517241 & 0.65625 & 0.578512 \\
Race-LK & 0.483221 & 0.712871 & 0.576 && 0.684783 & 0.45 & 0.543103 \\
Race-RK & 0.427835 & 0.954023 & 0.590747& & 0.692308 & 0.075 & 0.135338 \\
Sex-LK & 0.426752 & 0.930556 & 0.585153 && 0.666667 & 0.1 & 0.173913 \\
Sex-RK & 0.425532 & 0.689655 & 0.526316 && 0.590909 & 0.325 & 0.419355 \\
Gain-LK & 0.471429 & 0.767442 & 0.584071 && 0.69697 & 0.383333 & 0.494624 \\
Gain-RK & 0.44186 & 0.655172 & 0.527778 && 0.615385 & 0.4 & 0.484848 \\
Loss-LK &  \bf0.555556 & 0.714286 & 0.625 && 0.75 & 0.6 & 0.666667 \\
Loss-RK & 0.25 & 0.023077 & 0.042253 && 0.573826 & 0.95 &  \bf0.715481 \\
WH-LK & 0.228571 & 0.137931 & 0.172043 && 0.514563 &  \bf0.6625 & 0.579235 \\
WH-RK & 0.419087 & 1 & 0.590643& & 0 & 0 & 0 \\
NP-LK & 0.434066 & 0.908046 & 0.587361 && 0.68 & 0.141667 & 0.234483 \\
NP-RK & 0.435065 & 0.930556 & 0.59292 && 0.722222 & 0.13 & 0.220339 \\ \hline
\end{tabu}}
\end{center}
\label{tab:adultFeature}
\end{table*}

\begin{table*}[htbp]
\caption{Generalization Performance of different algorithms on Adult dataset.}
\begin{center}
{\tabulinesep=1.5mm                                           \begin{tabu}{ccccccccccr}
\hline
\multicolumn{ 2}{|c|}{\backslashbox{Methods$\downarrow$}{Data Size$\rightarrow$}}&10 & 20 & 30 & 40 & 50 & 60 & 70 & 80 & 90 \\ \hline
\multirow{ 2}{*}{Concat}& F+  &0.34 & 0.21 & 0.28 & 0.31 & 0.33 & 0.28 & 0.21 & 0.33 & 0.48 \\
\multicolumn{ 1}{c}{} & F-     	&0.65 & 0.76 & 0.89 & 0.88 & 0.84 & 0.82 & 0.86 & 0.82 & 0.79 \\ \hline
\multirow{2}{*}{F-EC}& F+  	&0.08 & 0.29 & 0.28 & 0.33 & 0.34 & 0.2 & 0.26 & 0.27 & 0.44 \\
\multicolumn{ 1}{c}{} & F-     	&0.49 & 0.61 & 0.67 & 0.89 & 0.76 & 0.79 & 0.78 & 0.57 & 0.79 \\ \hline
\multirow{2}{*}{SG-MKL}&F+ 	&0.52 & 0.51 & 0.44 & 0.56 & 0.57 & 0.58 & 0.51 & 0.51 & 0.52 \\
\multicolumn{ 1}{c}{} & F-  		&0.42 & 0.57 & 0.55 & 0.55 & 0.61 & 0.49 & 0.56 & 0.58 & 0.5 \\ \hline
\multirow{2}{*}{L-MKL}& F+ 	&0.58 & 0.63 & 0.63 & 0.6 & 0.65 & 0.6 & 0.67 & 0.69 & 0.59 \\
\multicolumn{ 1}{c}{} & F-  		&0.22 & 0.34 & 0.41 & 0.27 & 0.26 & 0.3 & 0.35 & 0.38 & 0.36 \\ \hline
\multirow{2}{*}{F-MKL}& F+ 	&0.51 & 0.47 & 0.56 & 0.6 & 0.65 & 0.58 & 0.62 & 0.76 & 0.69 \\
\multicolumn{ 1}{c}{} & F- 		&0.53 & 0.49 & 0.57 & 0.61 & 0.66 & 0.62 & 0.64 & 0.78 & 0.71 \\ \hline
\multirow{2}{*}{S-MKL}& F+ 	&0.49 & 0.55 & 0.69 & 0.76 & 0.78 & 0.79 & 0.78 & 0.81 & 0.81 \\
\multicolumn{ 1}{c}{} & F- 		&0.54 & 0.59 & 0.79 & 0.78 & 0.72 & 0.84 & 0.81 & 0.81 & 0.82 \\ \hline
\end{tabu}}
\end{center}
\label{tab:adultGen}
\end{table*}

\begin{table*}[htbp]
\caption{\small{The averages and standard deviations (in braces) of performances of different classifiers on Adult dataset when trained with $60\%$ of available data and the experiments are repeated $10$ times.}}
\begin{center}
{\tabulinesep=1.5mm                                           \begin{tabu}{ccccccccc}
\hline
\multicolumn{ 1}{c}{Methods $\downarrow$} & \multicolumn{ 3}{c}{Positive} & & \multicolumn{ 3}{c}{Negative} & \multicolumn{ 1}{c}{Support} \\ \cline{ 2- 4} \cline{6-8}
\multicolumn{ 1}{l}{} & \multicolumn{1}{l}{Precision} & \multicolumn{1}{l}{Recall} & \multicolumn{1}{l}{F-Measure} && \multicolumn{1}{l}{Precision} & \multicolumn{1}{l}{Recall} & \multicolumn{1}{l}{F-Measure} & \multicolumn{ 1}{c}{ Vectors} \\ \hline
CONCAT & 0.7(0.105) & 0.17(0.0047) & 0.28(0.02) && 0.7(0.0456) &  \bf0.98(0.0012) & 0.82(0.14) & 0.79(0.0333) \\
F-EC & 0.15(0.2141) & 0.3(0.0409) & 0.2(0.102) && 0.8(0.0012) & 0.78(0.174) & 0.79(0.012) & 0.8(0.2104) \\
SGMKL & 0.46(0.0201) & 0.78(0.0082) & 0.58(0.0018) && 0.53(0.0049) & 0.45(0.305) & 0.49(0.0001) & 0.62(0.0053) \\
L-MKL & 0.72(0.047) & 0.51(0.0059) & 0.6(0.0025) && 0.29(0.0039) & 0.31(0.508) & 0.3(0.5) & 0.56(0.0187) \\
F-MKL & 0.63(0.0248) & 0.53(0.0144) & 0.58(0.019) && 0.69(0.0037) & 0.56(0.0113) & 0.62(0.0001) & 0.54(0.0076) \\
S-MKL & \bf 0.79(0.0332) &  \bf 0.79(0.0541) &  \bf 0.79(0.015) && \bf 0.85(0.0009) & 0.83(0.0009) &  \bf0.84(0.010) &  \bf0.31(0.0025) \\ \hline
\end{tabu}}
\end{center}
\label{tab:adultPerform}
\end{table*}

\clearpage

\subsection{Commercial Database}
\label{subsec:commercial}
\begin{sidewaysfigure*}[htbp]
\centerline{\includegraphics[width=1\textheight]{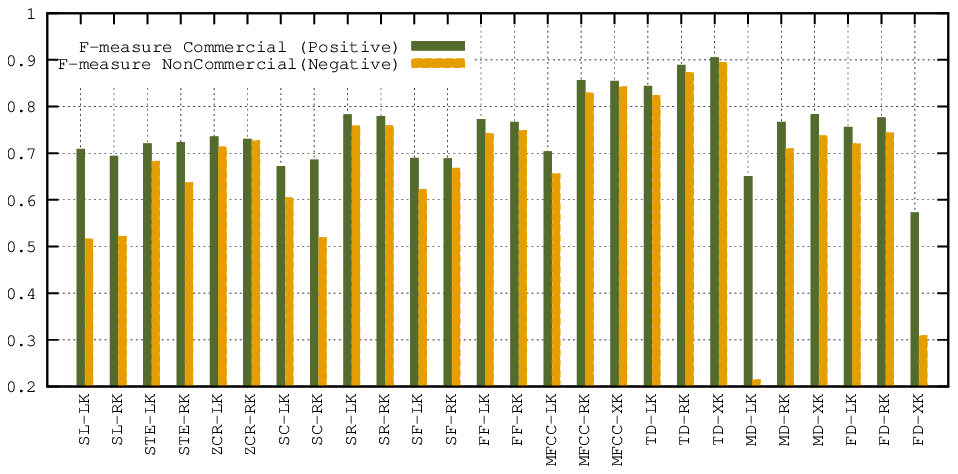}}
\caption{\small{Visualization of the performance analysis data presented in Table~\ref{tab:commercialFeature}. The precision, recall and f-measures for different feature kernel combinations are shown for the Commercial dataset.}}
\label{fig:commercialFeature}
\end{sidewaysfigure*}

\begin{figure*}
\centerline{\includegraphics[width=1\textwidth]{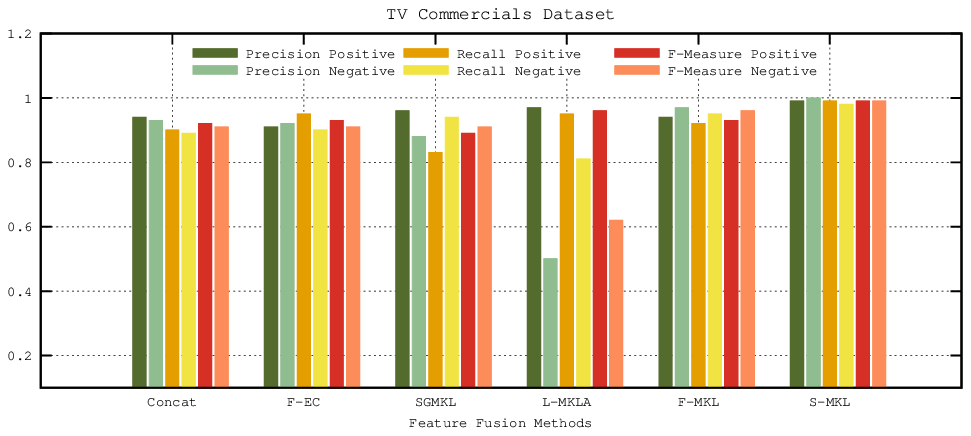}}
\caption{\small{Visualization of the performance analysis data presented in Table~\ref{tab:commercialPerform}}}
\label{fig:commercialPerform}
\end{figure*}

The Adult dataset consists of $1,30410$ samples with $64\%$ Positive sample. Each sample is represented by $4117$ dimensional feature vector representing $11$ distinct attributes of a video shot viz. --  Shot Length (SL), Short time energy(STE) , Zero crossing rate (ZCR), spectral centroid(SC) , spectral roll off (SR), spectral flux (SF), Fundamental Frequency (FF) ,  MFCC Bag of Audio words, (MFCC), Text Distribution (TD), Motion Distribution(MD) and Frame difference(FD). We have used Linear (LK) and RBF (RK) kernels with each attribute  and $\chi^2$ kernel with MFCC , TD, MD, and FD resulting in a total of $26$ feature-kernel combinations.  Performance of individual feature kernel combinations are tabulated in table~\ref{tab:commercialFeature} and are visualized in figure~\ref{fig:commercialFeature}. Table~\ref{tab:commercialGen} and Figure~\ref{fig:commercialGen} shows the Generalization  performance of different classifiers on Commercial dataset while Table~\ref{tab:commercialPerform} and Figure~\ref{fig:commercialPerform} presents the detailed performance analysis of different classifiers when trained on $60\%$ of total available data.

\begin{table*}[htbp]
\caption{Feature performance analysis of Commercial Dataset}
\begin{center}
{\tabulinesep=1.5mm                                           \begin{tabu}{cccccccc}
\hline
\multicolumn{1}{|c|}{Features} & \multicolumn{3}{|c|}{Commercials}  & \multicolumn{3}{|c|}{Non-Commercials} \\ \cline{ 2-7}
&Precision&Recall&F Measure&&Precision&Recall&F Measure \\ \hline
SL-LK & 0.611388 & 0.842166 & 0.708215 && 0.703682 & 0.408906 & 0.516159 \\
SL-RK   & 0.609649 & 0.803649 & 0.693045 && 0.66564 & 0.43097 & 0.522043 \\
STE-LK   & 0.712166 & 0.728676 & 0.719804 && 0.69287 & 0.673743 & 0.682501 \\
STE-RK   & 0.706963 & 0.754294 & 0.722575 && 0.689079 & 0.627839 & 0.636981 \\
ZCR-LK   & 0.744561 & 0.727353 & 0.734674 && 0.707304 & 0.722128 & 0.713309 \\
ZCR-RK   & 0.766339 & 0.698263 & 0.729788 && 0.696002 & 0.762029 & 0.726789 \\
SC-LK   & 0.646622 & 0.698227 & 0.671189 && 0.634555 & 0.578167 & 0.604682 \\
SC-RK   & 0.623457 & 0.771391 & 0.685286 && 0.581053 & 0.47114 & 0.519581 \\
SR-LK   & 0.782196 & 0.783999 & 0.78225 && 0.761506 & 0.756933 & 0.758193 \\
SR-RK   & 0.783589 & 0.774684 & 0.778726 && 0.75487 & 0.76315 & 0.758562 \\
SF-LK   & 0.663339 & 0.719649 & 0.689122 && 0.658423 & 0.593593 & 0.622077 \\
SF-RK   & 0.700379 & 0.677523 & 0.688065 && 0.65727 & 0.679608 & 0.667559 \\
FF-LK   & 0.763587 & 0.782458 & 0.772156 && 0.754464 & 0.73153 & 0.741923 \\
FF-RK   & 0.778201 & 0.759331 & 0.766103 && 0.744201 & 0.7575 & 0.748525 \\
MFCC-LK   & 0.687405 & 0.722497 & 0.703297 && 0.679296 & 0.637486 & 0.655946 \\
MFCC-RK   & 0.827443 & 0.887211 & 0.855649 && 0.867505 & 0.795397 & 0.828922 \\
MFCC-XK   & 0.86052 & 0.852115 & 0.854092 && 0.843012 & 0.845083 & 0.84198 \\
TD-LK   & 0.836876 & 0.849778 & 0.843002 && 0.831505 & 0.816351 & 0.823525 \\
TD-RK   & 0.874281 & 0.903055 & 0.888071 && 0.890738 & 0.85669 & 0.872885 \\
TD-XK   &  \bf0.905058 &  \bf0.904275 &  \bf0.904346 &&  \bf0.894666 &  \bf0.89425 & \bf 0.894094 \\
MD-LK   & 0.53048 & 0.854906 & 0.650084 && 0.371486 & 0.167121 & 0.214643 \\
MD-RK   & 0.729196 & 0.807577 & 0.765914 && 0.758942 & 0.667216 & 0.709307 \\
MD-XK   & 0.753872 & 0.817752 & 0.782925& & 0.781093 & 0.702846 & 0.737551 \\
FD-LK & 0.743288 & 0.769383 & 0.755488 && 0.737028 & 0.706311 & 0.720472 \\
FD-RK   & 0.763488 & 0.790462 & 0.775784 && 0.761931 & 0.72889 & 0.743792 \\
FD-XK   & 0.497437 & 0.678758 & 0.572143 && 0.42743 & 0.251136 & 0.308627 \\ \hline
\end{tabu}}
\end{center}
\label{tab:commercialFeature}
\end{table*}

\begin{table*}[htbp]
\caption{\small{The averages and standard deviations (in braces) of performances of different classifiers on Commercial dataset when trained with $60\%$ of available data and the experiments are repeated $10$ times. }}
\begin{center}
{\tabulinesep=1.5mm                                           \begin{tabu}{ccccccccccr}
\hline
\multicolumn{ 2}{|c|}{\backslashbox{Methods$\downarrow$}{Data Size$\rightarrow$}}& 10 & 20 & 30 & 40 & 50 & 60 & 70 & 80 & \multicolumn{1}{c|}{90} \\ \hline
\multirow{ 2}{*}{Concat} & F+ & 0.89 & 0.88 & 0.91 & 0.92 & 0.91 & 0.92 & 0.92 & 0.9 & 0.92 \\
\multicolumn{ 1}{c}{} & F- & 0.9 & 0.91 & 0.91 & 0.91 & 0.92 & 0.91 & 0.9 & 0.91 & 0.92 \\ \hline
\multirow{2}{*}{F-EC} & F+ & 0.88 & 0.85 & 0.92 & 0.93 & \multicolumn{1}{r|}{0.92} & 0.93 & 0.91 & 0.93 & 0.92 \\
\multicolumn{ 1}{c}{} & F- & 0.89 & 0.88 & 0.9 & 0.91 & 0.9 & 0.91 & 0.9 & 0.9 & 0.9 \\ \hline
\multirow{2}{*}{SG-MKL} & F+ & 0.73 & 0.8 & 0.86 & 0.88 & 0.88 & 0.89 & 0.89 & 0.88 & 0.88 \\
\multicolumn{ 1}{c}{} & F- & 0.69 & 0.79 & 0.8 & 0.86 & 0.75 & 0.91 & 0.89 & 0.9 & 0.91 \\ \hline
\multirow{2}{*}{L-MKL} & F+ & 0.79 & 0.76 & 0.86 & 0.88 & 0.89 & 0.96 & 0.95 & 0.94 & 0.96 \\
\multicolumn{ 1}{c}{} & F- & 0.72 & 0.74 & 0.73 & 0.78 & 0.7 & 0.62 & 0.65 & 0.69 & 0.7 \\ \hline
\multirow{2}{*}{F-MKL} & F+ & 0.87 & 0.88 & 0.86 & 0.9 & 0.92 & 0.93 & 0.91 & 0.92 & 0.89 \\
\multicolumn{ 1}{c}{} & F- & 0.89 & 0.94 & 0.94 & 0.94 & 0.93 & 0.96 & 0.95 & 0.94 & 0.93 \\ \hline
\multirow{2}{*}{S-MKL} & F+ & 0.66 & 0.78 & 0.83 & 0.89 & 0.92 & 0.99 & 0.99 & 1 & 0.99 \\
\multicolumn{ 1}{c}{} & F- & 0.59 & 0.81 & 0.86 & 0.93 & 0.95 & 0.99 & 0.98 & 0.99 & 0.98 \\ \hline
\end{tabu}}
\end{center}
\label{tab:commercialGen}
\end{table*}

\begin{figure*}
\centerline{\includegraphics[width=0.5\textwidth]{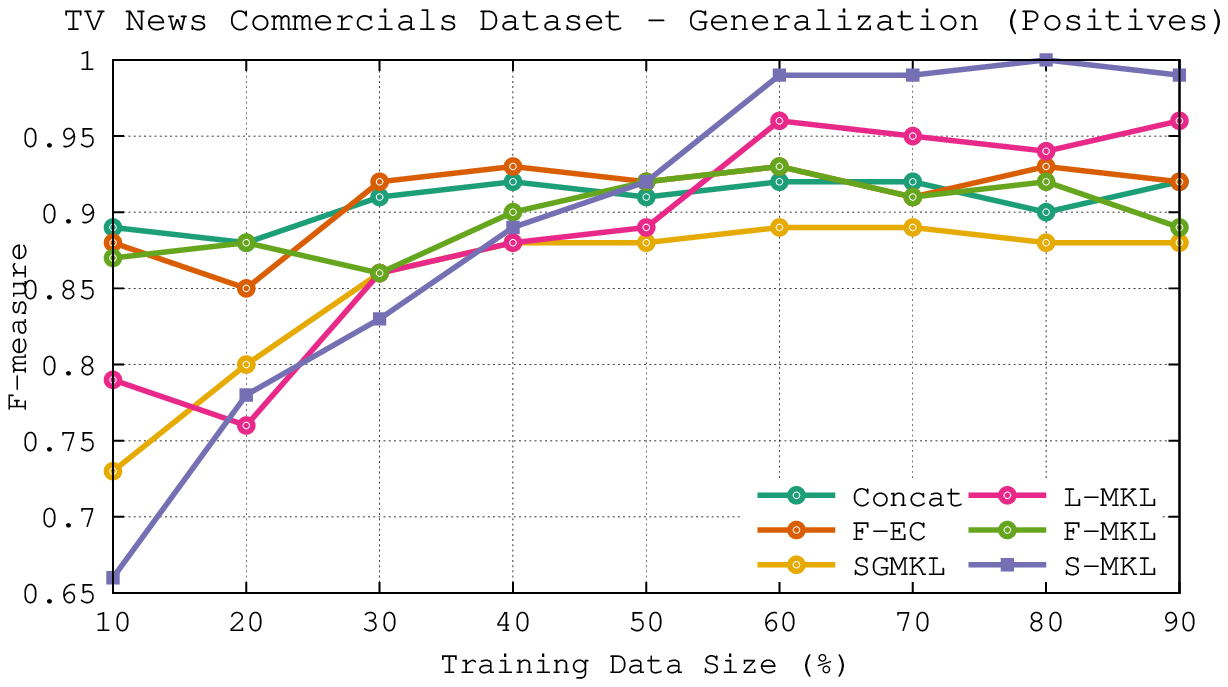}
\includegraphics[width=0.5\textwidth]{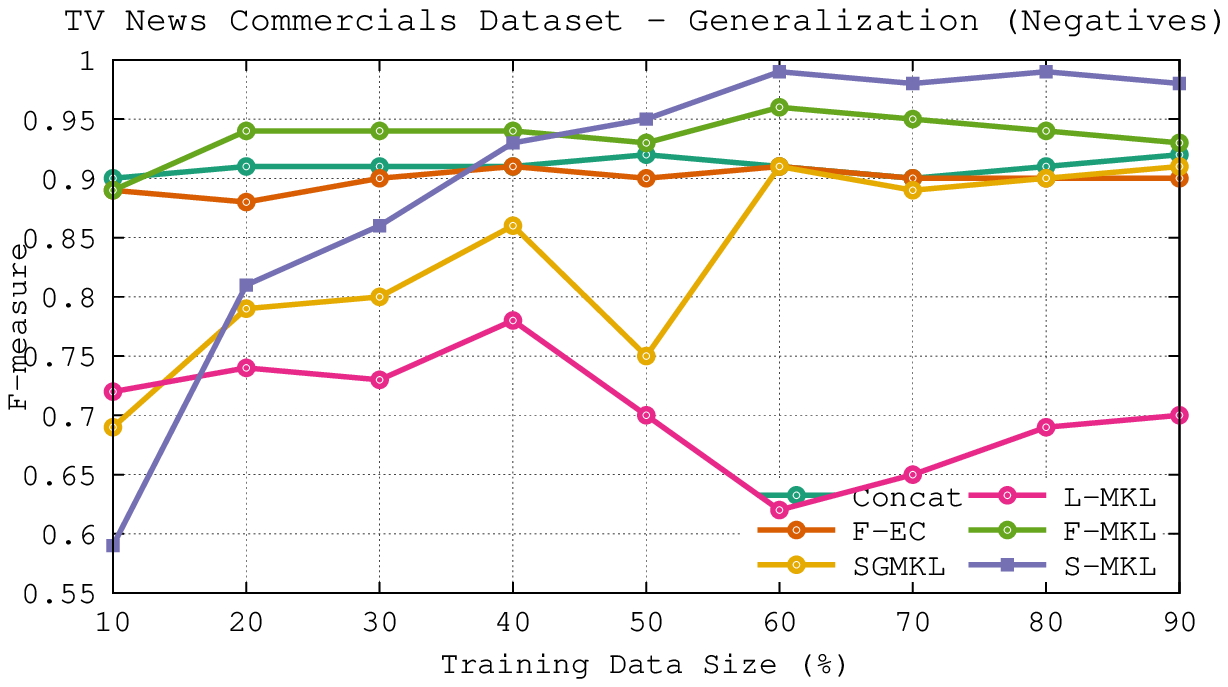}}
\caption{\small{Visualization of generalization performance data presented in Table~\ref{tab:commercialGen}. The variations of f-measures for (a) positive and (b) negative categories are presented with respect to changing training set size.}}
\label{fig:commercialGen}
\end{figure*}


\begin{table*}[htbp]
\caption{}
\begin{center}
{\tabulinesep=1.5mm                                           \begin{tabu}{ccccccccc}
\hline
\multicolumn{ 1}{c}{Methods $\downarrow$} & \multicolumn{ 3}{c}{Positive} & & \multicolumn{ 3}{c}{Negative} & \multicolumn{ 1}{c}{Support} \\ \cline{ 2- 4} \cline{6-8}
\multicolumn{ 1}{l}{} & \multicolumn{1}{l}{Precision} & \multicolumn{1}{l}{Recall} & \multicolumn{1}{l}{F-Measure} && \multicolumn{1}{l}{Precision} & \multicolumn{1}{l}{Recall} & \multicolumn{1}{l}{F-Measure} & \multicolumn{ 1}{c}{ Vectors} \\ \hline
CONCAT & 0.94(0.0109) & 0.90(0.005) & 0.92(0.0001) && 0.93(0.0123) & 0.89(0.0124) & 0.91(0.0001) & 0.51(0.031) \\
F-EC & 0.91(0.0260) & 0.95(0.0126) & 0.93(0.0011) && 0.92(0.0172) & 0.90(0.0246) & 0.91(0.001) & 0.47(0.0761) \\
SGMKL & 0.96(0.0159) & 0.83(0.12) & 0.89(0.009) && 0.88(0.0221) & 0.94(0.0058) & 0.91(0.0001) & 0.57(0.0562) \\
L-MKL & 0.97(0.0013) & 0.95(0.0025) & 0.96(0.0001) && 0.5(0.451) & 0.81(0.0055) & 0.62(0.0014) & 0.68(0.0902) \\
F-MKL & 0.94(0.0610) & 0.92(0.0038) & 0.93(0.0004) && 0.97(0.0049) & 0.95(0.0438) & 0.96(0.0004) & 0.6(0.0834) \\
S-MKL &  \bf0.99(0.0001) & \bf 0.99(0.0021) &  \bf0.99(0.0001) &&  \bf1(0.0003) &  \bf0.98(0.0039) &  \bf0.99(0.0002) &  \bf0.32(0.0057) \\ \hline
\end{tabu}}
\end{center}
\label{tab:commercialPerform}
\end{table*}

\end{document}